%% file: main.tex
\documentclass[runningheads]{llncs}

\usepackage[mobile]{eccv}

\usepackage{eccvabbrv}
\usepackage{tabularx}
\usepackage{multirow}

\usepackage{graphicx}
\usepackage{booktabs}
\usepackage[dvipsnames]{xcolor}

\usepackage[accsupp]{axessibility}  % Improves PDF readability for those with disabilities.

\usepackage{hyperref}

\usepackage{orcidlink}
\usepackage[linesnumbered,ruled,vlined]{algorithm2e}
\usepackage{sidecap}

\newcolumntype{Y}{>{\centering\arraybackslash}X}
\newcommand*{\V}[1]{\mathbf{#1}}

\SetAlCapHSkip{0pt}
\SetKwInput{KwInput}{Inputs}
\SetKwInput{KwOutput}{Outputs}
\SetKwInOut{KwParam}{Parameters}

\newcommand*{\methodname}[0]{\textsc{ReGround}}

\begin{document}

\title{ReGround: Improving Textual and Spatial Grounding at No Cost} 

\titlerunning{ReGround: Improving Textual and Spatial Grounding at No Cost}

\author{Phillip Y. Lee\orcidlink{0009-0000-7614-261X} $\quad$
Minhyuk Sung\orcidlink{0000-0001-7428-9570}}

\authorrunning{P.Y.~Lee and M.~Sung}

\institute{KAIST\\
\email{\{phillip0701,mhsung\}@kaist.ac.kr}
}

\maketitle

\input{figures/teaser.tex}

\input{sections/00_abstract}
\input{sections/01_introduction}
\input{sections/02_related_work}
\input{sections/03_background}
\input{sections/04_method}
\input{sections/05_experiments}
\input{sections/06_conclusion}

\bibliographystyle{splncs04}
\bibliography{main}

\clearpage

\noindent
\section{Appendix}
In this supplementary material, we provide additional details on the evaluation setup (Sec.~\ref{subsec:supp_eval_setup}) and more quantitative comparisons of our \methodname{} and GLIGEN~\cite{li2023gligen} (Sec.~\ref{subsec:supp_more_quantitative}). Moreover, we showcase the effect of \methodname{} as a backbone of zero-shot layout-guided image generation methods (Sec.~\ref{subsec:supp_more_backbone}) and finally provide extensive qualitative comparisons of Stable Diffusion (SD)~\cite{rombach2022high}, GLIGEN~\cite{li2023gligen}, and our \methodname{} (Sec.~\ref{subsec:supp_more_qualitative}).

\input{sections_supp/01_dataset_details}
\input{sections_supp/02_additional_quantitative}
\input{sections_supp/03_more_backbones}
\input{sections_supp/04_additional_qualitative}

\end{document}

%% file: figures/teaser.tex
{
    \centering
    \footnotesize{
        \renewcommand{\arraystretch}{0.5}
        \setlength{\tabcolsep}{0.0em}
        \setlength{\fboxrule}{0.0pt}
        \setlength{\fboxsep}{0pt}
        
        \begin{tabularx}{\textwidth}{>{\centering\arraybackslash}m{0.20\textwidth} >{\centering\arraybackslash}m{0.20\textwidth} >{\centering\arraybackslash}m{0.20\textwidth} >{\centering\arraybackslash}m{0.20\textwidth} >{\centering\arraybackslash}m{0.20\textwidth}}
        Layout & SD & GLIGEN$_{\gamma=1.0}$ & GLIGEN$_{\gamma=0.2}$ & \textbf{ReGround} \\
        %%%%%%%%%%%%%%%%%%%%%%%%%
        \multicolumn{5}{c}{\includegraphics[width=\textwidth]{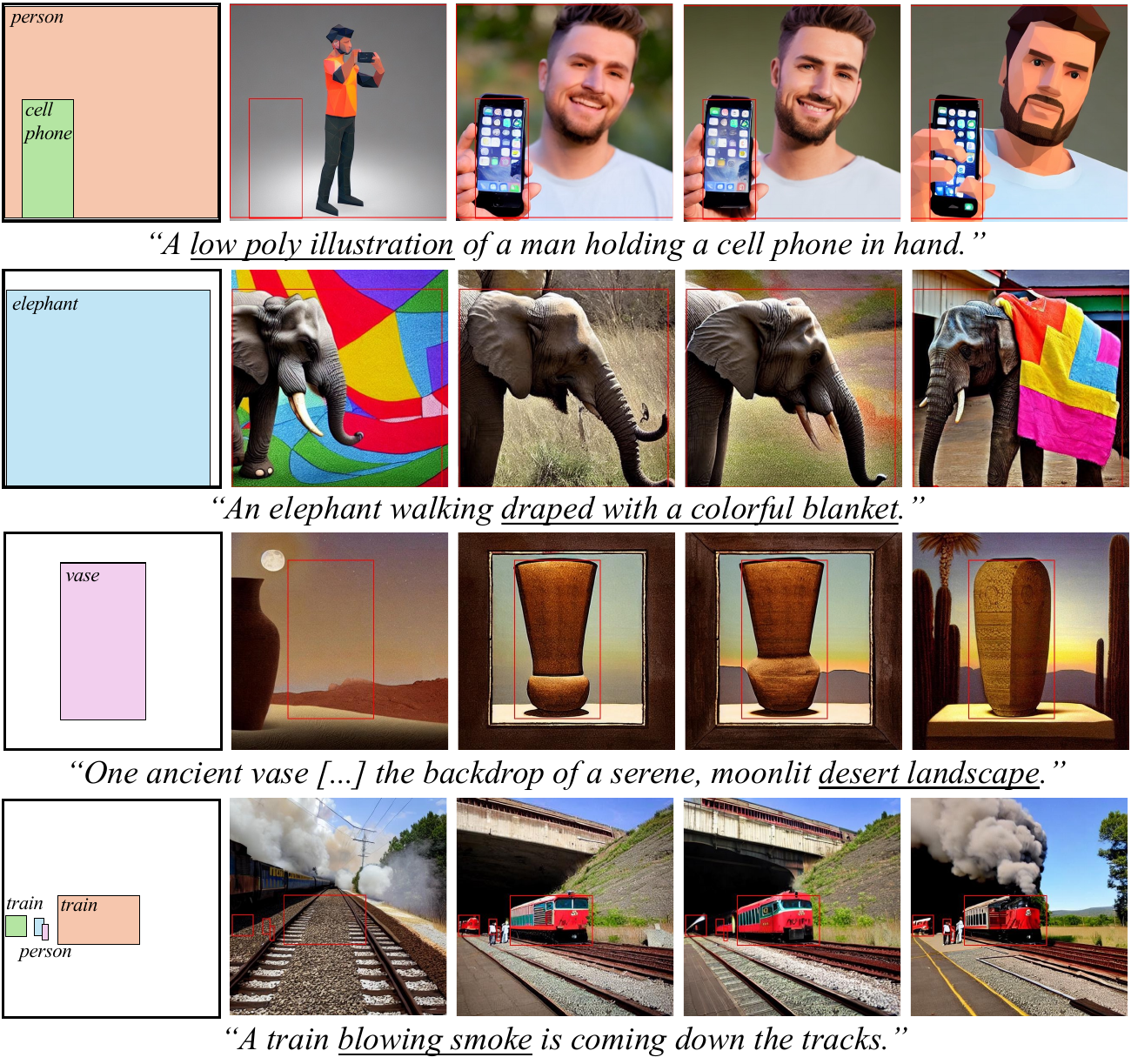}} \\
        %%%%%%%%%%%%%
        \end{tabularx}
    }
    \captionof{figure}{Comparison across Stable Diffusion (SD)\cite{rombach2022high}, GLIGEN\cite{li2023gligen}, and our \methodname{}. SD (2nd column) can generate an image aligned with the input prompt (shown below each row), while it does not allow taking spatial constraints such as bounding boxes and labels. GLIGEN (3rd column) enables spatial grounding using gated self-attention, although it often disregards some descriptions in the input prompt due to a bias towards bounding box conditions. Such trends also occur when only activating gated self-attention for 0.2 fraction of the initial denoising steps (4th column). Our~\methodname{} (last column) resolves the issue of description omission while accurately reflecting the bounding box information.}
    \label{fig:teaser}
}
% -----------------------------------------------------------------------

%% file: sections/00_abstract.tex
\begin{abstract}
When an image generation process is guided by both a \emph{text} prompt and \emph{spatial} cues, such as a set of bounding boxes, do these elements work in harmony, or does one dominate the other? Our analysis of a pretrained image diffusion model that integrates gated self-attention into the U-Net reveals that spatial grounding often outweighs textual grounding due to the \emph{sequential} flow from gated self-attention to cross-attention. We demonstrate that such bias can be significantly mitigated without sacrificing accuracy in either grounding by simply rewiring the network architecture, changing from sequential to \emph{parallel} for gated self-attention and cross-attention. This surprisingly simple yet effective solution does not require any fine-tuning of the network but substantially reduces the trade-off between the two groundings. Our experiments demonstrate significant improvements from the original GLIGEN to the rewired version in the trade-off between textual grounding and spatial grounding. The project webpage is at \url{https://re-ground.github.io}.
\keywords{Textual Grounding \and Spatial Grounding \and Network Rewiring}
\end{abstract}

%% file: sections/01_introduction.tex
\section{Introduction}
\label{sec:intro}
The emergence of diffusion models~\cite{ho2020DDPM,song2021SDE,song2021DDIM} has markedly propelled the field of text-to-image (T2I) generation forward, allowing users to generate high-quality images from text prompts. In a bid to further augment the creativity and controllability, recent efforts~\cite{li2023gligen,yang2023reco,xie2023boxdiff,phung2023grounded,ma2023directed,chen2024training,bansal2024universal,balaji2022ediffi,kim2023dense,bar2023multidiffusion,couairon2023zero,zhang2023adding,avrahami2023spatext} have focused on enabling these models to understand and interpret \emph{spatial instructions}, such as 
layouts~\cite{li2023gligen,yang2023reco,xie2023boxdiff,phung2023grounded,ma2023directed,chen2024training,bansal2024universal}, segmentation masks~\cite{balaji2022ediffi,kim2023dense,bar2023multidiffusion,couairon2023zero,bansal2024universal,avrahami2023spatext} and sketches~\cite{voynov2023sketch,zhang2023adding}.

Among them, \emph{bounding boxes} are extensively employed in downstream image generation tasks~\cite{li2023gligen,yang2023reco,phung2023grounded,ma2023directed,chen2024training,bansal2024universal}. GLIGEN~\cite{li2023gligen} is a pioneering work in terms of enhancing existing T2I models with the capability to incorporate additional spatial cues in the form of bounding boxes. 
Its core component, \emph{gated self-attention}, is a simple attention module~\cite{vaswani2017attention} that is plugged into each U-Net~\cite{ronneberger2015unet} layer of a pretrained T2I model such as Stable Diffusion~\cite{rombach2022high}, and is trained to accurately position various entities in their designated areas. A notable advantage of GLIGEN is that the original parameters of the underlying model remain unchanged, inheriting the generative capability of the T2I model while introducing the novel functionality of spatial grounding using bounding boxes. This capability has been leveraged by numerous studies to facilitate high-quality, layout-guided image generation~\cite{feng2023layoutgpt, xie2023boxdiff, chen2023llava, liu2023grounding}.

However, our analysis reveals that GLIGEN's integration of the gated self-attention into an existing T2I model is not optimal for blending new spatial guidance from bounding boxes with the original textual guidance.
It often leads to the \emph{omission of specific details} from the text prompts.
For instance, in the first row and third column of Fig.~\ref{fig:teaser}, GLIGEN fails to reflect the description \textit{``low poly illustration''} from the input text prompt. Also in the second row, a crucial detail in the text prompt, \textit{``draped with a colorful blanket''}, is neglected in the output image. We refer to this issue as \textbf{description omission}. Such outcomes imply that the current architectural design of GLIGEN does not effectively harmonize the new spatial guidance and the text conditioning in the given T2I model. Considering the widespread applications of GLIGEN in various layout-based generation tasks~\cite{feng2023layoutgpt, xie2023boxdiff, phung2023grounded, zhao2023loco, xiao2023r, chen2023llava, liu2023grounding}, these limitations represent a significant bottleneck.

To address the observed neglect of textual grounding in GLIGEN, we first analyze the root causes. Our investigation reveals that the issue arises from the \emph{sequential} arrangement of the spatial grounding and textual grounding modules. Specifically, the output of the gated self-attention is directed to a cross-attention module in each layer of the U-Net architecture (Fig.~\ref{fig:pipeline_image}-(b)). 

Building on this insight, we propose a straightforward yet impactful solution: \emph{network rewiring}. This approach alters the relationship between the two grounding modules from sequential to \emph{parallel} (Fig.~\ref{fig:pipeline_image}-(c)). Remarkably, this network modification significantly reduces the grounding trade-off between textual and spatial groundings without necessitating any adjustments to the network parameters. Importantly, \textbf{this rewiring does \emph{not} require additional network training, extra parameters, or changes in computational load and time.} Simply reconfiguring the attention modules of the pretrained GLIGEN, originally trained with the sequential architecture, during inference dramatically enhances performance.

In our experiments on MS-COCO~\cite{lin2014mscoco} and our newly introduced NSR-1K-GPT datasets, we demonstrate that rewiring the pretrained GLIGEN substantially reduces the trade-off between textual and spatial groundings. This is evidenced by the evaluation of text prompt alignment (measured using CLIP score~\cite{radford2021clip}, PickScore~\cite{kirstain2024pick} and user study) and bounding box alignment (assessed by YOLO score~\cite{wang2023yolov7}). Furthermore, we show that our rewiring also leads to better outcomes in other frameworks using GLIGEN as a backbone, including BoxDiff~\cite{xie2023boxdiff}.

%% file: sections/02_related_work.tex
\section{Related Work}
\label{sec:related_work}

\subsection{Zero-Shot Guidance in Diffusion Models}
\label{subsec:diffusion_zero_shot}
The progress in diffusion models~\cite{ho2020DDPM,song2021SDE,song2021DDIM} has significantly elevated the capabilities of text-to-image (T2I) generation, resulting in foundation models~\cite{rombach2022high,podell2023sdxl,ramesh2021dalle,ramesh2022dalle2,betker2023dalle3} that exhibit remarkable generative performance. Leveraging the robust performance of these models, recent studies~\cite{li2023gligen,yang2023reco,xie2023boxdiff,phung2023grounded,ma2023directed,chen2024training,bansal2024universal,balaji2022ediffi,kim2023dense,bar2023multidiffusion,couairon2023zero,zhang2023adding} have introduced efficient guidance techniques designed to further improve the image generation process.
Notably, numerous works~\cite{li2023gligen,yang2023reco,xie2023boxdiff,phung2023grounded,ma2023directed,chen2024training,si2023freeu} focus on the internal architecture (Fig.~\ref{fig:pipeline_image}-(a)) of the denoising U-Net of Latent Diffusion Models (LDMs)~\cite{rombach2022high}, where self-attention and cross-attention modules are intertwined to facilitate inter-pixel communication and text conditioning.
The self-attention of U-Net can be utilized to improve image quality~\cite{hong2023improving} or facilitate image translation~\cite{tumanyan2023plug} and image editing tasks~\cite{cao2023masactrl}. Since text conditions are integrated via cross-attention, the intermediate attention maps have been leveraged to improve text faithfulness~\cite{hertz2023prompt} or enable spatial manipulation of the generation process~\cite{parmar2023zeroshot}. Recently, FreeU~\cite{si2023freeu} analyzed the contributions of the backbone and residuals of the U-Net and proposed a \emph{free-lunch} strategy to enhance image quality: reweighting the backbone and residual features maps. In contrast to previous works that only deal with self- and cross-attention in standard LDMs, we introduce a method to enhance GLIGEN~\cite{li2023gligen} by reconnecting its gated self-attention with the other attention modules, thereby achieving performance improvement in zero-shot without any tuning of the network parameters.

\subsection{Layout-Guided Image Generation}
The use of layouts, particularly in the form of bounding boxes, has become a popular intermediary to bridge the gap between textual inputs and the images generated~\cite{zhao2019image, sun2019image, li2019object, li2021image, hong2018inferring, johnson2018image, sylvain2021object, herzig2020learning, yang2022modeling}.
Layout2Im~\cite{zhao2019image} samples object latent codes from a normal distribution, eliminating the need to predict instance masks as done in prior works~\cite{hong2018inferring, johnson2018image}.
LostGAN~\cite{sun2019image} controls the style of each object by devising an extension of the feature normalization layer used in StyleGAN~\cite{karras2019stylegan,karras2020stylegan2,karras2021stylegan3}, while  OC-GAN~~\cite{sylvain2021object} incorporates the spatial relationships between objects using a scene-graph representation.
LAMA~\cite{li2021image} introduces a mask adaptation module that mitigates the semantic ambiguity arising from overlaps in the input layout.
While these developments have greatly improved user control over image generation, their applicability is confined to the categories found in the training data, such as those of the MS-COCO~\cite{lin2014mscoco} dataset.

In contrast, recent studies~\cite{li2023gligen, yang2023reco, chen2024training, phung2023grounded, xie2023boxdiff, zheng2023layoutdiffusion, bar2023multidiffusion, cheng2023layoutdiffuse} have extended layout-guided image generation towards \textit{open-vocabulary}, building on the advancements of foundational text-to-image (T2I) models~\cite{rombach2022high}.
Training-free approaches~\cite{xie2023boxdiff, balaji2022ediffi, phung2023grounded, chen2024training, bar2023multidiffusion} aim to improve the spatial grounding of T2I models through straightforward guidance mechanisms.
GLIGEN~\cite{li2023gligen}, on the other hand, introduces gated self-attention, which is injected into the U-Net architecture of the Latent Diffusion Model~\cite{rombach2022high}, and is trained to equip the underlying model with spatial grounding abilities. Given the simple architecture of GLIGEN and its robust grounding accuracy with the input bounding boxes, numerous studies~\cite{phung2023grounded,xie2023boxdiff,zhao2023loco,xiao2023r} build upon its framework and propose further refinements to increase performance. In this work, we identify and address a significant performance bottleneck in GLIGEN related to description omission and propose a simple yet effective solution.

%% file: sections/03_background.tex
\section{Background --- Latent Diffusion Models~\cite{rombach2022high}}
\label{sec:ldm_architecture}

\begin{figure}[t!]
    \centering
    \includegraphics[width=\textwidth]{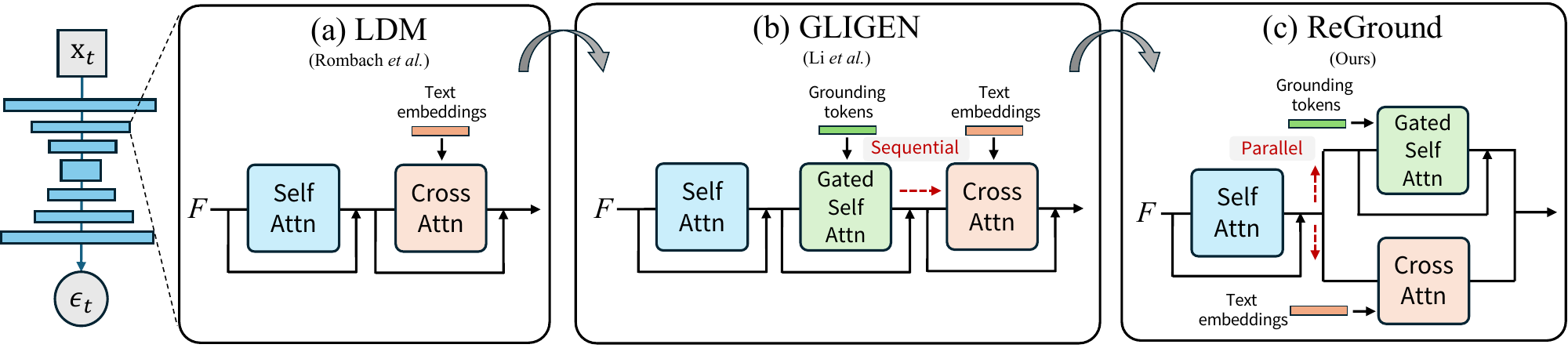}
    \caption{Comparison between the U-Net architectures of (a) Latent Diffusion Model (LDM)~\cite{rombach2022high}, (b) GLIGEN~\cite{li2023gligen} and (c) our \methodname{}. From LDM, GLIGEN enables spatial grounding by injecting Gated Self-Attention before cross-attention, forming a \textit{sequential} flow of them. Based on GLIGEN, our \methodname{} changes the relationship of the two attention modules to become \textit{parallel}, resulting in noticeable improvement in textual grounding while preserving the spatial grounding capability. (The residual block before self-attention is omitted.)}
    \label{fig:pipeline_image}
\end{figure}

Rombach \textit{et al.}~\cite{rombach2022high} proposed Latent Diffusion Model (LDM), a text-to-image (T2I) diffusion model with a U-Net as the noise prediction network. It is trained to generate an image from an input text prompt by predicting the noise $\epsilon (\V{x}_t, t, c)$ conditioned both on the timestep $t$ and the text embedding $c$. Each layer of LDM's U-Net consists of three core components: a convolutional residual block, followed by a self-attention ($\text{SA}$), and a cross-attention ($\text{CA}$) module (Fig.~\ref{fig:pipeline_image}-(a)).
In each $l$-th layer of the U-Net, its residual block first extracts intermediate visual features $F = (f_1, ..., f_{N_l})^T$ from the output of the previous layer.
The self-attention module then facilitates interaction between the features in $F$. Subsequently, the cross-attention module enables the interaction between each visual feature $f_i$ and the text embedding $c$.
Throughout this process, the output feature of the previous module is also forwarded through a residual connection, as illustrated in lines 4-5 and 7 of Alg.~\ref{alg:gligen_pseudocode} and also in Fig.~\ref{fig:pipeline_image}-(a).

\section{GLIGEN~\cite{li2023gligen} and Description Omission}
\label{sec:gsa_intro}
In this section, we review GLIGEN~\cite{li2023gligen} and its key idea of employing gated self-attention for spatial grounding. Then, we present our key observations on the description omission issue that occurs due to the addition of gated self-attention.

\subsection{Gated Self-Attention}
\label{subsec:gsa}
Li \textit{et al.}\cite{li2023gligen} propose a plug-in spatial grounding module, named \textit{gated self-attention}, which adopts the gated attention mechanism\cite{alayrac2022flamingo} to equip a pretrained T2I diffusion model~\cite{rombach2022high} with spatial grounding capabilities (Fig.~\ref{fig:pipeline_image}-(b)). Given a set of bounding boxes and text labels for each of them, let $b_i$ be the $xy$-coordinates of the $i$-th bounding box's top-left and bottom-right corners, and $p_i$ be the corresponding text label.
Then, the $i$-th grounding token is defined as $g_i := \mathcal{G} \left(\mathcal{T}\left(p_i\right), \mathcal{F} \left(b_i\right)\right)$, where $\mathcal{T} (\cdot)$ is a pretrained text encoder~\cite{radford2021clip,ilharco2021openclip}, $\mathcal{F} (\cdot)$ is the Fourier embedding~\cite{mildenhall2020nerf,tancik2020fourfeat} and $\mathcal{G} (\cdot, \cdot)$ is a shallow MLP network that concatenates the two given embeddings, respectively. Given a set of grounding tokens $\{ g_i \}$, gated self-attention learns the self-attention among the unified feature set $(f_1, ..., f_{N_l}, g_1,..., g_M)$ where $\{ f_i \}$ is the set of intermediate visual features in the $l$-th layer of U-Net, and $M$ is the number of bounding boxes.

As shown in Fig.~\ref{fig:pipeline_image}-(b), gated self-attention receives the output of the self-attention along with the residual features as its input and forwards the output features to the cross-attention module. By incorporating gated self-attention into each layer of the U-Net, the model enables the placement of the entity specified in the text label $p_i$ at the location indicated by the bounding box $b_i$. Note that the integration of gated self-attention does not require training the network from scratch or fine-tuning it, but can be accomplished simply by training the gated self-attention parameters while keeping all other parameters in the backbone model frozen.

Alg.~\ref{alg:gligen_pseudocode} shows the pseudocode of the U-Net forward-pass including the plug-in of gated self-attention in line 6.
Note that $\beta_t$ is set to 1 for GLIGEN. If $\beta_t = 0$, the algorithm is identical to that of LDM~\cite{rombach2022high}.

% -----------------------------------
\begin{algorithm}[h!]
\caption{Noise Prediction U-Net with Gated Self-Attention.}
\label{alg:gligen_pseudocode} 
{
    \footnotesize
    \KwParam{$\beta_{t} $\tcp*{Weight for GSA.}}
    \KwInput{$\V{x}_{t}, c, \{ g_i \}_{i=0 \cdots N-1}$\tcp*{Noisy data at timestep $t$, text condition, and grounding tokens}}
    \KwOutput{$ \V{\epsilon}_{t} $\tcp*{Noise at timestep $t-1$.}}
    
    % Set Function Names
    \SetKwFunction{FSD}{U-Net}
    
    % Write Function with word ``Function''
    \SetKwProg{Fn}{Function}{:}{}
    \Fn{\FSD{$ \V{x}_{t}, c, \{ g_i \} $ }}{
        $F \leftarrow \V{x}_t$ \\
        \For{$i = 0, \dots, L-1$} {
            $F_{\text{RS}} \leftarrow \text{Conv} (F) + F $\tcp*{Residual block.}
            $F_{\text{SA}} \leftarrow \text{SA} (F_{\text{RS}}) + F_{\text{RS}} $\tcp*{Self-Attention module.}
            $F_{\text{GSA}} \leftarrow \beta_t \cdot \text{GSA} (F_{\text{SA}}, \{ g_i \}) + F_{\text{SA}} $\tcp*{Gated Self-Attention module.}
            $F \leftarrow \text{CA} (F_{\text{GSA}}, c) + F_{\text{GSA}} $\tcp*{Cross-Attention module.}
        }
        $\V{\epsilon}_{t} \leftarrow F$\;
        \KwRet $ \V{\epsilon}_{t} $\;
    }
}
\end{algorithm}
% -----------------------------------

\subsection{Description Omission}
\label{subsec:prompt_neglection}
Despite its high accuracy in spatial grounding, GLIGEN~\cite{li2023gligen} frequently struggles to capture essential attributes specified in the input text prompt. As illustrated in Fig.~\ref{fig:gligen_trade-off_image}, the leftmost image shows \textit{``a person''} and \textit{``a skateboard''} accurately placed in their designated regions. However, a critical detail from the input text prompt, \textit{``black and white photography''}, is absent in the output image. This discrepancy often emerges when the input comprises distinct but equally important descriptions regarding the image, presented through text prompts and bounding boxes. Such omissions not only fail to convey the stylistic intent of the image but also tend to overlook significant objects mentioned within the text prompt. Additional examples of this problem are showcased in Fig.~\ref{fig:teaser}, where the second row demonstrates the absence of a \textit{``blanket''} in the generated image, a key element from the text prompt. This limitation significantly hampers GLIGEN's fidelity to user-provided text prompts, a challenge we term as \textbf{description omission}.

%% file: sections/04_method.tex
\section{ReGround: Rewiring Attention Modules}
\label{sec:method}

\begin{figure}[t!]
    \centering
    \includegraphics[width=\textwidth]{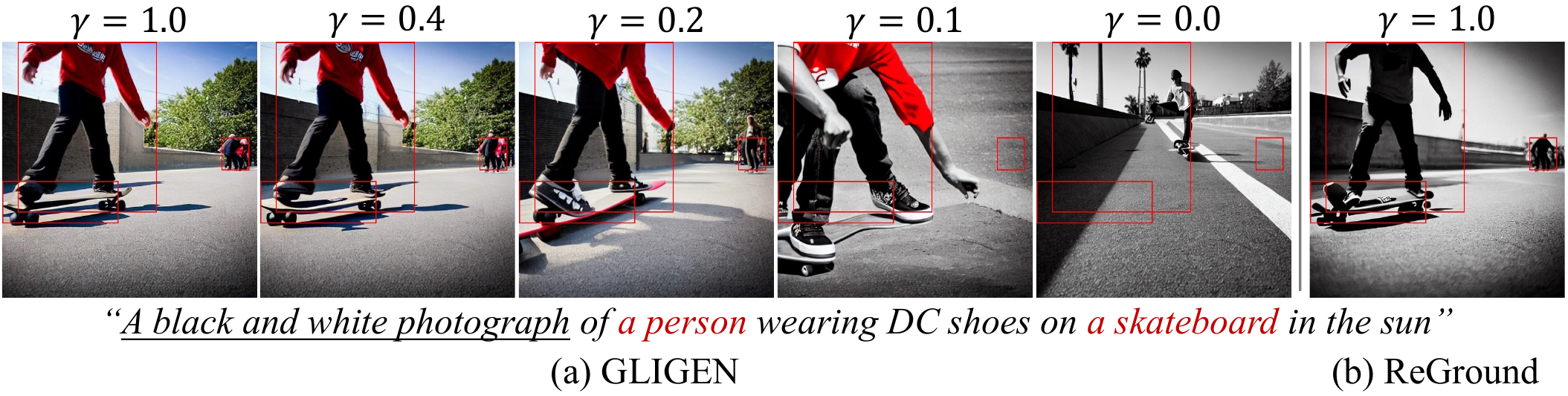}
    \caption{(a) Images generated by GLIGEN~\cite{li2023gligen} with varying activation duration of gated self-attention $\gamma$ in scheduled sampling (Sec.~\ref{subsec:grounding_trade-off}). The \textbf{\textcolor{BrickRed}{red}} words in the text prompt denote the words used as labels of the input bounding boxes. Note that for GLIGEN to reflect the \underline{underlined} description in the text prompt in the final image, $\gamma$ must be decreased to 0.1, which compromises spatial grounding accuracy. (b) In contrast, our \methodname{} reflects the \underline{underlined} phrase even when $\gamma=1.0$, therefore achieving high accuracy in both textual and spatial grounding.}
    \label{fig:gligen_trade-off_image}
\end{figure}

Gated self-attention and cross-attention each play a crucial role in enabling spatial and textual groundings, by taking bounding boxes and text prompts as inputs, respectively. To tackle the issue of description omission, we first examine the impact of attention modules on the groundings they do not address: the effect of gated self-attention on textual grounding (Sec.~\ref{subsec:grounding_trade-off}), and the influence of cross-attention on spatial grounding (Sec.~\ref{subsec:grounding_analysis}). Building on this analysis, we propose an approach for network reconfiguration, modifying the connections among self-attention, gated self-attention, and cross-attention modules (Sec.~\ref{subsec:network_rewiring}).

\subsection{Impact of Gated Self-Attention on Textual Grounding}
\label{subsec:grounding_trade-off}
As the issue of description omission arises due to the newly added gated self-attention in GLIGEN~\cite{li2023gligen}, we first attempt to mitigate the impact of gated self-attention by using \emph{scheduled sampling}~\cite{li2023gligen}, activating gated self-attention only in a few initial steps of the denoising process. This approach is inspired by the observation that the coarse structure of the final image is established within the first few denoising steps. The scheduling is applied by setting the weight of gated self-attention $\beta_t$ (line 6 of Alg.~\ref{alg:gligen_pseudocode}) as
\begin{equation}
    \beta_t =\begin{cases} 1 & (t \leq \gamma \cdot T) \\
                     0 &  (t > \gamma \cdot T),
       \end{cases}
\end{equation}
where $\gamma \in [0, 1]$ represents the fraction of the initial denoising steps to activate gated self-attention.

Fig.~\ref{fig:gligen_trade-off_image}-(a) shows an example of generated images while incrementally adjusting $\gamma$ from 1.0 to 0.0. As $\gamma$ is reduced from 1.0 to 0.0, the details specified in the text prompt, \textit{``a black and white photograph''}, begin to be reflected starting at $\gamma=0.1$, demonstrating that longer activation of gated self-attention may interfere with the alignment of the output image with the text prompt. However, as gated self-attention is activated for shorter durations, the spatial grounding diminishes, as shown in the objects' reduced alignment with the input bounding boxes. This phenomenon illustrates the inherent trade-off between spatial and textual grounding, which cannot be resolved by controlling the duration of gated self-attention activation.

\subsection{Impact of Cross-Attention on Spatial Grounding} 
\label{subsec:grounding_analysis}
We also investigate whether cross-attention has influence on spatial grounding. For this, we conduct a toy experiment by removing cross-attention modules in GLIGEN~\cite{li2023gligen}, allowing the output of the gated self-attention to be directly passed to the next layer of the U-Net. This modification is equivalent to changing line 7 of Alg.~\ref{alg:gligen_pseudocode} to $F \leftarrow F_{GSA}$.

The results are displayed in Fig.~\ref{fig:ca_reweight_experiment}. Note that, while the appearance of the background and objects changes, the silhouettes of the cat (left) and the individuals (right) remain precisely positioned within their respective bounding boxes \emph{without} cross-attention. This observation indicates that while gated self-attention that is performed before cross-attention may compromise textual grounding, cross-attention that processes the output of gated self-attention does not affect spatial grounding.

\input{figures/CA_reweight_figure}

\subsection{Network Rewiring: From Sequential to Parallel}
\label{subsec:network_rewiring}

Building on the analyses above, we propose a simple yet effective modification to the grounding mechanism, changing the relationship between gated self-attention and cross-attention from sequential to \emph{parallel}. This change eliminates the placement of gated self-attention before cross-attention, thus preventing the reduction of text grounding caused by gated self-attention. Moreover, in this parallel arrangement, the preservation of spatial grounding is assured, as gated self-attention for spatial grounding does not require subsequent cross-attention.

Specifically, recall that in GLIGEN~\cite{li2023gligen}, the output of gated self-attention is added to the residual from self-attention, which is then passed to the cross-attention module as follows:
\begin{equation}
\begin{split}
    F_{GSA} \leftarrow \underbrace{\text{GSA} (F_{SA}, \{ g_i \})}_{\textcolor{purple}{spatial\;grounding}} + F_{SA};\\
    F \leftarrow \underbrace{\text{CA} (F_{GSA}, c)}_{\textcolor{teal}{textual\;grounding}} + F_{GSA};
    \label{eq:combined}
\end{split}
\end{equation}

We propose to transform this sequence grounding pipeline into two parallel processes as follows:
\begin{align}
    F \leftarrow \underbrace{\text{GSA} (F_{SA}, \{ g_i \})}_{\textcolor{purple}{spatial\;grounding}} + \underbrace{\text{CA} (F_{SA}, c)}_{\textcolor{teal}{textual\;grounding}} + \underbrace{F_{SA}}_{residual};
\end{align}

Refer to Fig.~\ref{fig:pipeline_image} for the visualization of network architecture changes ((b) $\rightarrow$ (c)).
This \textit{network rewiring} is feasible because the input to gated self-attention remains unchanged, while the input to cross-attention shifts to $F_{SA}$, for which it was originally designed in the context of Latent Diffusion Models~\cite{rombach2022high}.

It is important to note that the modification is effective even when applied to the pretrained GLIGEN, which was trained with the sequential structure of the attention modules. Therefore, \textbf{our rewiring does not require any additional training or fine-tuning, introduces no extra parameters, and does not affect computation time or memory usage during the generation process.} The only requirement is the simple reconfiguration of the attention modules at inference time.

%% file: figures/CA_reweight_figure.tex
% -----------------------------------------------------------------------
\begin{figure*}[h!]
    \centering
    \footnotesize{
        \renewcommand{\arraystretch}{0.5}
        \setlength{\tabcolsep}{0.0em}
        \setlength{\fboxrule}{0.0pt}
        \setlength{\fboxsep}{0pt}
        
        \begin{tabularx}{\textwidth}{>{\centering\arraybackslash}m{0.16\textwidth} >{\centering\arraybackslash}m{0.16\textwidth} >{\centering\arraybackslash}m{0.166\textwidth} | >{\centering\arraybackslash}m{0.166\textwidth} >{\centering\arraybackslash}m{0.16\textwidth} >{\centering\arraybackslash}m{0.16\textwidth}}
        \scriptsize{Layout} & \scriptsize{w/ CA} & \scriptsize{w/o CA} & \scriptsize{Layout} & \scriptsize{w/ CA} & \scriptsize{w/o CA} \\
        \includegraphics[width=0.156\textwidth]{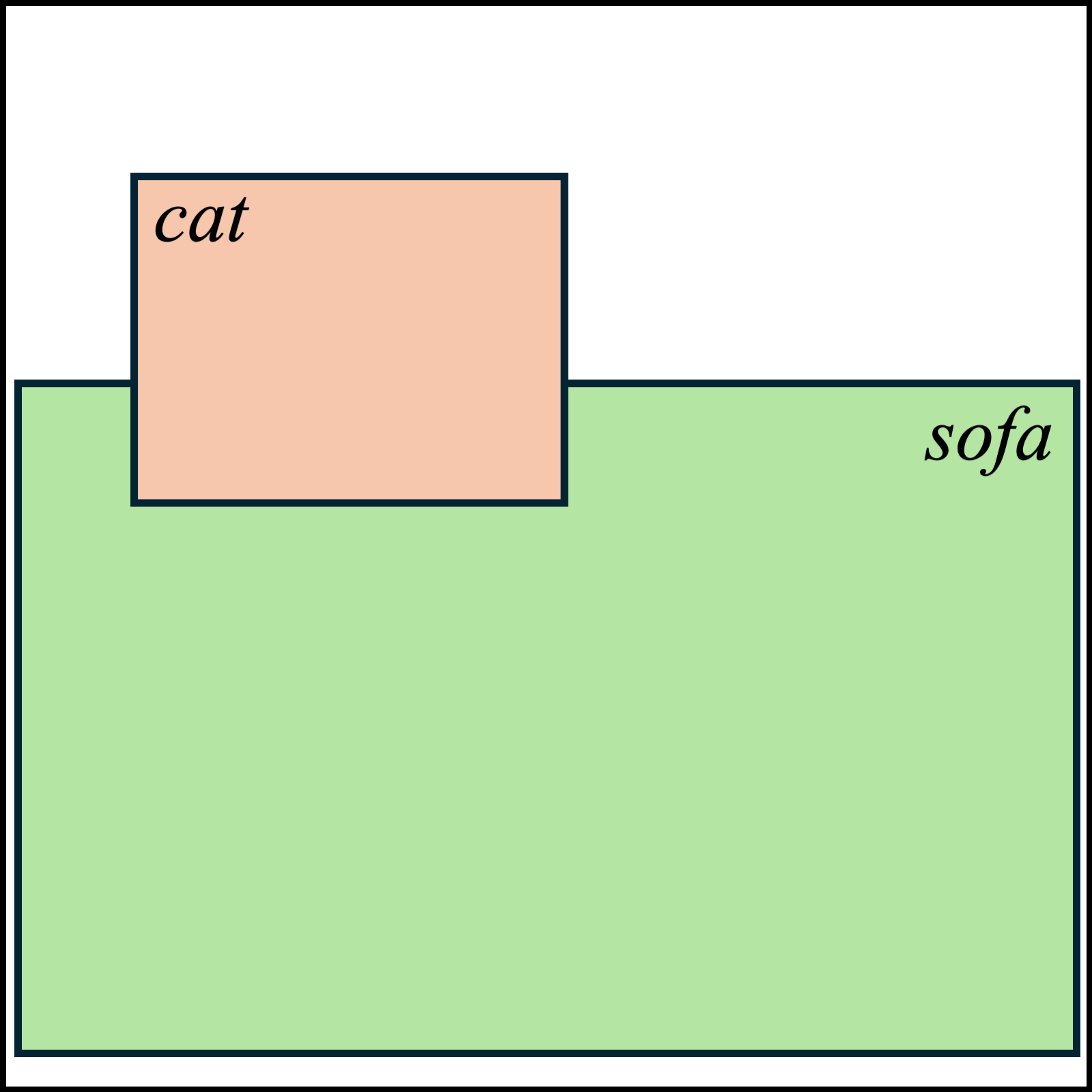} &
        \includegraphics[width=0.156\textwidth]{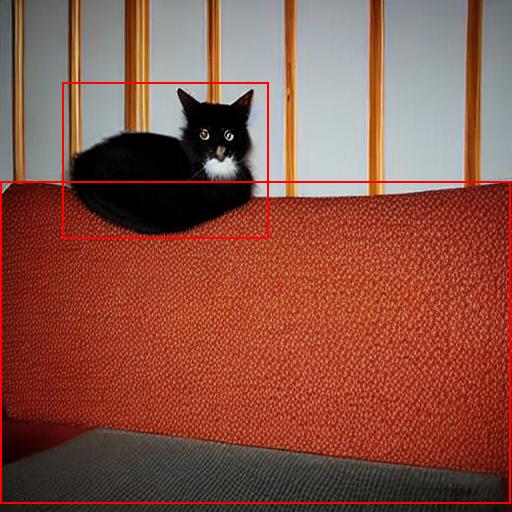} &
        \includegraphics[width=0.156\textwidth]{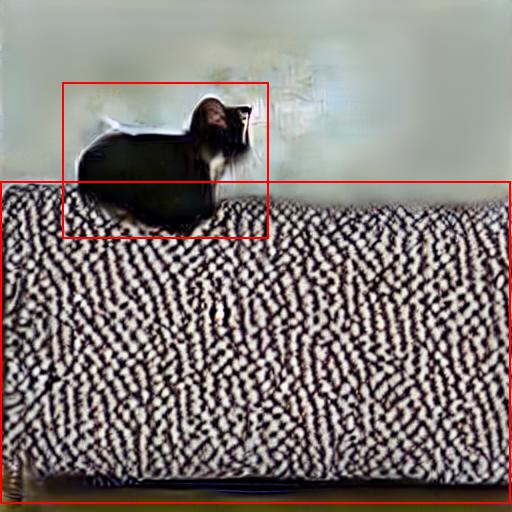} &
        \includegraphics[width=0.156\textwidth]{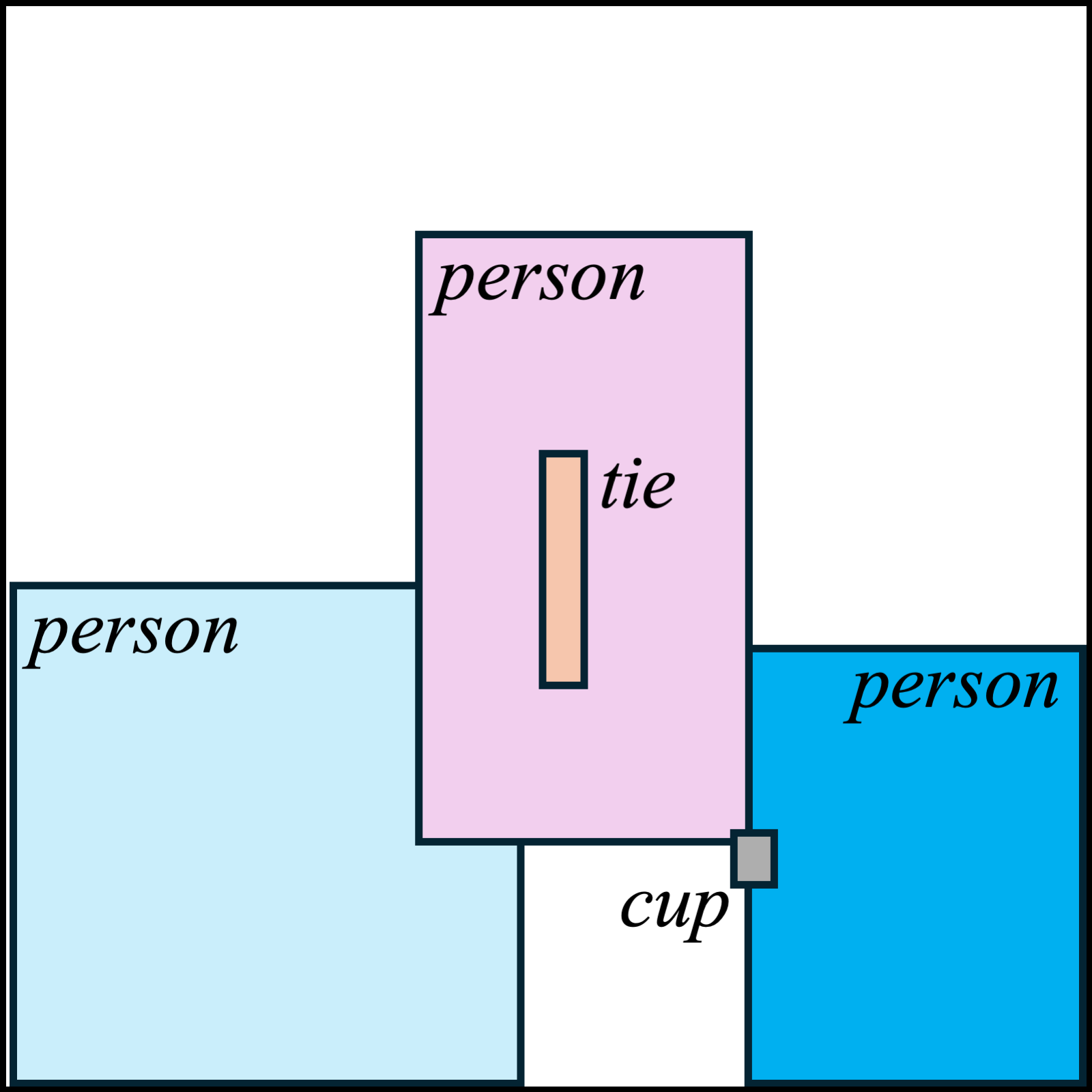} &
        \includegraphics[width=0.156\textwidth]{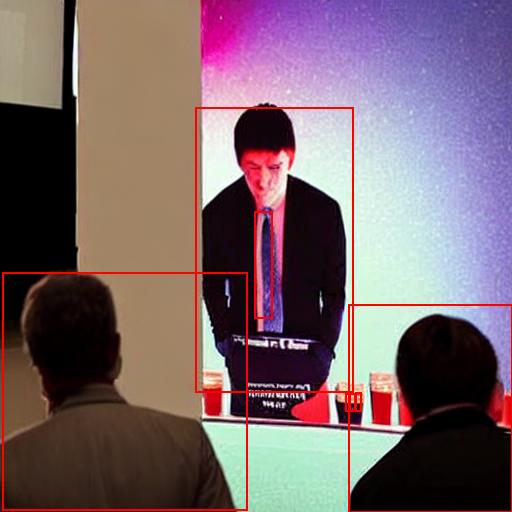} &
        \includegraphics[width=0.156\textwidth]{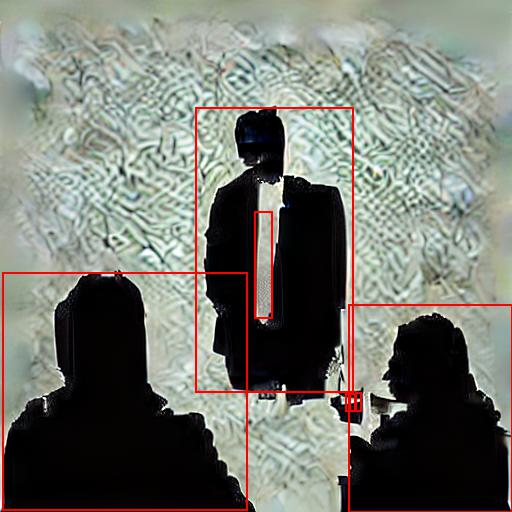}
        \end{tabularx}
    }
    \caption{Comparison of the output of GLIGEN~\cite{li2023gligen} with and without cross-attention. While the absence of cross-attention reduces realism and quality of the image, the silhouette of objects remains grounded within the given bounding boxes, as shown in the third column of each case.}
    \label{fig:ca_reweight_experiment}
\end{figure*}
% -----------------------------------------------------------------------

%% file: sections/05_experiments.tex
\section{Experiments}
\label{sec:experiments}
\input{figures/main_qualitative}
In this section, we show the effectiveness of our \methodname{} by evaluating the spatial grounding on existing layout-caption datasets~\cite{lin2014mscoco, feng2023layoutgpt} and the textual grounding on text-image alignment metrics~\cite{radford2021clip, kirstain2024pick}. We use the official GLIGEN~\cite{li2023gligen} checkpoint which is trained based on Stable Diffusion v1.4~\cite{rombach2022high}.

\subsection{Datasets}
\label{subsec:datasets}
\paragraph{\textbf{MS-COCO.}} 
We use the validations sets of both MS-COCO-2014 and MS-COCO-2017 datasets~\cite{lin2014mscoco}. Each dataset provides image-captions pairs and the $xy$-coordinates of bounding boxes along with their corresponding object categories.

\paragraph{\textbf{NSR-1K-GPT.}}
We also use the NSR-1K benchmark~\cite{feng2023layoutgpt} for evaluation. Based on each subset of NSR-1K---\emph{Counting} and \emph{Spatial}---we develop a new benchmark, \textit{NSR-1K-GPT}, augmenting each original caption in NSR-1K using GPT-4~\cite{openai2022chatgpt}. The instructions for augmentation are to (i) elaborate on the descriptions of each mentioned entity and (ii) provide additional details about the background of the image. More details on the evaluation datasets are provided in the \textbf{Appendix (Sec.~\ref{subsec:supp_eval_setup})}.

\subsection{Evaluation Metrics}
\label{subsec:eval_metrics}

\begin{itemize}
    \item \textbf{YOLO score:} Spatial grounding accuracy is assessed using YOLO score~\cite{wang2023yolov7}. We employ YOLOv7~\cite{wang2023yolov7} to detect objects in each generated image and compute the average precision (AP) based on the ground truth bounding box annotations from MS-COCO~\cite{lin2014mscoco}.
    \item \textbf{CLIP score:} Textual grounding accuracy is assessed using CLIP score~\cite{hessel2021clipscore}.
    \item \textbf{FID:} Image quality and diversity are evaluated using FID~\cite{heusel2018fid}.
    \item \textbf{User Study and PickScore}: We conduct a user study to assess human preferences for the generated images based on each input text prompt. Additionally, we use PickScore~\cite{kirstain2024pick}, a human preference predictor, to further analyze the results.
\end{itemize}

\subsection{Comparison with GLIGEN}
\label{subsec:gligen_comparison}

% main graph - camera ready
\begin{figure}[t!]
    \centering
    \includegraphics[width=\textwidth]{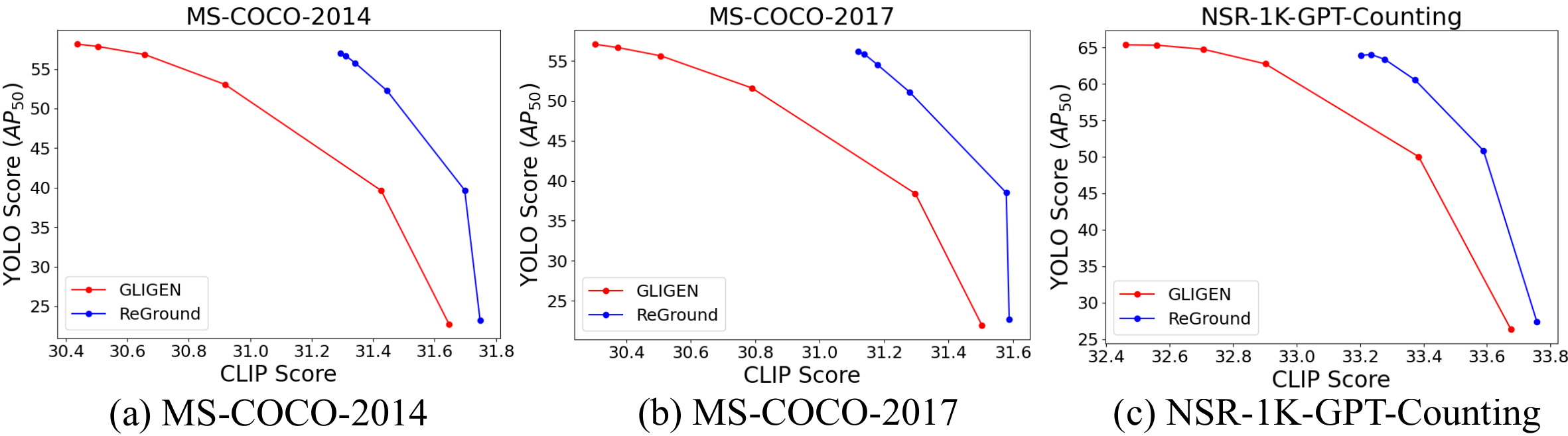}
    \caption{Comparisons on MS-COCO~\cite{lin2014mscoco} and NSR-1K-GPT. Each plot shows the relationship between textual grounding (\ie CLIP score~\cite{hessel2021clipscore}) and spatial grounding (\ie YOLO score~\cite{wang2023yolov7}) accuracy of GLIGEN~\cite{li2023gligen} and our method. Note that the plot of our \methodname{} is positioned in the top-right quadrant relative to GLIGEN, signifying that it alleviates the inherent trade-off between textual and spatial grounding.}
    \label{fig:graph_main_coco}
\end{figure}

\paragraph{\textbf{Textual-Spatial Grounding Trade-off.}}
We first examine the trade-off between textual and spatial groundings for both GLIGEN~\cite{li2023gligen} and our \methodname{}, the rewired version of GLIGEN, while varying the scheduled sampling parameter $\gamma$ from 1.0 to 0.1.

Fig.~\ref{fig:graph_main_coco}-(a), (b) present the graphs of CLIP score~\cite{hessel2021clipscore} and YOLO score~\cite{wang2023yolov7} measured on the MS-COCO datasets~\cite{lin2014mscoco}.
In MS-COCO-2014, when reducing $\gamma$ from 1.0 to 0.1, the CLIP score of GLIGEN varies from 30.44 to 31.65, while the YOLO score significantly drops from 58.13 to 22.75 (\textcolor{red}{red} in Fig.~\ref{fig:graph_main_coco}).
In contrast, our \methodname{}, (\textcolor{blue}{blue} in Fig.~\ref{fig:graph_main_coco}), demonstrates a notably superior trade-off between textual and spatial grounding.
Specifically, with $\gamma$ set to 1.0, \methodname{} already achieves a CLIP score of 31.29, accounting for 70.25\% of GLIGEN's total improvement in CLIP score when $\gamma$ is reduced from 1.0 to 0.1.
Despite this significant increase in CLIP score, the YOLO score remains largely unchanged, marking 56.96 which represents only a 3.31\% decrease in the range of YOLO score variation for GLIGEN when $\gamma$ is adjusted from 1.0 to 0.1.
Moreover, when varing the $\gamma$, the plot for \methodname{} (\textcolor{blue}{blue}) is constantly on the upper right side of GLIGEN (\textcolor{red}{red}), signifying a more advantageous trade-off across varying $\gamma$.
The same pattern is observed in MS-COCO-2017, where our \methodname{} achieves 68.33\% of the increase in CLIP score of GLIGEN while only compromising YOLO score by 2.62\% compared to the decrease for GLIGEN.

Fig.~\ref{fig:graph_main_coco}-(c) further shows a quantitative comparison on the \emph{Counting} subset of the newly generated NSR-1K-GPT benchmark. The plot reveals a consistent trend with the MS-COCO datasets. By reducing $\gamma$ from 1.0 to 0.1, GLIGEN's CLIP score is increased from 32.46 to 33.67, while the YOLO score is decreased from 65.36 to 26.38. In contrast, when $\gamma=1.0$, \methodname{} achieves a CLIP score of 33.20, which is equal to 61.16\% of GLIGEN's total improvement in CLIP score, while the compromise in YOLO score is equal to only 3.69\% of the total decrease in the YOLO score of GLIGEN from $\gamma=1.0$ to $\gamma=0.1$.
Moreover, a comparison on the \emph{Spatial} subset of NSR-1K-GPT is provided in the \textbf{Appendix (Sec.~\ref{subsec:supp_more_quantitative})}.
These results highlight that the advantage of our \methodname{} holds robustly for the realistic image captions provided in the MS-COCO~\cite{lin2014mscoco}, as well as for diverse text prompts generated by GPT-4~\cite{openai2022chatgpt}.

\begin{figure}[t!]
    \centering
    \includegraphics[width=0.9\textwidth]{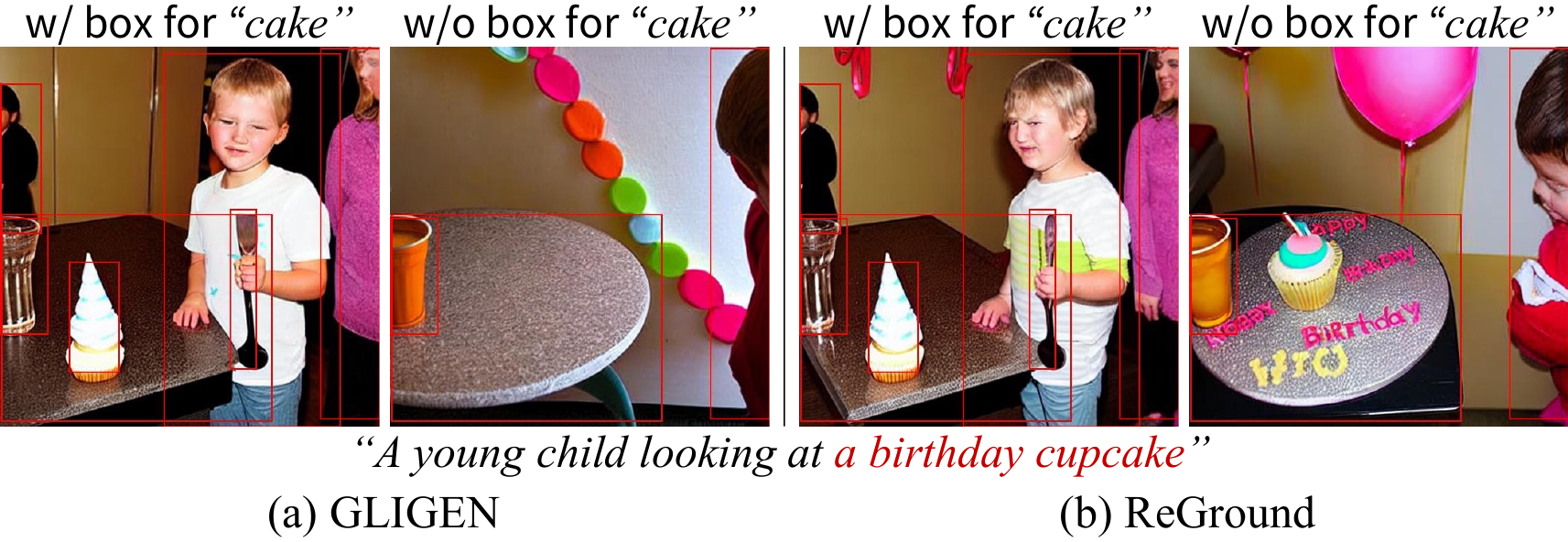}
    \caption{Generated images from the text prompt and bounding boxes from the MS-COCO-2017 (left of each column) and our COCO-Drop (right of each column). While GLIGEN~\cite{li2023gligen} fails to generate \textit{``\textcolor{BrickRed}{a birthday cupcake}''} when the corresponding bounding box is removed, our \methodname{} successfully generates a cupcake on the table.}
    \label{fig:box_drop_example}
\end{figure}

\paragraph{\textbf{Random Box Dropping.}}
To further assess the extent of description omission in each method, we modify the MS-COCO-2017 dataset~\cite{lin2014mscoco} to make the \textit{COCO-Drop} dataset.
In this version, the bounding boxes for 50\% of the categories are randomly removed from each image, thereby preventing every entity described in the text prompt from being included within the bounding boxes.

\begin{figure}[t!]
\centering
\begin{minipage}{.49\textwidth}
    \centering
    \includegraphics[width=0.9\linewidth]{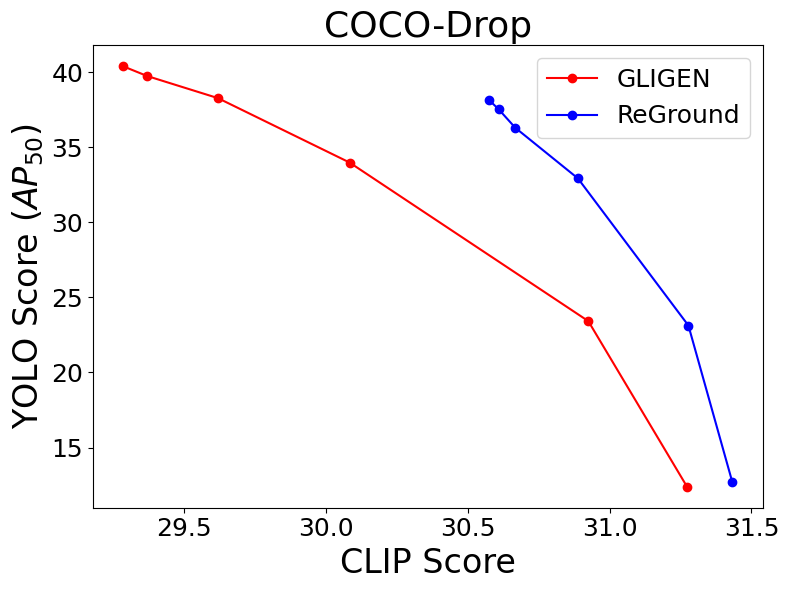}
    \caption{Comparison on the COCO-Drop dataset.}
    \label{fig:coco_drop_graph}
\end{minipage}\hfill% 
\begin{minipage}{.49\textwidth}
    \centering
    \includegraphics[width=0.9\linewidth]{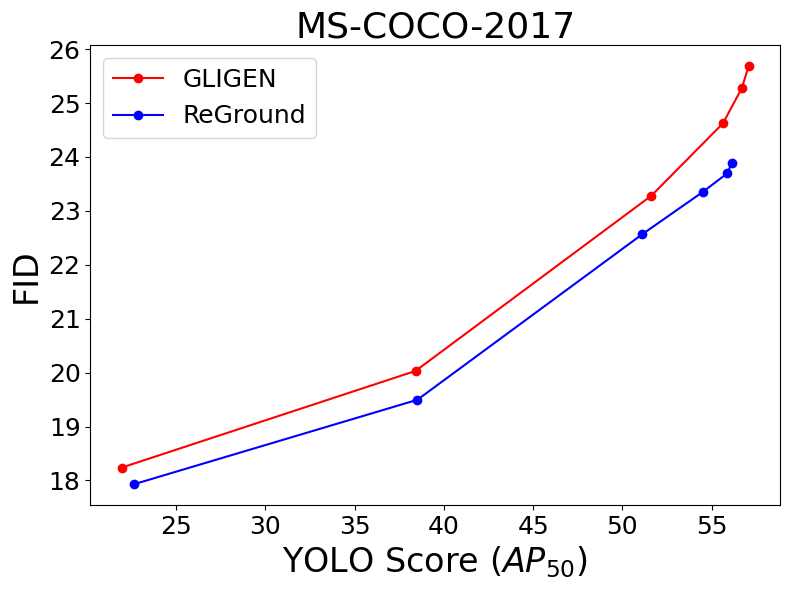}
    \caption{Comparison of FID~\cite{heusel2018fid} on MS-COCO-2017~\cite{lin2014mscoco} dataset.}
    \label{fig:fid_graph}
\end{minipage}
\end{figure}

Fig.~\ref{fig:coco_drop_graph} shows the quantitative comparison of \methodname{} and GLIGEN on COCO-Drop.
In this case, \methodname{} shows a larger advantage over GLIGEN in CLIP score, obtaining a gap in CLIP score which is 1.57 times that of the original MS-COCO-2017 dataset before box dropping for $\gamma=1.0$.
Such a larger gap in CLIP score demonstrates that compared to GLIGEN, our \methodname{} better reflects the text prompts even when some entities in the text prompt are not provided as a bounding box.
Fig.~\ref{fig:box_drop_example} displays a representative example, where GLIGEN fails to generate a \textit{``cupcake''} when its corresponding bounding box is removed in COCO-Drop, whereas our \methodname{} robustly generates the cupcake even when it it not provided as a bounding box.

\paragraph{\textbf{Image Quality.}}
Fig.~\ref{fig:fid_graph} displays the relationship between YOLO score~\cite{wang2023yolov7} and FID~\cite{heusel2018fid} for each method on MS-COCO-2017. Note that the FID of \methodname{} is constantly lower than that of GLIGEN~\cite{li2023gligen}, meaning that our network rewiring also results in higher image quality and diversity.

\paragraph{\textbf{User Study.}}
We conducted a user study to compare GLIGEN and our ReGround in terms of faithfulness to input text prompts. We used GPT-4~\cite{openai2022chatgpt} to generate 100 prompts each containing two different objects, along with a bounding box for each object. 
Participants were given the text prompt along with two images---one from each method---and asked to choose the image that \texttt{``better includes all the objects from the prompt.''}
Among the 92 out of 100 participants who passed the vigilance tests, our \methodname{} surpassed GLIGEN, with a preference rate of 70.05\% compared to 29.95\%. Further details on the user study are provided in the \textbf{Appendix (Sec.~\ref{subsec:supp_eval_setup})}.

\paragraph{\textbf{PickScore.}}
We further compare the PickScore~\cite{kirstain2024pick} of GLIGEN~\cite{li2023gligen} and our \methodname{} given each input text prompt. On MS-COCO-2017, \methodname{} is preferred over GLIGEN by 55.66\% to 44.34\%, and on COCO-Drop, \methodname{} is preferred by 57.57\% to 42.43\%.

\begin{figure}[h!]
    \centering
    \includegraphics[width=\textwidth]{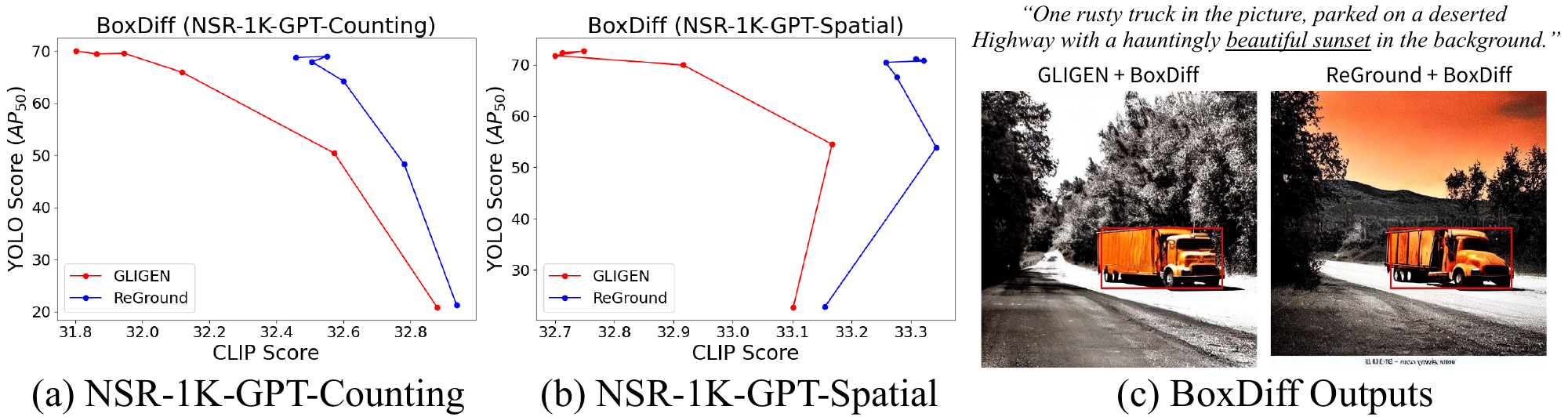}
    \caption{Comparison of applying BoxDiff~\cite{xie2023boxdiff} on GLIGEN~\cite{li2023gligen} and our \methodname{}, respectively. (a) and (b) show that our \methodname{} further improves the grounding quality of BoxDiff on NSR-1K-GPT datasets. (c) While BoxDiff with GLIGEN (left) also shows description omission---omitting \underline{\textit{``beautiful sunset''}} from the text prompt---BoxDiff with our \methodname{} contains the sunset in the final image (right).}
    \label{fig:boxdiff_results}
\end{figure}

\subsection{Impact of \methodname{} as a Backbone}
\label{subsec:plug_in_comparison}

We demonstrate that applying our rewiring of attention modules can also improve text-image alignment in other layout-guided generation methods that use GLIGEN as a backbone.
For instance, BoxDiff~\cite{xie2023boxdiff} is a notable example that uses GLIGEN as its foundation and improves spatial grounding with respect to the bounding boxes by leveraging cross-attention maps as additional spatial cues in a zero-shot manner.
Our network rewiring can also be combined with the zero-shot guidance of BoxDiff. Fig.~\ref{fig:boxdiff_results} illustrates the results on the NSR-1K-GPT datasets (a) when BoxDiff uses GLIGEN as the base, and (b) when it uses our \methodname{}, the rewired GLIGEN, as the base. It depicts that for the same range of spatial grounding accuracies, \methodname{} obtains noticeably higher textual grounding (\ie CLIP score~\cite{hessel2021clipscore}).
Also, as shown in Fig.~\ref{fig:boxdiff_results}-(c), our network rewiring allows for a more detailed description to accurately appear in the final image, both for the entities in the bounding boxes (\textit{``truck''}) and the entities that are given as a text prompt (\textit{``sunset''}).

%% file: figures/main_qualitative.tex
% -----------------------------------------------------------------------
\begin{figure*}[h!]
    \centering
    \footnotesize{
        \renewcommand{\arraystretch}{0.5}
        \setlength{\tabcolsep}{0.0em}
        \setlength{\fboxrule}{0.0pt}
        \setlength{\fboxsep}{0pt}
        
        \begin{tabularx}{\textwidth}{>{\centering\arraybackslash}m{0.198\textwidth} >{\centering\arraybackslash}m{0.198\textwidth} >{\centering\arraybackslash}m{0.198\textwidth} >{\centering\arraybackslash}m{0.198\textwidth} >{\centering\arraybackslash}m{0.198\textwidth}}
        Layout & SD & GLIGEN$_{\gamma=1.0}$ & GLIGEN$_{\gamma=0.2}$ & \textbf{ReGround} \\
        %%%%%
        \multicolumn{5}{c}{\includegraphics[width=0.99\textwidth]{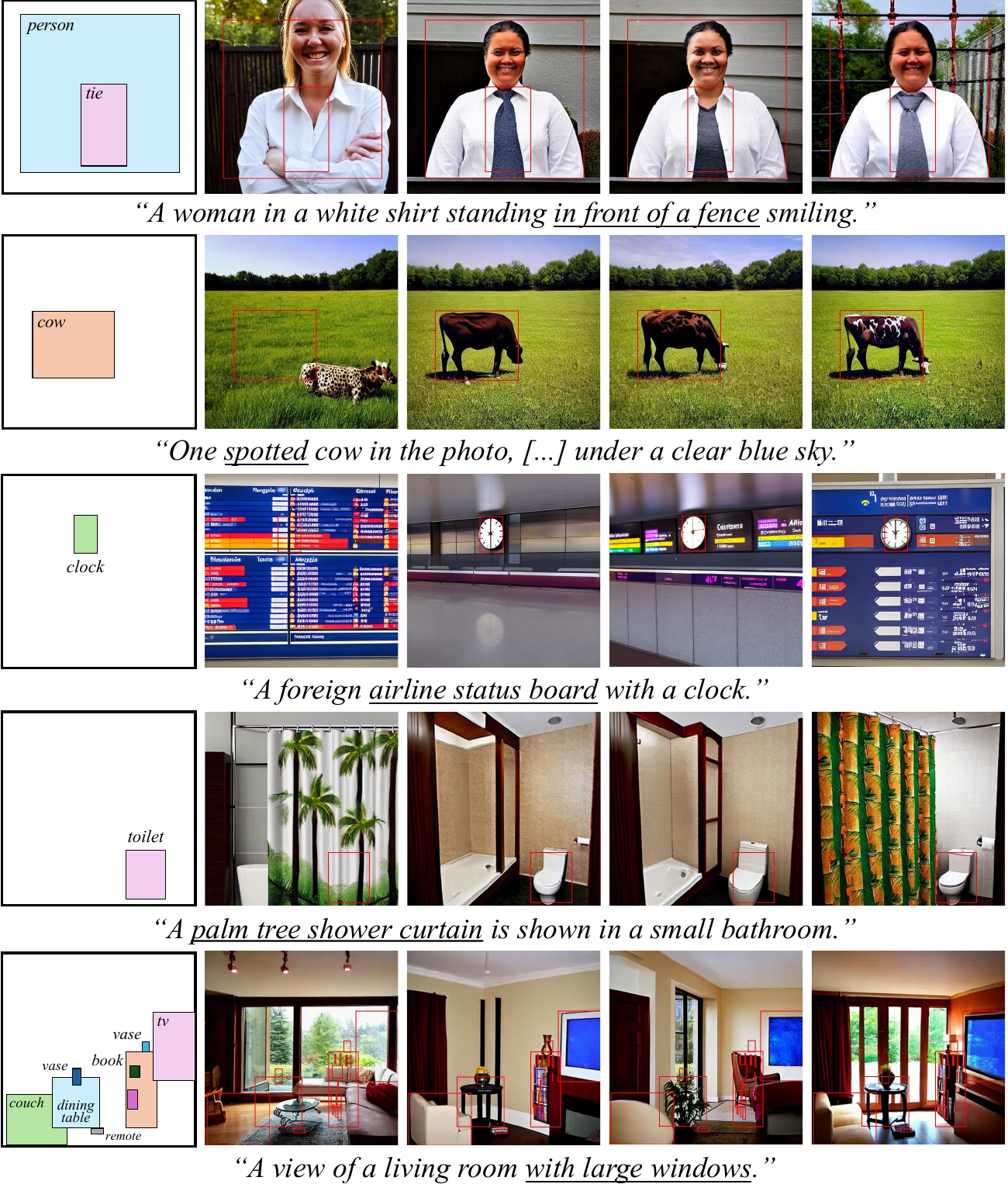}} \\
        \end{tabularx}
    }
    \caption{Qualitative comparisons. Stable Diffusion (SD, 2nd column) generates images that align with the given text descriptions, including the \underline{underlined} phrase in each row, but cannot take bounding boxes as input. GLIGEN (3rd column) creates images that match the input layouts but suffers from \textit{description omission}, failing to reflect the \underline{underlined} descriptions. Scheduled sampling strategy (4th column) can partially address this issue (for instance, in the 5th row, where \textit{``window''} appears in the room), but it results in a noticeable decline in spatial accuracy (as seen in the 1st row, where the tie is not generated). In contrast, our method (last column) accurately incorporates the \underline{underlined} text descriptions while maintaining precise spatial representation.} 
    \label{fig:main_qualitative}
\end{figure*}
% -----------------------------------------------------------------------

%% file: sections/06_conclusion.tex
\section{Conclusion}
\label{sec:conclusion}
We have demonstrated that a simple network rewiring of attention modules, making the gated self-attention and cross-attention parallel, surprisingly improves the trade-off between textual and spatial grounding at no additional cost --- without introducing any new parameters, any fine-tuning of the network, or any changes in generation time and memory. Using the pretrained GLIGEN~\cite{li2023gligen}, which was trained with the original sequential architecture of the two attention modules, the reconfiguration at inference time has led to achieving higher CLIP scores, indicating the noticeable improvement in textual grounding accuracy.
Moreover, our \methodname{} improves the textual grounding while preserving the spatial grounding accuracy -- achieving 70.25\% and 68.33\% of GLIGEN's total improvement with the scheduled sampling in CLIP score while compromising YOLO score only 3.31\% and 2.62\% for the MS-COCO-2014 and MS-COCO-2017 datasets, respectively.
We also showcased that this simple yet effective solution for the textual-spatial grounding trade-off can lead to improvements in diverse frameworks using GLIGEN as a base.

% \paragraph{\textbf{Supplementary.}}
% Due to limited space, we provide the following contents in the Supplementary: details on the evaluation setup (Sec. S1), additional quantitative (Sec. S2) and qualitative (Sec. S4) comparisons, and more results of \methodname{} as a backbone for other layout-guided generation methods (Sec. S3).
\paragraph{\textbf{Appendix.}}
Due to limited space, we provide the following contents in the Appendix: details on the evaluation setup (Sec.~\ref{subsec:supp_eval_setup}), additional quantitative (Sec.~\ref{subsec:supp_more_quantitative}) and qualitative (Sec.~\ref{subsec:supp_more_qualitative}) comparisons, and more results of \methodname{} as a backbone for other layout-guided generation methods (Sec.~\ref{subsec:supp_more_backbone}).

% Camera-ready: add acknowledgements
% \paragraph{Acknowledgements.}
\section*{Acknowledgments}
This work was supported by NRF grant (RS-2023-00209723), IITP grants (2022-0-00594, RS-2023-00227592, RS-2024-00399817), and Alchymist Project Program (RS-2024-00423625) funded by the Korean government (MSIT and MOTIE), and grants from the DRB-KAIST SketchTheFuture Research Center, NAVER-intel, Adobe Research, Hyundai NGV, KT, and Samsung Electronics.
% \section*{Acknowledgments}
% This work was supported by NRF grant (RS-2023-00209723), IITP grants (2022-0-00594, RS-2023-00227592, RS-2024-00399817), and KEIT grant (RS-2024-00423625) funded by the Korean government, and grants from the DRB, NAVER-intel, Adobe Research, Hyundai NGV, KT, Samsung Electronics.

%% file: sections_supp/01_dataset_details.tex
\subsection{Details on Evaluation Setup}
\label{subsec:supp_eval_setup}
This section provides further descriptions on the evaluation datasets (Sec.~\ref{subsec:datasets}) and the user study setup (Sec.~\ref{subsec:eval_metrics}).

\paragraph{\textbf{MS-COCO.}} 
The validation set of the MS-COCO-2017 dataset~\cite{lin2014mscoco} consists of 5,000 image-annotation pairs.
Since GLIGEN~\cite{li2023gligen} is trained to handle a maximum of 30 bounding boxes per image, we excluded pairs with more than 30 bounding boxes or no bounding boxes, resulting in a total of 4,952 images. For the validation set of the MS-COCO-2014 dataset~\cite{lin2014mscoco}, we randomly sampled 5,000 pairs for evaluation.

\paragraph{\textbf{NSR-1K-GPT.}}
\emph{Numerical and Spatial Reasoning (NSR-1K)}~\cite{feng2023layoutgpt} is a collection of layout-caption pairs designed to assess the numerical and spatial reasoning capabilities of image generation methods. The object labels and bounding boxes are from MS-COCO~\cite{lin2014mscoco}, while the captions are newly annotated based on the spatial relationships and numerical properties of objects.
NSR-1K consists of two subsets: \textit{Counting} and \textit{Spatial}.
We randomly sampled 1,000 pairs from the Counting set and used all 1,021 pairs from the Spatial set.

\begin{figure}[h!]
    \centering
    \includegraphics[width=0.50\textwidth]{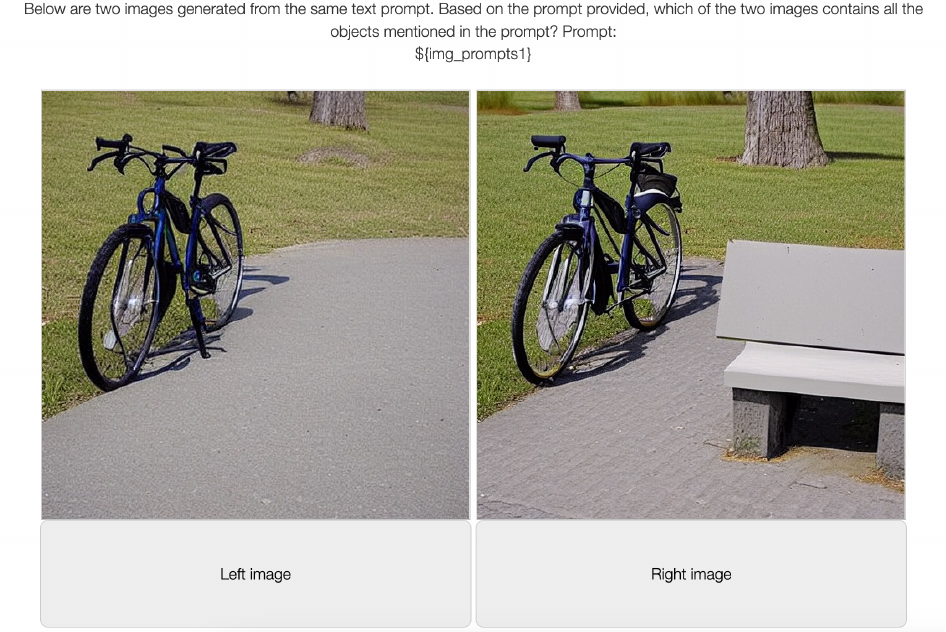}
    \caption{User study template. In the above example, the text prompt \textit{``A photo of a bicycle and a bench''} was displayed to the respondents.}
    \label{fig:supp_user_study}
\end{figure}

\paragraph{\textbf{User Study.}}
We conducted the user study through Amazon Mechanical Turk using the template displayed in Fig.~\ref{fig:supp_user_study}. Based on the text prompt and bounding boxes generated from GPT-4~\cite{openai2022chatgpt}, images were generated by both GLIGEN~\cite{li2023gligen} and our \methodname{}. 
Since \methodname{} aims to resolve the failure cases of GLIGEN, we re-generated both images when the differences between them were minimal (\ie, if the LPIPS value~\cite{zhang2018unreasonable} was less than 0.3), resulting in an average of 2.4 iterations per image. Each participant answered 20 questions and 5 vigilance tests.

%% file: sections_supp/02_additional_quantitative.tex
\subsection{Additional Quantitative Comparisons}
\label{subsec:supp_more_quantitative}

In addition to Sec.~\ref{subsec:gligen_comparison}, this section provides quantitative comparisons between GLIGEN~\cite{li2023gligen} and our \methodname{} on the Spatial subset of NSR-1K-GPT, and with a different version of Stable Diffusion~\cite{rombach2022high} as the base image diffusion model. 

\begin{figure}[b!]
    \centering
    \includegraphics[width=0.95\textwidth]{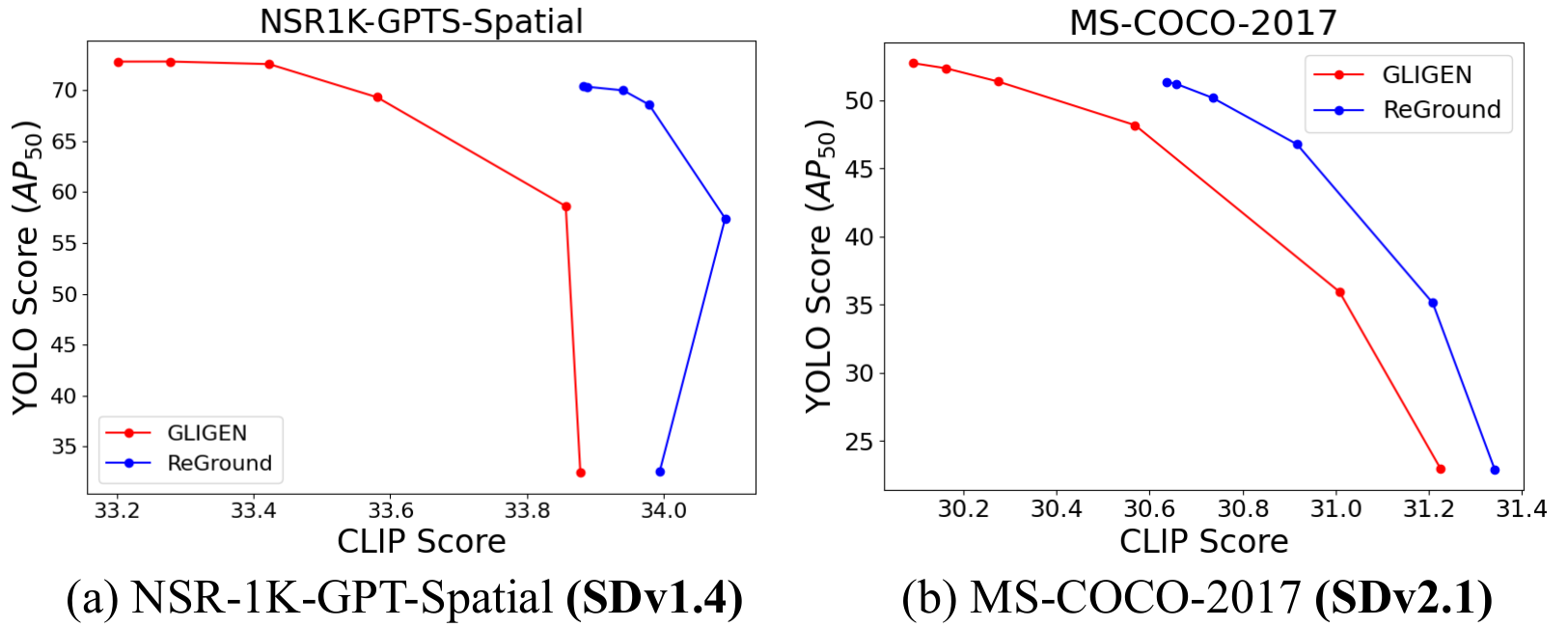}
    \caption{Quantitative comparisons (a) on the \emph{Spatial} subset of NSR-1K-GPT and (b) using SDv2.1 as the base image diffusion model. Consistent with the findings from Fig. 6 of the main paper, our \methodname{} demonstrates improved performance in textual and spatial groundings, as seen by the higher CLIP score~\cite{hessel2021clipscore} for the same range of YOLO score~\cite{wang2023yolov7}.}
    \label{fig:supp_main_graph}
\end{figure}

\paragraph{\textbf{Comparison on NSR-1K-GPT-Spatial.}}
Fig.~\ref{fig:supp_main_graph}-(a) shows the CLIP score~\cite{hessel2021clipscore} and YOLO score~\cite{wang2023yolov7} measured on the \emph{Spatial} subset of NSR-1K-GPT.
The minimum CLIP score of our \methodname{} (33.89 at $\gamma=1.0$) is already higher than GLIGEN's maximum CLIP score (33.88 at $\gamma=0.1$), indicating that \methodname{} obtains a significant enhancement in textual grounding while preserving the spatial grounding.

% supp graph - camera ready

\paragraph{\textbf{Results with SDv2.1 as Base Diffusion Model.}}
In Sec.~\ref{sec:experiments}, we conducted experiments using the GLIGEN~\cite{li2023gligen} checkpoint based on Stable Diffusion v1.4 \textbf{(SDv1.4)}. Additionally, we provide quantitative comparisons with an unofficial GLIGEN checkpoint~\cite{lian2023llmgrounded} that was trained with \textbf{SDv2.1} as the base image diffusion model. The results, presented in Fig.~\ref{fig:supp_main_graph}-(b), clearly demonstrate the significant outperformance of our \methodname{} over GLIGEN.

%% file: sections_supp/03_more_backbones.tex
\subsection{More Results with \methodname{} as Backbone}
\label{subsec:supp_more_backbone}

\begin{figure*}[h!]
    \centering
    \scriptsize{
        \renewcommand{\arraystretch}{0.5}
        \setlength{\tabcolsep}{0.0em}
        \setlength{\fboxrule}{0.0pt}
        \setlength{\fboxsep}{0pt}
        
        \begin{tabularx}{\textwidth}{>{\centering\arraybackslash}m{0.20\textwidth} >{\centering\arraybackslash}m{0.20\textwidth} >{\centering\arraybackslash}m{0.20\textwidth} >{\centering\arraybackslash}m{0.20\textwidth} >{\centering\arraybackslash}m{0.20\textwidth}}
        Layout & BoxDiff w/ GLIGEN & BoxDiff w/ \textbf{\methodname{}} & Attn-Refocus w/ GLIGEN & Attn-Refocus w/ \textbf{\methodname{}} \\
        %%%%%
        \multicolumn{5}{c}{\includegraphics[width=\textwidth]{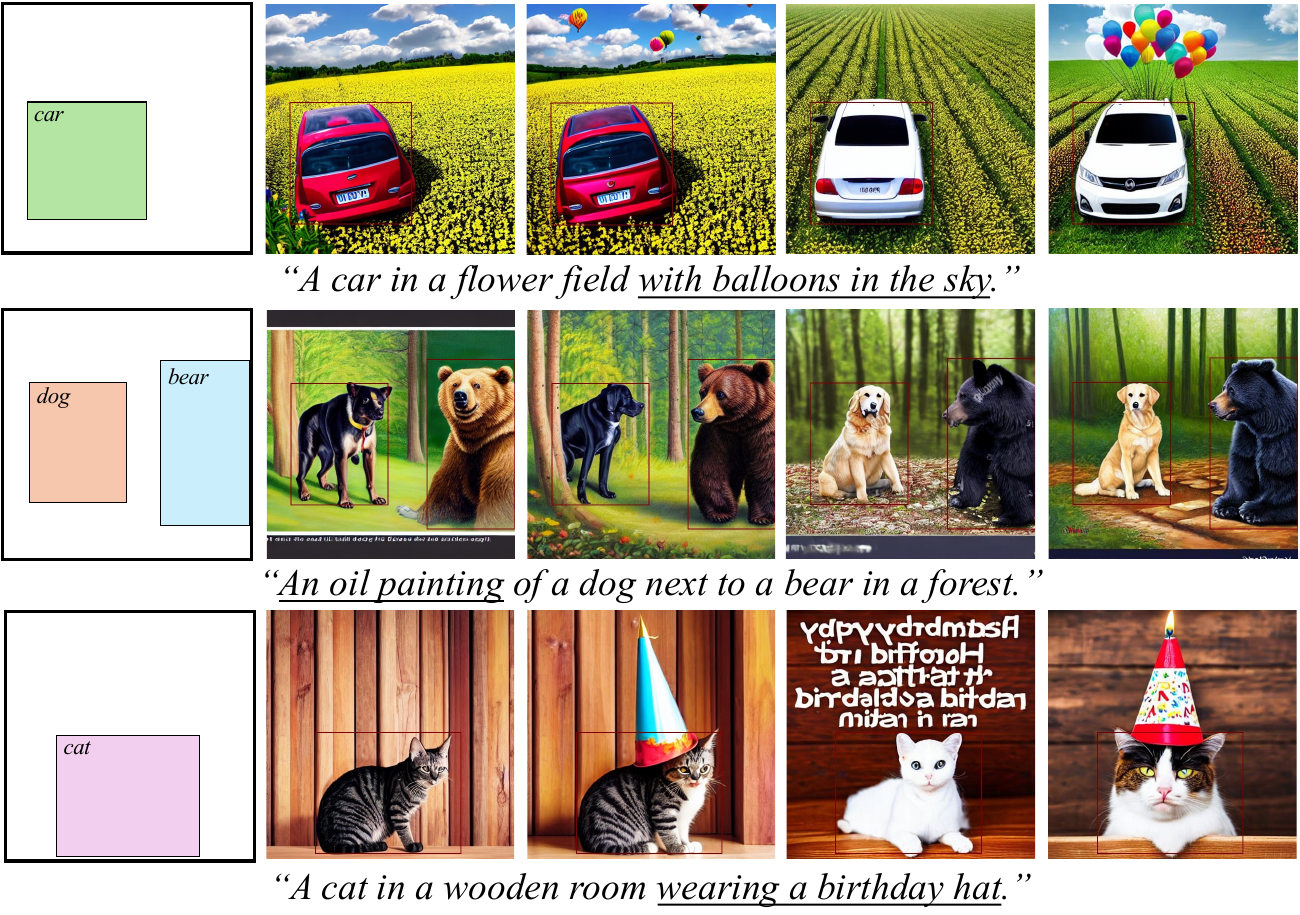}} \\
        \end{tabularx}
    }
    \caption{Comparisons of GLIGEN~\cite{li2023gligen} and our \methodname{} as a backbone for BoxDiff~\cite{xie2023boxdiff} and Attention Refocusing (Attn-Refocus)~\cite{phung2023grounded}.} 
    \label{fig:supp_backbone_1}
\end{figure*}

\begin{figure*}[h!]
    \centering
    \scriptsize{
        \renewcommand{\arraystretch}{0.5}
        \setlength{\tabcolsep}{0.0em}
        \setlength{\fboxrule}{0.0pt}
        \setlength{\fboxsep}{0pt}
        
        \begin{tabularx}{\textwidth}{>{\centering\arraybackslash}m{0.20\textwidth} >{\centering\arraybackslash}m{0.20\textwidth} >{\centering\arraybackslash}m{0.20\textwidth} >{\centering\arraybackslash}m{0.20\textwidth} >{\centering\arraybackslash}m{0.20\textwidth}}
        Layout & BoxDiff w/ GLIGEN & BoxDiff w/ \textbf{\methodname{}} & Attn-Refocus w/ GLIGEN & Attn-Refocus w/ \textbf{\methodname{}} \\
        %%%%%
        \multicolumn{5}{c}{\includegraphics[width=\textwidth]{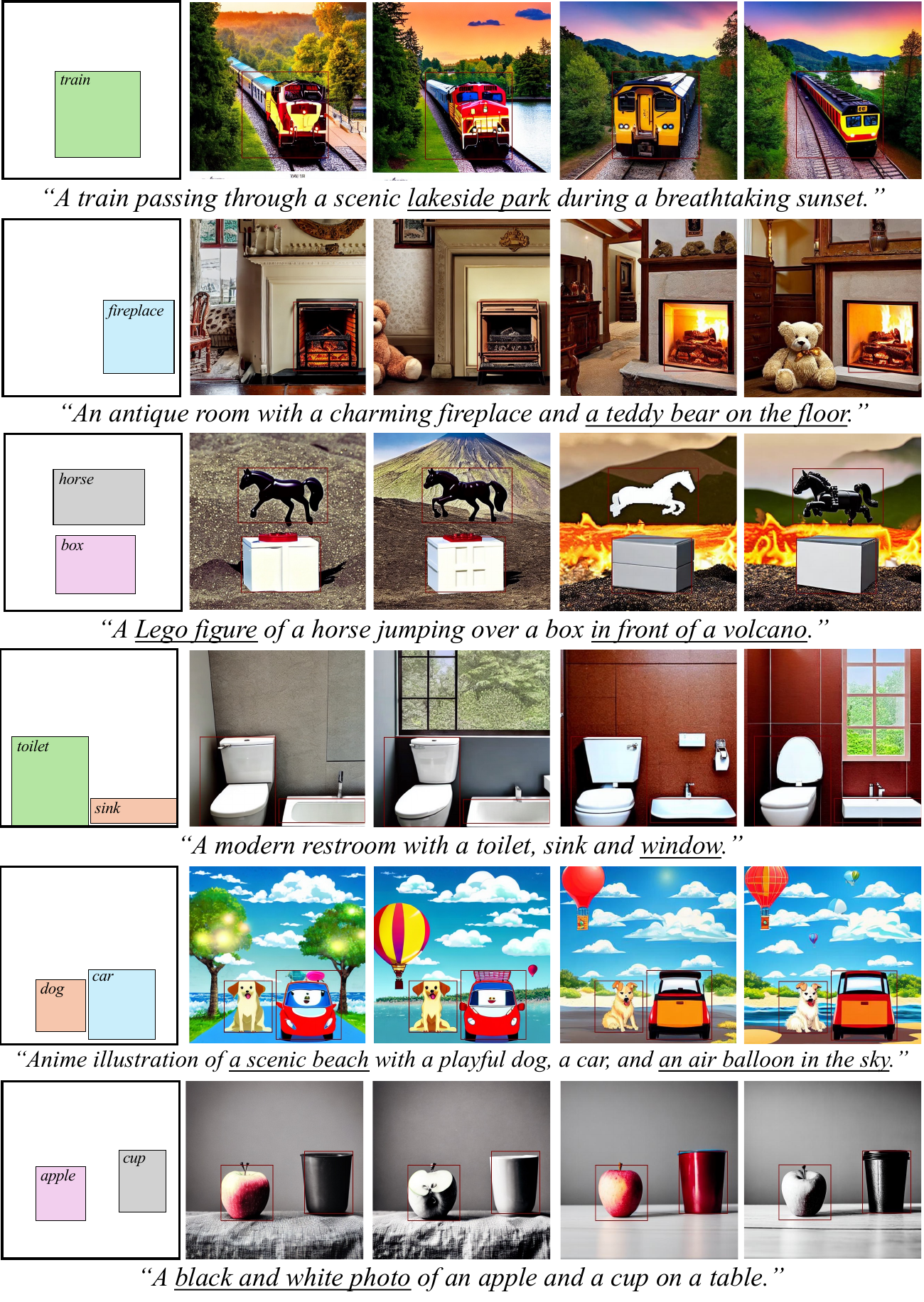}} \\
        \end{tabularx}
    }
    \caption{More comparisons on BoxDiff~\cite{xie2023boxdiff} and Attention Refocusing (Attn-Refocus)~\cite{phung2023grounded}.}
    \label{fig:supp_backbone_2}
\end{figure*}

In addition to Sec.~\ref{subsec:plug_in_comparison}, we provide qualitative comparisons of different layout-guided generation methods using GLIGEN~\cite{li2023gligen} and our \methodname{} as backbones, respectively (Fig.~\ref{fig:supp_backbone_1}, \ref{fig:supp_backbone_2}). 
The results on BoxDiff~\cite{xie2023boxdiff} and Attention Refocusing~\cite{phung2023grounded} illustrate that our network rewiring substantially improves the performance of layout-guided generation methods built upon the GLIGEN framework.

% --- full comparison - 1
% -----------------------------------------------------------------------

% --- full comparison - 2
% -----------------------------------------------------------------------

%% file: sections_supp/04_additional_qualitative.tex
\clearpage
\subsection{Additional Qualitative Comparisons}
\label{subsec:supp_more_qualitative}

In this section, we provide extensive qualitative comparisons of Stable Diffusion (SD)~\cite{rombach2022high}, GLIGEN~\cite{li2023gligen}, and our \methodname{} on layout-guided image generation. Note that $\gamma \in [0, 1]$ denotes the fraction of the initial denoising steps during which gated self-attention is activated, as discussed in Sec.~\ref{subsec:grounding_trade-off}.

In each row, the input layout is presented in the first column, with the input text prompt displayed below the images. The phrase \underline{underlined} in each prompt highlights the entity subject to \textit{description omission}, as mentioned in Sec.~\ref{subsec:prompt_neglection}. Furthermore, black arrows are used to denote bounding boxes that some methods fail to represent accurately, whereas other methods succeed in doing so precisely. Red arrows signify a failure in either spatial or textual grounding, while green arrows indicate successful grounding of a specific entity.

\input{sections_supp/supp_figures/additional_qualitative_NEW}

%% file: sections_supp/supp_figures/additional_qualitative_NEW.tex
% -----------------------------------------------------------------------
\begin{figure*}[h!]
    \centering
    \footnotesize{
        \renewcommand{\arraystretch}{0.5}
        \setlength{\tabcolsep}{0.0em}
        \setlength{\fboxrule}{0.0pt}
        \setlength{\fboxsep}{0pt}
        %%%%%%%%%%%%%
        \begin{tabularx}{\textwidth}{>{\centering\arraybackslash}m{0.20\textwidth} >{\centering\arraybackslash}m{0.20\textwidth} >{\centering\arraybackslash}m{0.20\textwidth} >{\centering\arraybackslash}m{0.20\textwidth} >{\centering\arraybackslash}m{0.20\textwidth}}
        Layout & SD & GLIGEN$_{\gamma=1.0}$ & GLIGEN$_{\gamma=0.2}$ & \textbf{ReGround} \\
        %%%%%%%%%%%%%%%%%%%%%%%%%
        \multicolumn{5}{c}{\includegraphics[width=\textwidth]{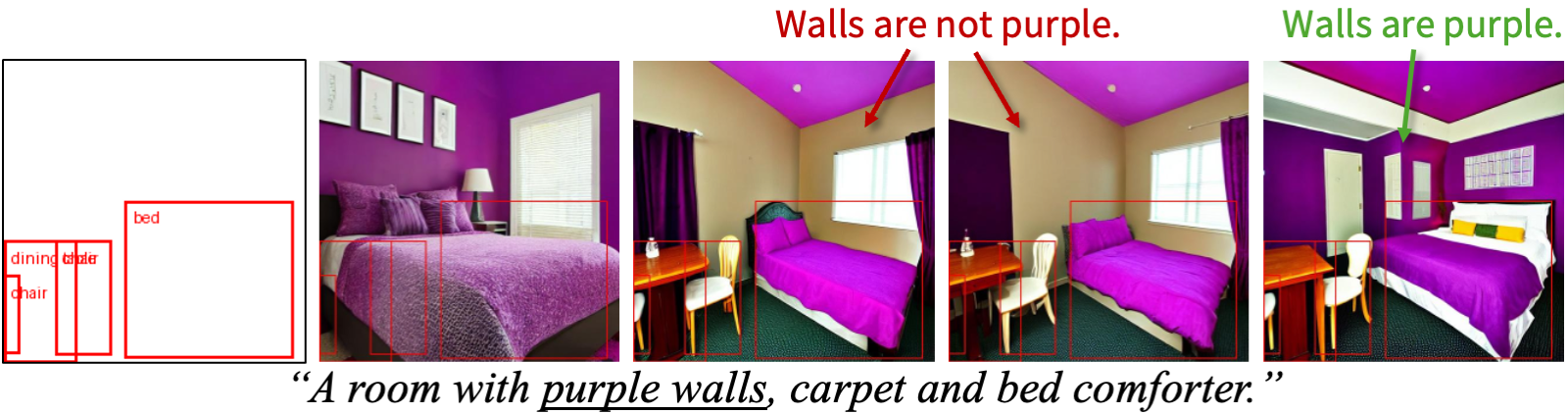}} \\
        \midrule
        %%%%%%%%%%%%%
        \multicolumn{5}{c}{\includegraphics[width=\textwidth]{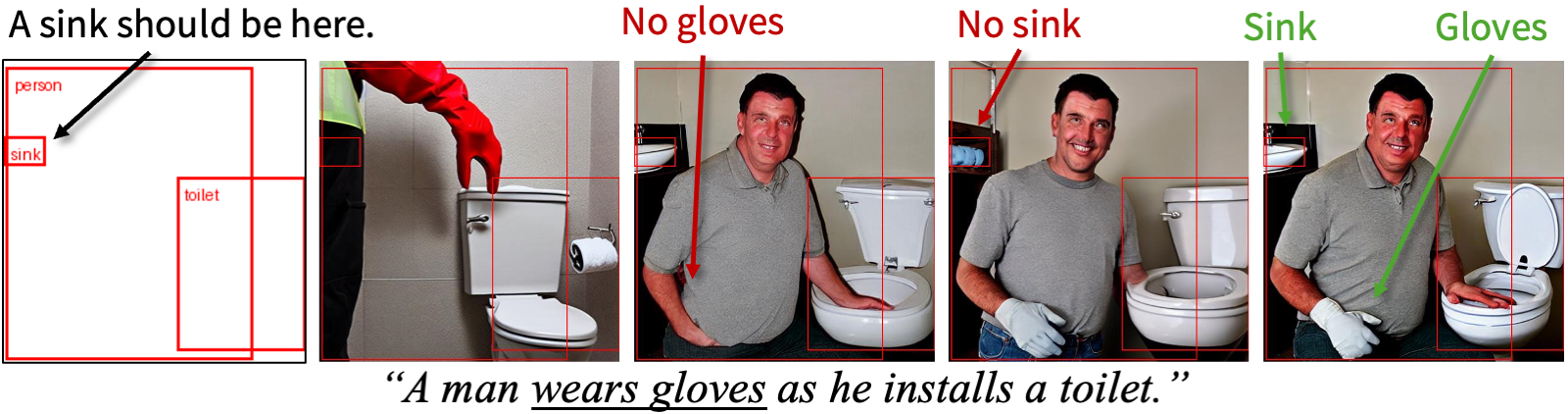}} \\
        \midrule
        % %%%%%%%%%%%%%
        \multicolumn{5}{c}{\includegraphics[width=\textwidth]{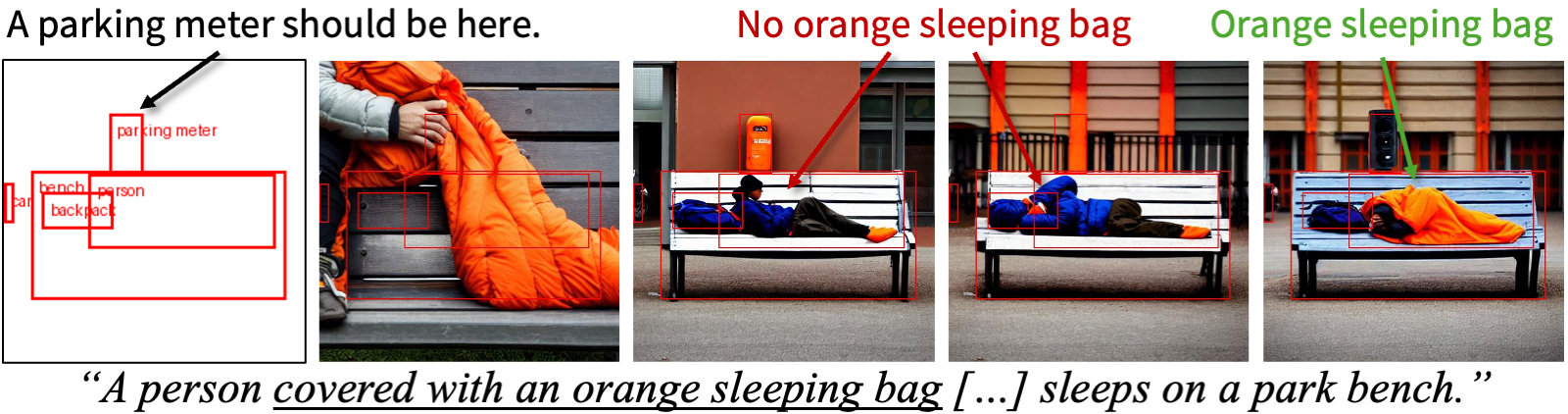}} \\
        \end{tabularx}
    }
    \label{fig:supp_more_qualitative_01}
\end{figure*}
% -----------------------------------------------------------------------

% -----------------------------------------------------------------------
\begin{figure*}[h!]
    \centering
    \footnotesize{
        \renewcommand{\arraystretch}{0.5}
        \setlength{\tabcolsep}{0.0em}
        \setlength{\fboxrule}{0.0pt}
        \setlength{\fboxsep}{0pt}
        %%%%%%%%%%%%%
        \begin{tabularx}{\textwidth}{>{\centering\arraybackslash}m{0.20\textwidth} >{\centering\arraybackslash}m{0.20\textwidth} >{\centering\arraybackslash}m{0.20\textwidth} >{\centering\arraybackslash}m{0.20\textwidth} >{\centering\arraybackslash}m{0.20\textwidth}}
        Layout & SD & GLIGEN$_{\gamma=1.0}$ & GLIGEN$_{\gamma=0.2}$ & \textbf{ReGround} \\
        %%%%%%%%%%%%%%%%%%%%%%%%%
        \multicolumn{5}{c}{\includegraphics[width=\textwidth]{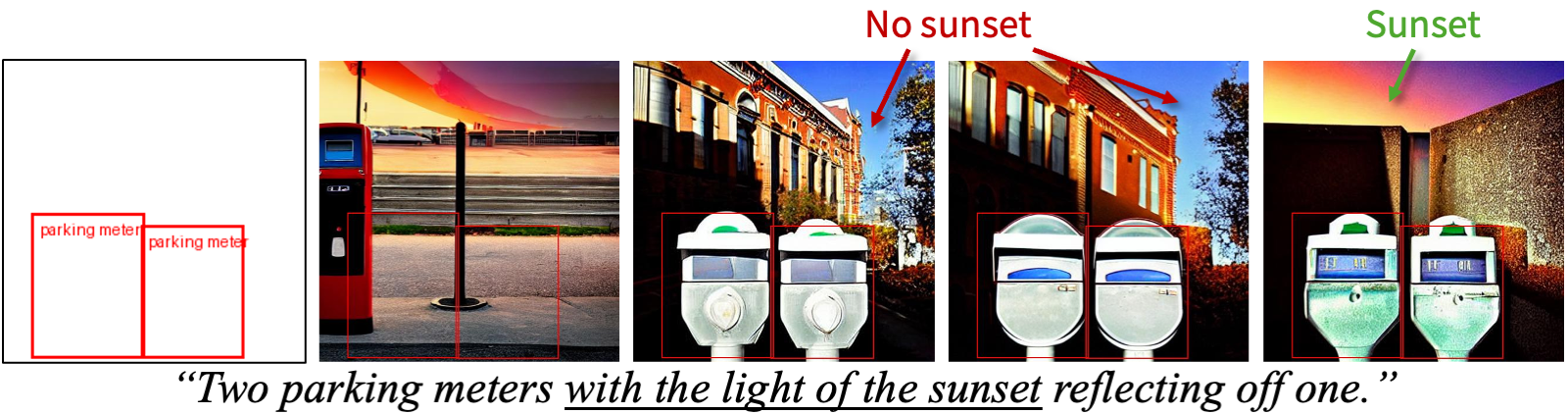}} \\
        \midrule
        %%%%%%%%%%%%%
        \multicolumn{5}{c}{\includegraphics[width=\textwidth]{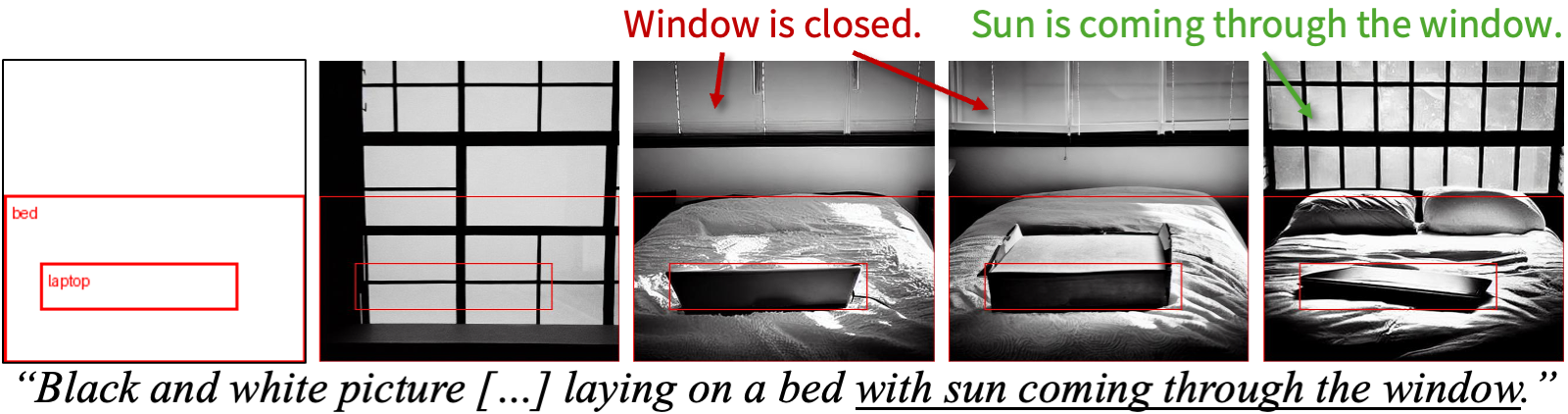}} \\
        \midrule
        %%%%%%%%%%%%%
        \multicolumn{5}{c}{\includegraphics[width=\textwidth]{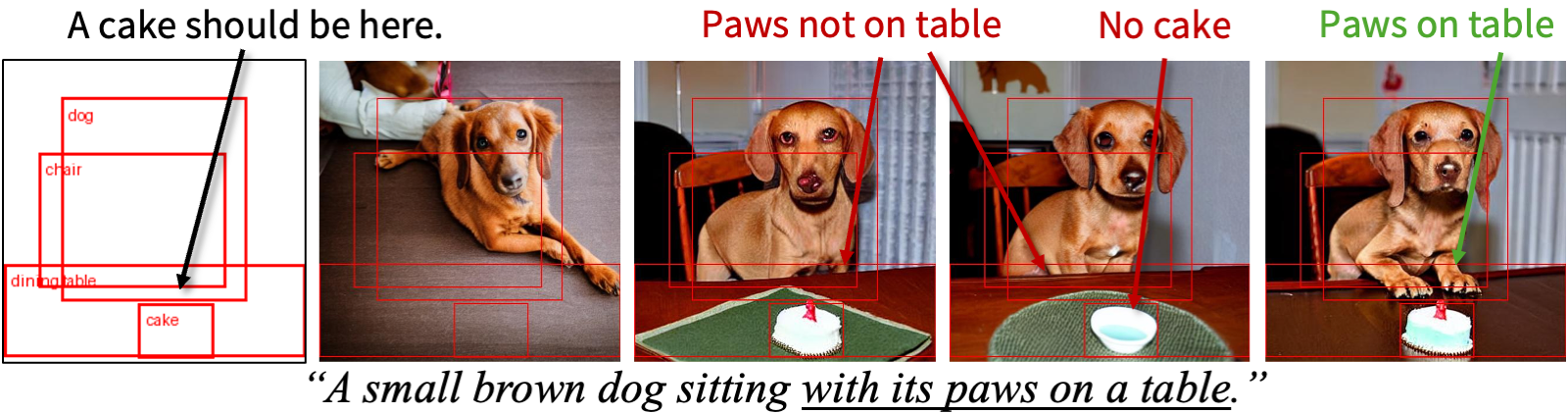}} \\
        \midrule
        %%%%%%%%%%%%%
        \multicolumn{5}{c}{\includegraphics[width=\textwidth]{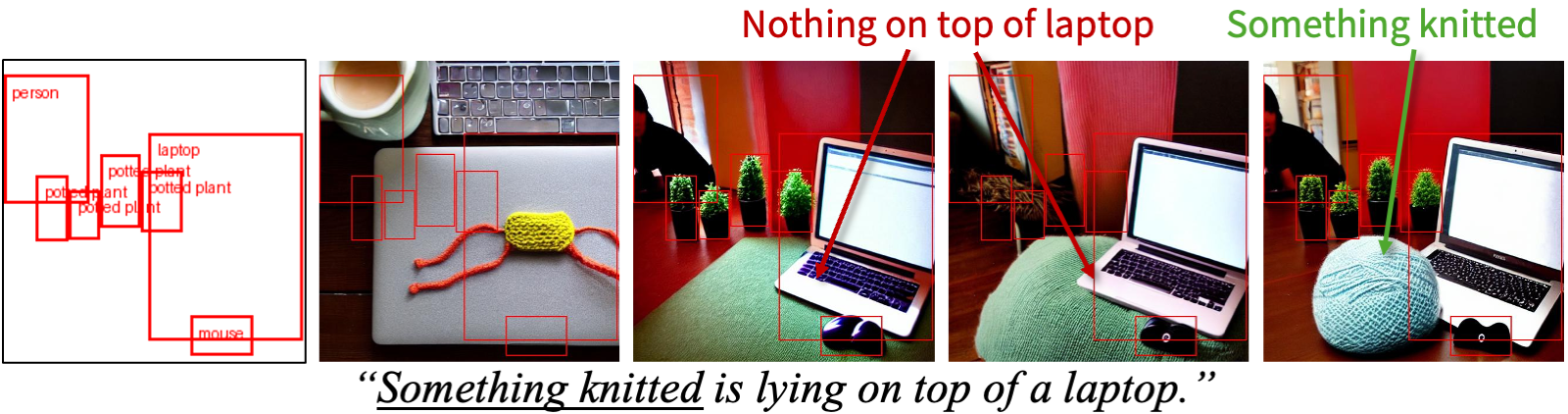}} \\
        \midrule
        %%%%%%%%%%%%%
        \multicolumn{5}{c}{\includegraphics[width=\textwidth]{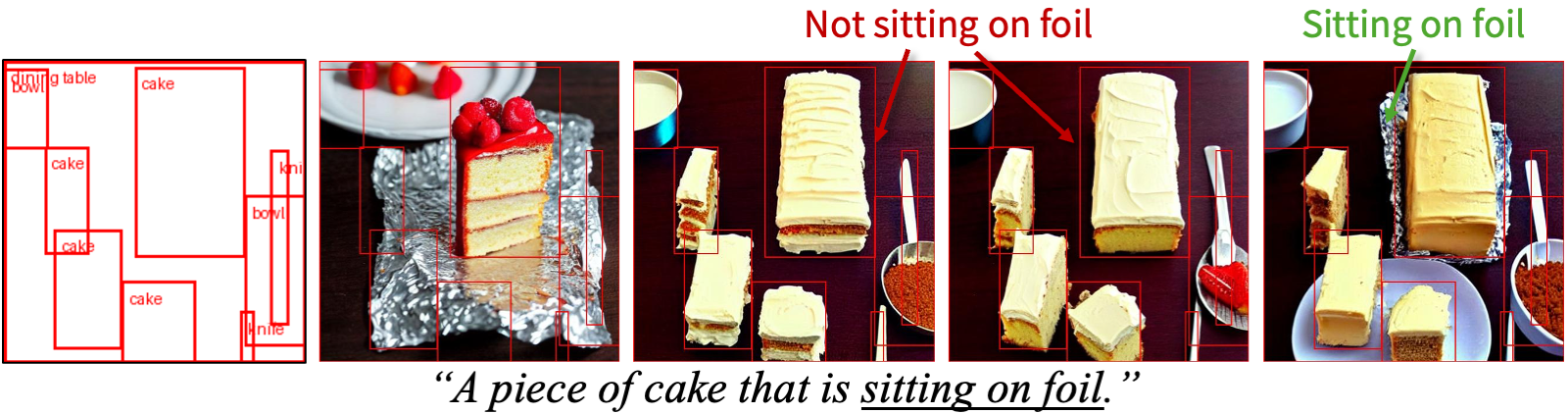}} \\
        %%%%%%%%%%%%%
        \end{tabularx}
    }
    \label{fig:supp_more_qualitative_02}
\end{figure*}
% -----------------------------------------------------------------------

% -----------------------------------------------------------------------
\begin{figure*}[h!]
    \centering
    \footnotesize{
        \renewcommand{\arraystretch}{0.5}
        \setlength{\tabcolsep}{0.0em}
        \setlength{\fboxrule}{0.0pt}
        \setlength{\fboxsep}{0pt}
        %%%%%%%%%%%%%
        \begin{tabularx}{\textwidth}{>{\centering\arraybackslash}m{0.20\textwidth} >{\centering\arraybackslash}m{0.20\textwidth} >{\centering\arraybackslash}m{0.20\textwidth} >{\centering\arraybackslash}m{0.20\textwidth} >{\centering\arraybackslash}m{0.20\textwidth}}
        Layout & SD & GLIGEN$_{\gamma=1.0}$ & GLIGEN$_{\gamma=0.2}$ & \textbf{ReGround} \\
        %%%%%%%%%%%%%%%%%%%%%%%%%
        \multicolumn{5}{c}{\includegraphics[width=\textwidth]{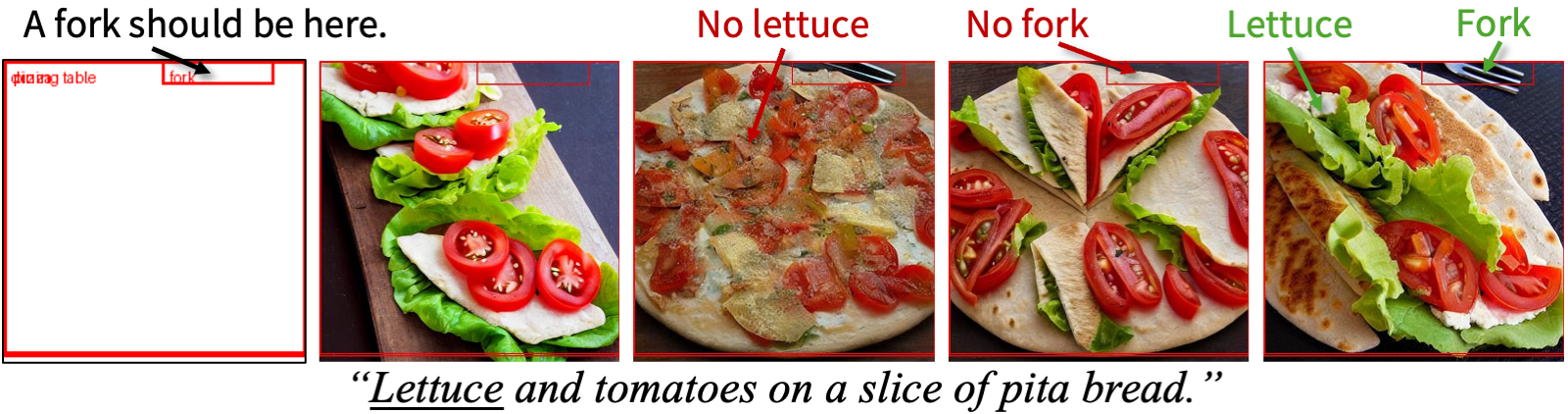}} \\
        \midrule
        %%%%%%%%%%%%%
        \multicolumn{5}{c}{\includegraphics[width=\textwidth]{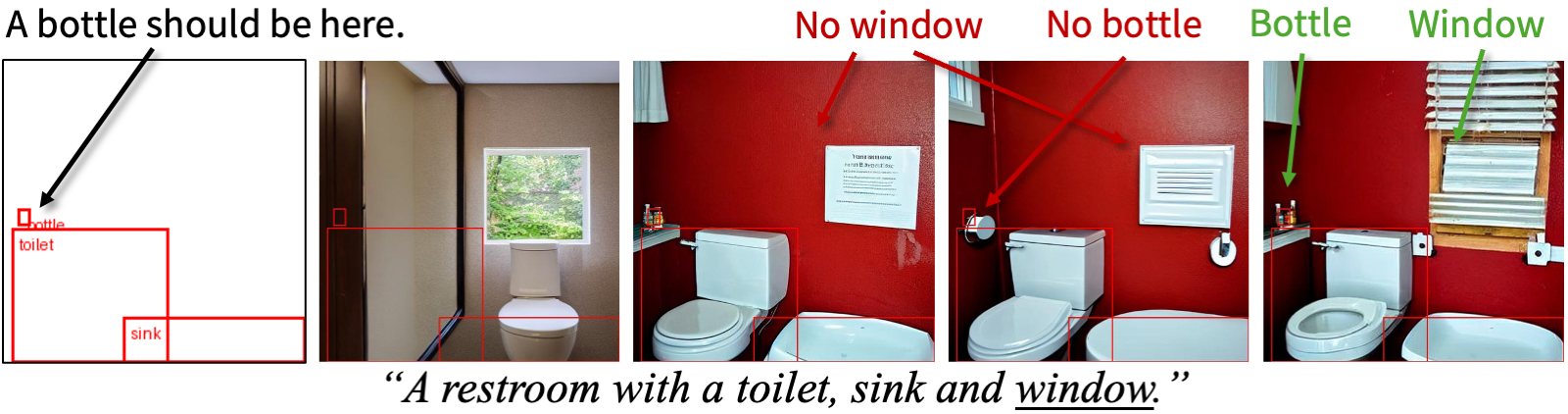}} \\
        \midrule
        %%%%%%%%%%%%%
        \multicolumn{5}{c}{\includegraphics[width=\textwidth]{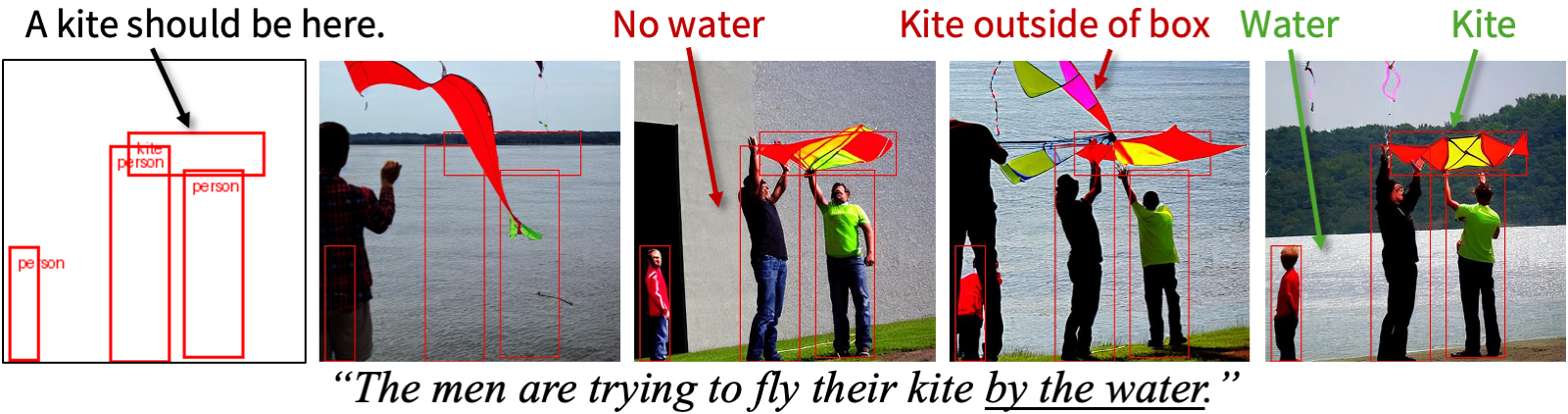}} \\
        \midrule
        %%%%%%%%%%%%%
        \multicolumn{5}{c}{\includegraphics[width=\textwidth]{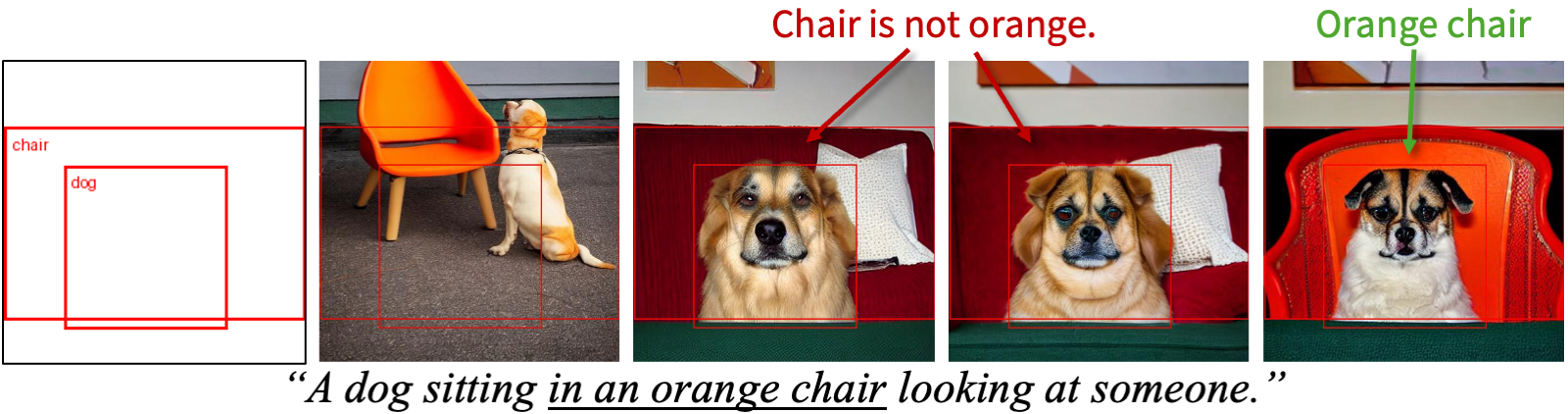}} \\
        \midrule
        %%%%%%%%%%%%%
        \multicolumn{5}{c}{\includegraphics[width=\textwidth]{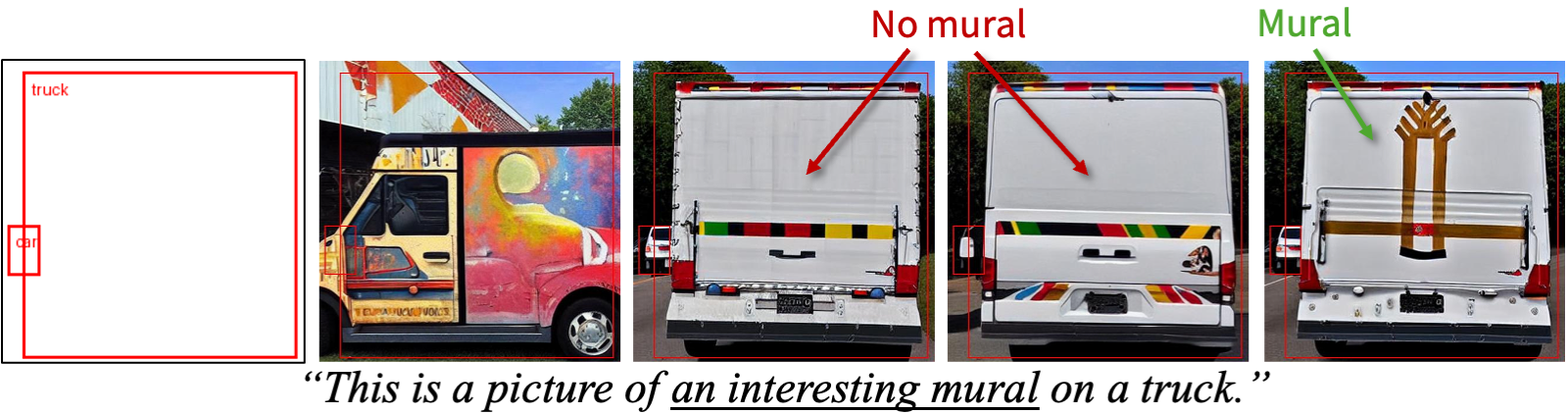}} \\
        %%%%%%%%%%%%%
        \end{tabularx}
    }
    \label{fig:supp_more_qualitative_03}
\end{figure*}
% -----------------------------------------------------------------------

% -----------------------------------------------------------------------
\begin{figure*}[h!]
    \centering
    \footnotesize{
        \renewcommand{\arraystretch}{0.5}
        \setlength{\tabcolsep}{0.0em}
        \setlength{\fboxrule}{0.0pt}
        \setlength{\fboxsep}{0pt}
        %%%%%%%%%%%%%
        \begin{tabularx}{\textwidth}{>{\centering\arraybackslash}m{0.20\textwidth} >{\centering\arraybackslash}m{0.20\textwidth} >{\centering\arraybackslash}m{0.20\textwidth} >{\centering\arraybackslash}m{0.20\textwidth} >{\centering\arraybackslash}m{0.20\textwidth}}
        Layout & SD & GLIGEN$_{\gamma=1.0}$ & GLIGEN$_{\gamma=0.2}$ & \textbf{ReGround} \\
        %%%%%%%%%%%%%%%%%%%%%%%%%
        \multicolumn{5}{c}{\includegraphics[width=\textwidth]{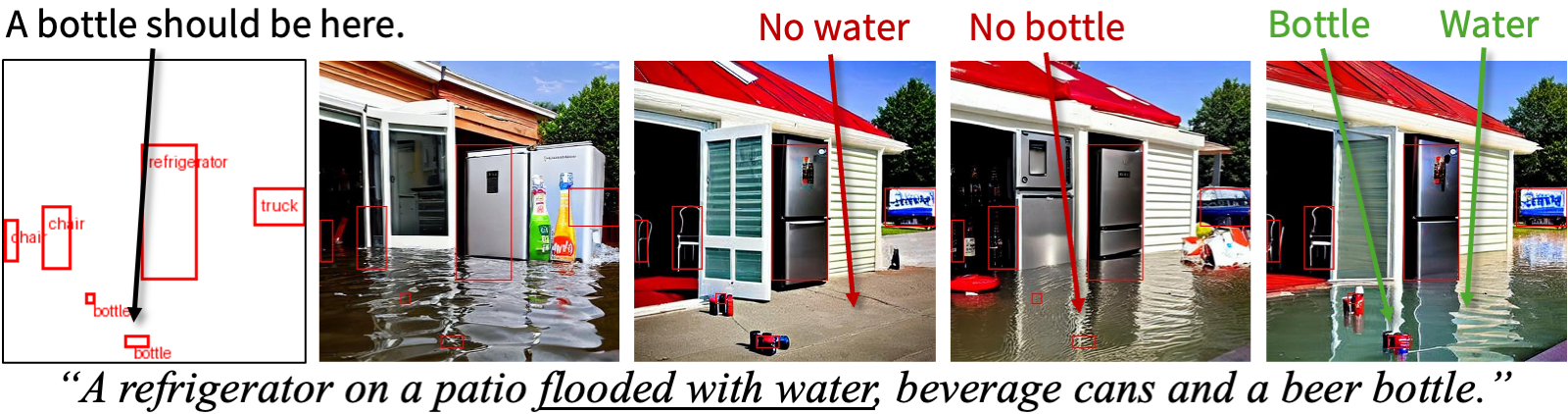}} \\
        \midrule
        %%%%%%%%%%%%%
        \multicolumn{5}{c}{\includegraphics[width=\textwidth]{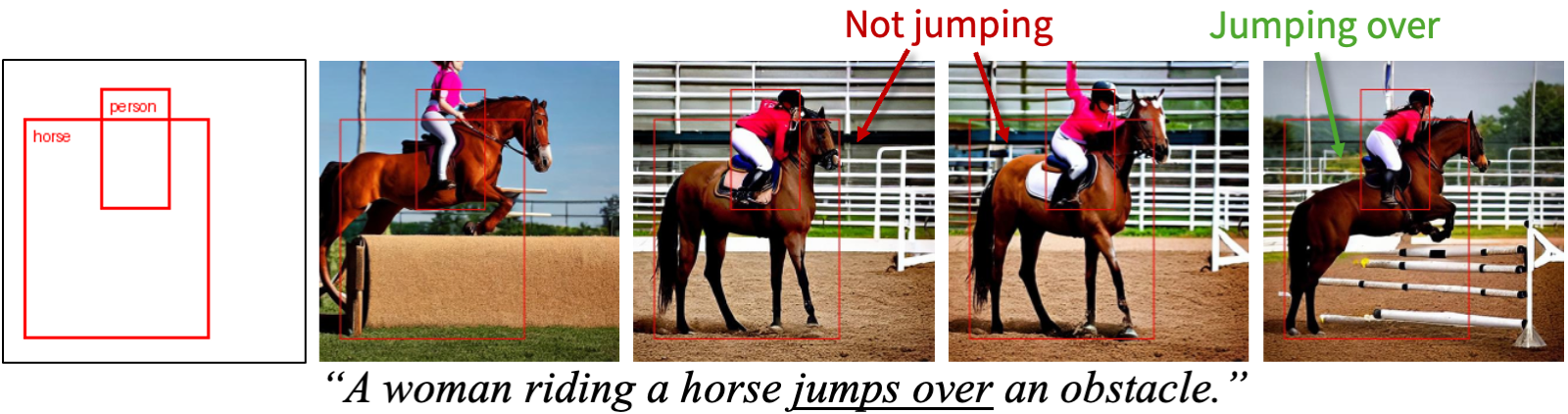}} \\
        \midrule
        %%%%%%%%%%%%%
        %%%%%%%%%%%%%
        \multicolumn{5}{c}{\includegraphics[width=\textwidth]{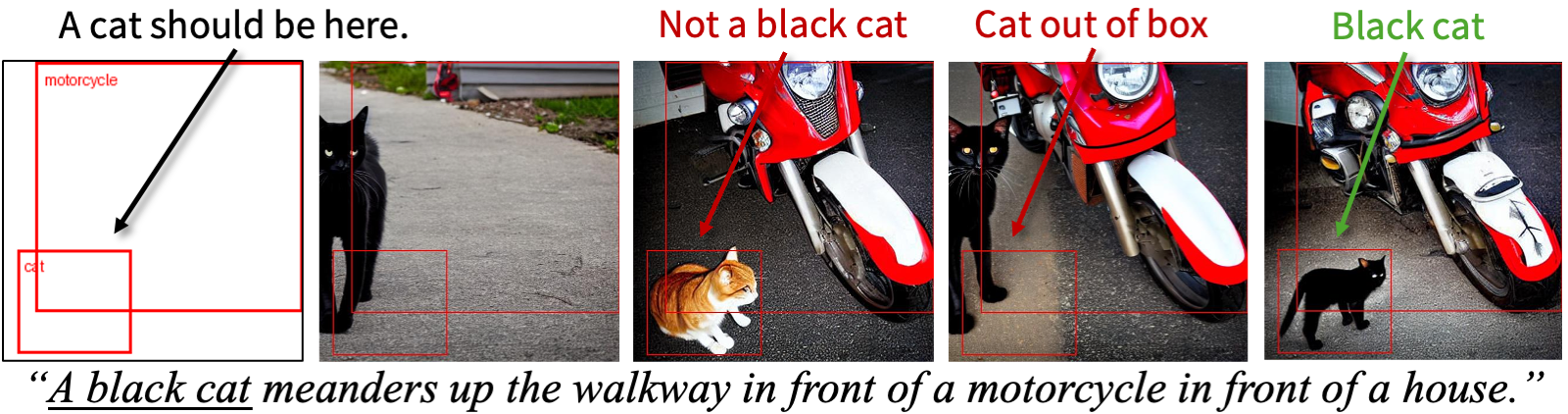}} \\
        \midrule
        %%%%%%%%%%%%%
        \multicolumn{5}{c}{\includegraphics[width=\textwidth]{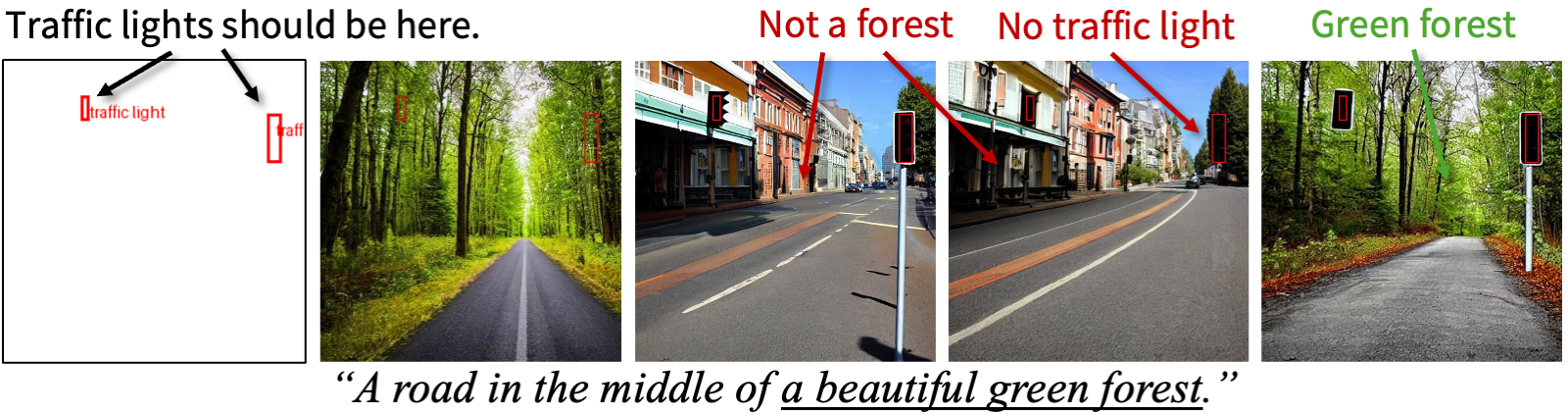}} \\
        \midrule
        %%%%%%%%%%%%%
        \multicolumn{5}{c}{\includegraphics[width=\textwidth]{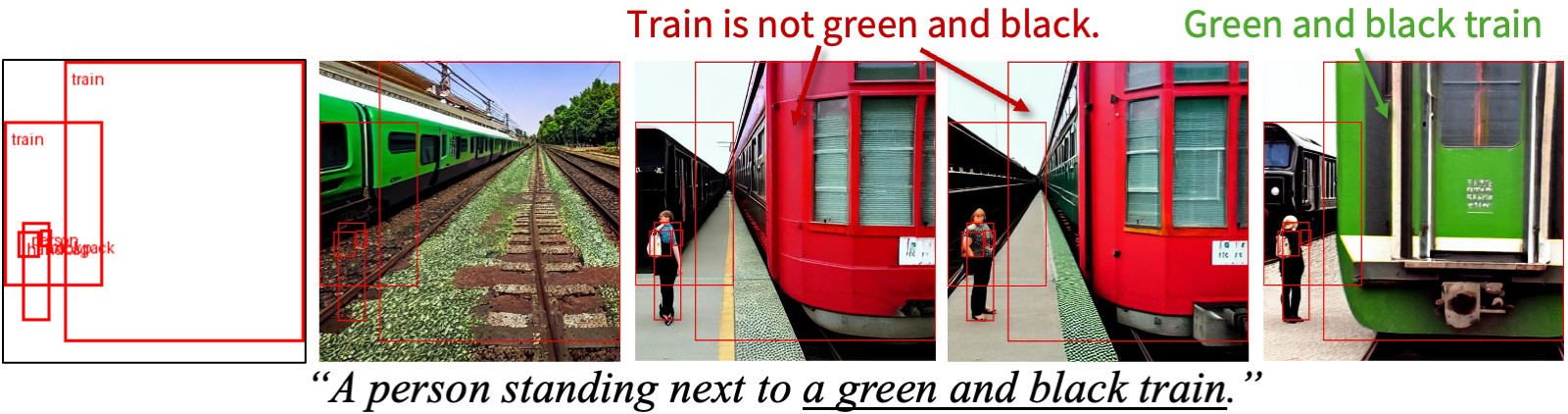}} \\
        \end{tabularx}
    }
    \label{fig:supp_more_qualitative_11}
\end{figure*}
% -----------------------------------------------------------------------

% -----------------------------------------------------------------------
\begin{figure*}[h!]
    \centering
    \footnotesize{
        \renewcommand{\arraystretch}{0.5}
        \setlength{\tabcolsep}{0.0em}
        \setlength{\fboxrule}{0.0pt}
        \setlength{\fboxsep}{0pt}
        %%%%%%%%%%%%%
        \begin{tabularx}{\textwidth}{>{\centering\arraybackslash}m{0.20\textwidth} >{\centering\arraybackslash}m{0.20\textwidth} >{\centering\arraybackslash}m{0.20\textwidth} >{\centering\arraybackslash}m{0.20\textwidth} >{\centering\arraybackslash}m{0.20\textwidth}}
        Layout & SD & GLIGEN$_{\gamma=1.0}$ & GLIGEN$_{\gamma=0.2}$ & \textbf{ReGround} \\
        %%%%%%%%%%%%%%%%%%%%%%%%
        \multicolumn{5}{c}{\includegraphics[width=\textwidth]{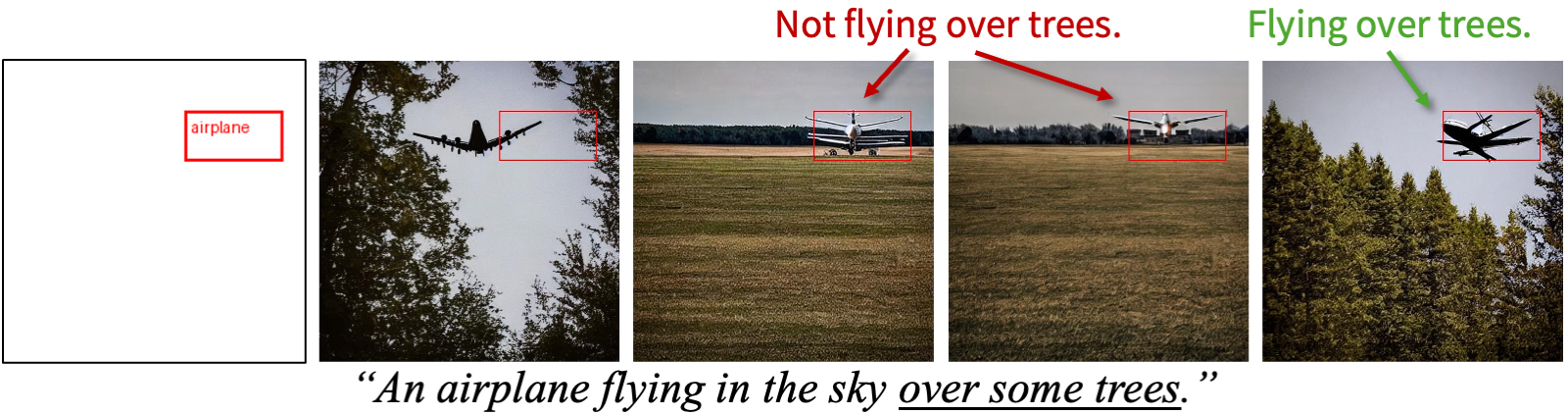}} \\
        \midrule
        %%%%%%%%%%%%%
        \multicolumn{5}{c}{\includegraphics[width=\textwidth]{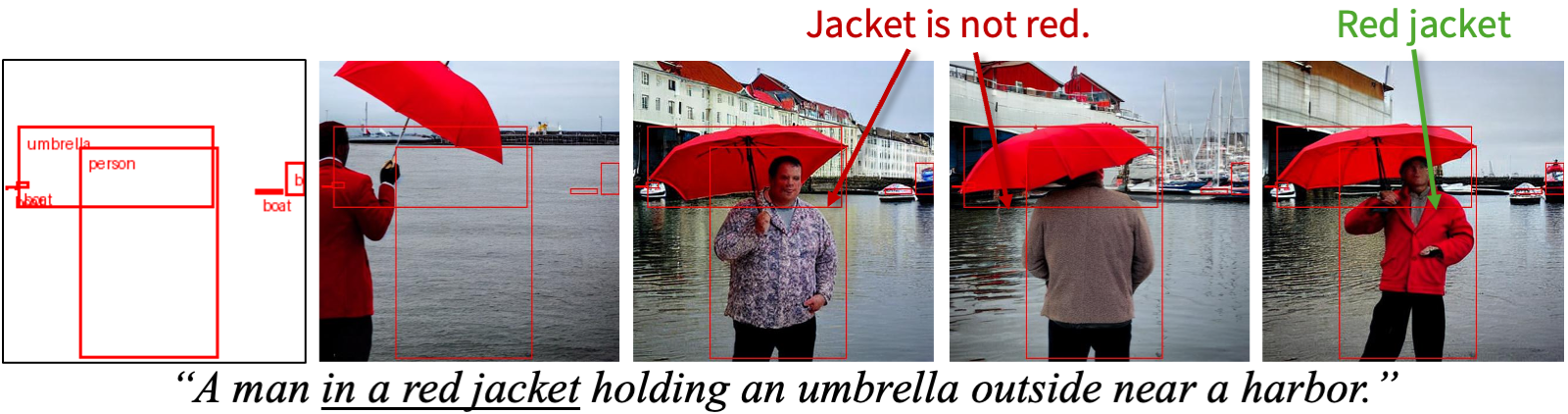}} \\
        \midrule
        % %%%%%%%%%%%%%
        % %%%%%%%%%%%%%
        \multicolumn{5}{c}{\includegraphics[width=\textwidth]{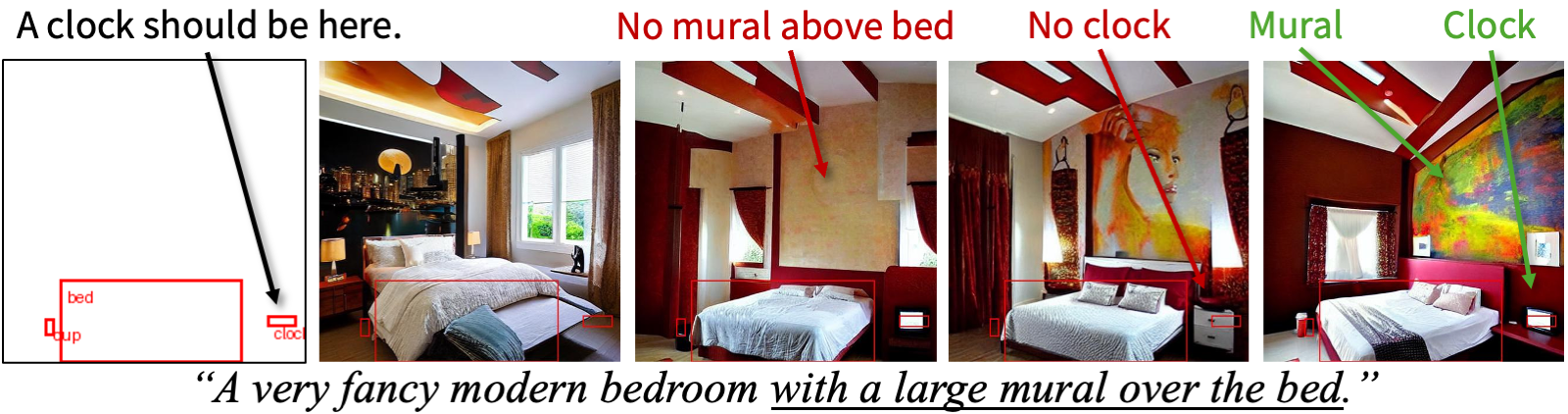}} \\
        \midrule
        % %%%%%%%%%%%%%
        \multicolumn{5}{c}{\includegraphics[width=\textwidth]{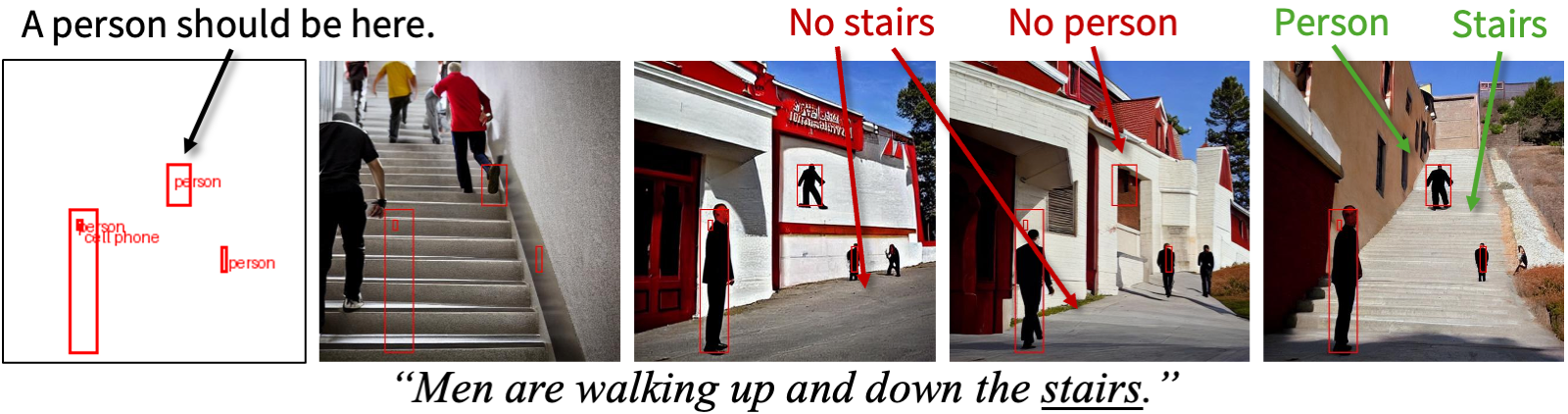}} \\
        \midrule
        % %%%%%%%%%%%%%
        \multicolumn{5}{c}{\includegraphics[width=\textwidth]{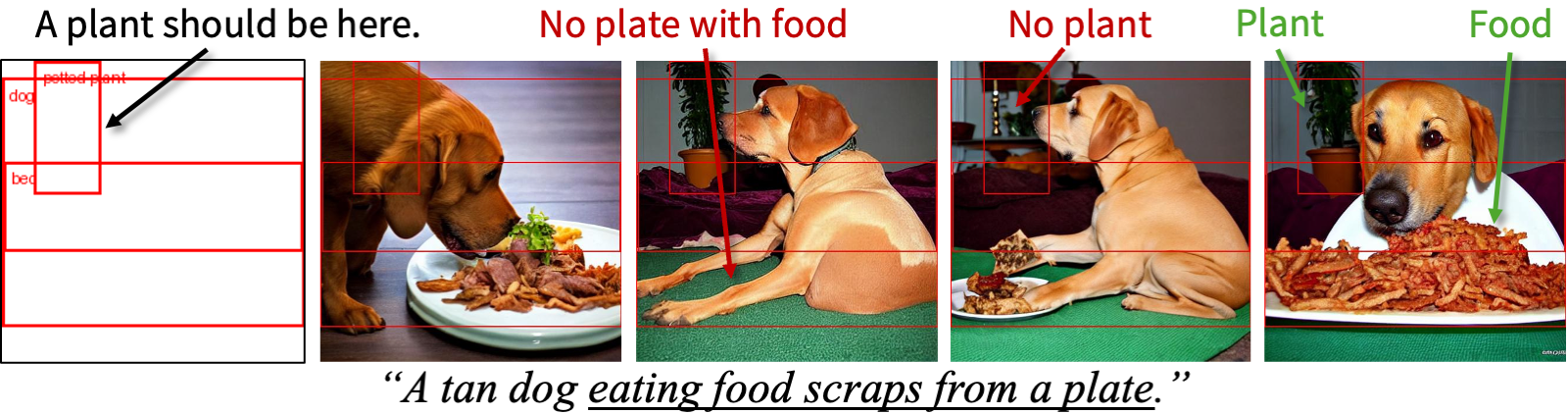}} \\
        \end{tabularx}
    }
    \label{fig:supp_more_qualitative_04}
\end{figure*}
% -----------------------------------------------------------------------

% -----------------------------------------------------------------------
\begin{figure*}[h!]
    \centering
    \footnotesize{
        \renewcommand{\arraystretch}{0.5}
        \setlength{\tabcolsep}{0.0em}
        \setlength{\fboxrule}{0.0pt}
        \setlength{\fboxsep}{0pt}
        %%%%%%%%%%%%%
        \begin{tabularx}{\textwidth}{>{\centering\arraybackslash}m{0.20\textwidth} >{\centering\arraybackslash}m{0.20\textwidth} >{\centering\arraybackslash}m{0.20\textwidth} >{\centering\arraybackslash}m{0.20\textwidth} >{\centering\arraybackslash}m{0.20\textwidth}}
        Layout & SD & GLIGEN$_{\gamma=1.0}$ & GLIGEN$_{\gamma=0.2}$ & \textbf{ReGround} \\
        %%%%%%%%%%%%%%%%%%%%%%%%
        \multicolumn{5}{c}{\includegraphics[width=\textwidth]{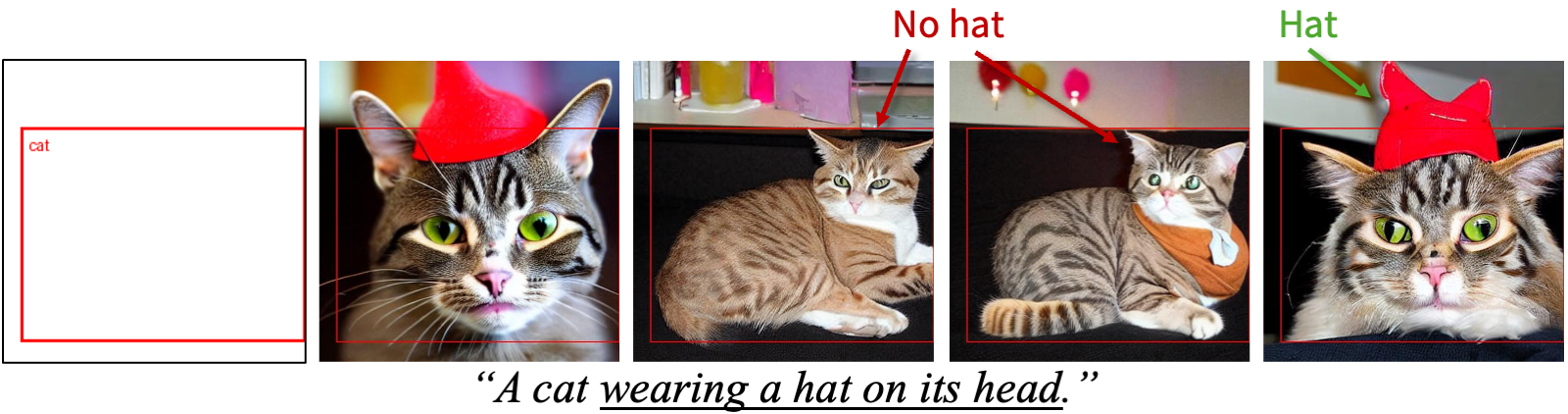}} \\
        \midrule
        %%%%%%%%%%%%%
        \multicolumn{5}{c}{\includegraphics[width=\textwidth]{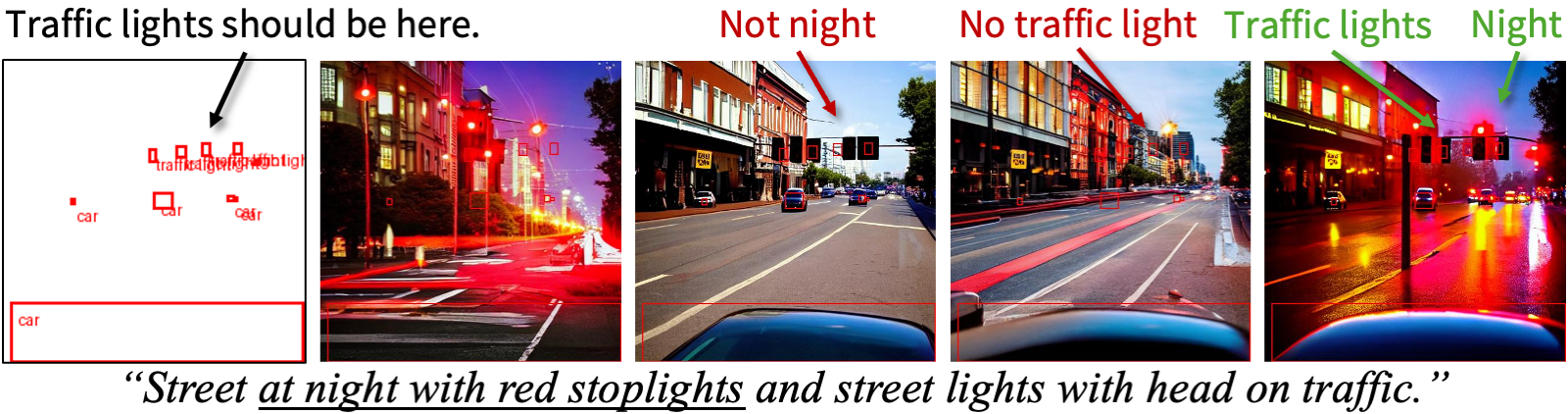}} \\
        \midrule
        % %%%%%%%%%%%%%
        % %%%%%%%%%%%%%
        \multicolumn{5}{c}{\includegraphics[width=\textwidth]{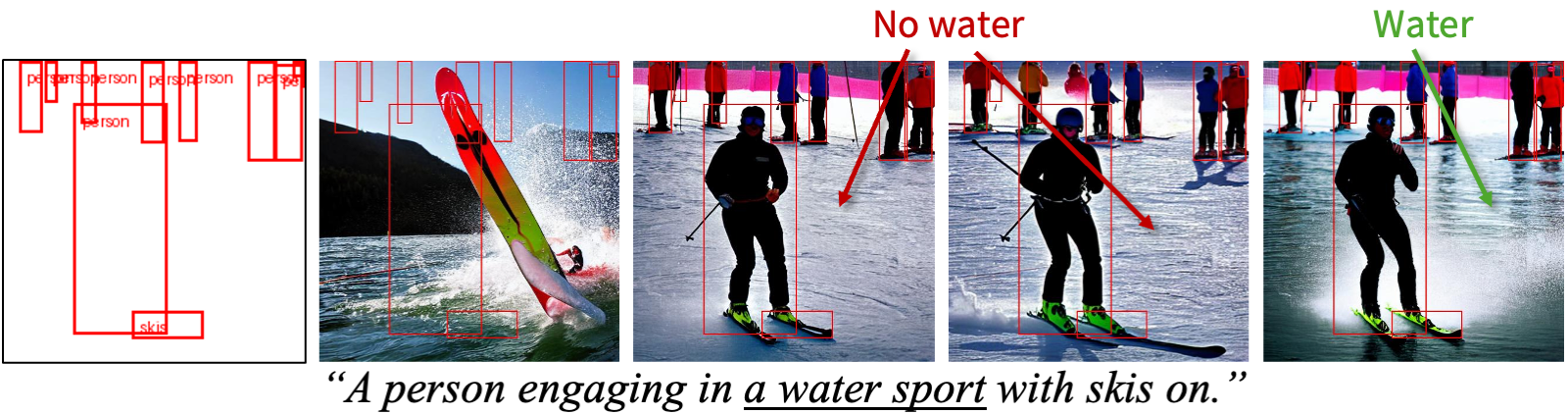}} \\
        \midrule
        % %%%%%%%%%%%%%
        \multicolumn{5}{c}{\includegraphics[width=\textwidth]{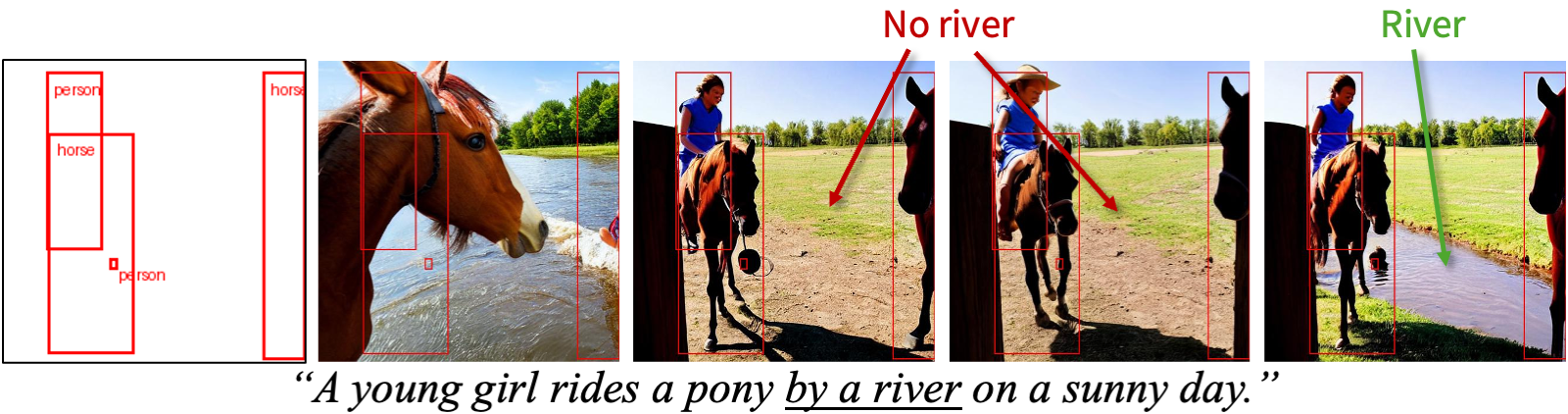}} \\
        \midrule
        % %%%%%%%%%%%%%
        \multicolumn{5}{c}{\includegraphics[width=\textwidth]{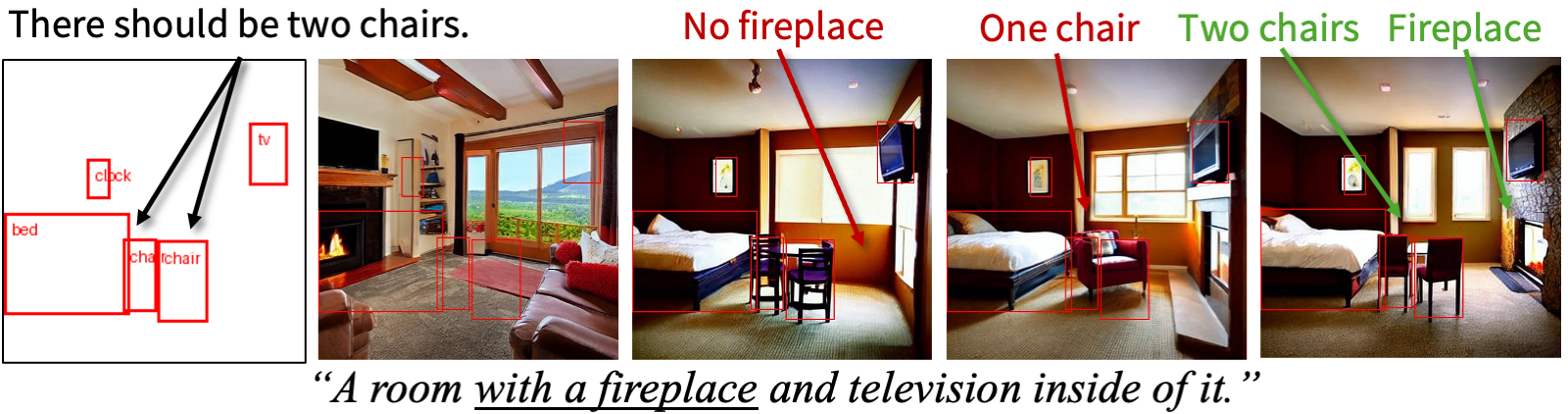}} \\
        \end{tabularx}
    }
    \label{fig:supp_more_qualitative_05}
\end{figure*}
% -----------------------------------------------------------------------

% -----------------------------------------------------------------------
\begin{figure*}[h!]
    \centering
    \footnotesize{
        \renewcommand{\arraystretch}{0.5}
        \setlength{\tabcolsep}{0.0em}
        \setlength{\fboxrule}{0.0pt}
        \setlength{\fboxsep}{0pt}
        %%%%%%%%%%%%%
        \begin{tabularx}{\textwidth}{>{\centering\arraybackslash}m{0.20\textwidth} >{\centering\arraybackslash}m{0.20\textwidth} >{\centering\arraybackslash}m{0.20\textwidth} >{\centering\arraybackslash}m{0.20\textwidth} >{\centering\arraybackslash}m{0.20\textwidth}}
        Layout & SD & GLIGEN$_{\gamma=1.0}$ & GLIGEN$_{\gamma=0.2}$ & \textbf{ReGround} \\
        %%%%%%%%%%%%%%%%%%%%%%%%
        \multicolumn{5}{c}{\includegraphics[width=\textwidth]{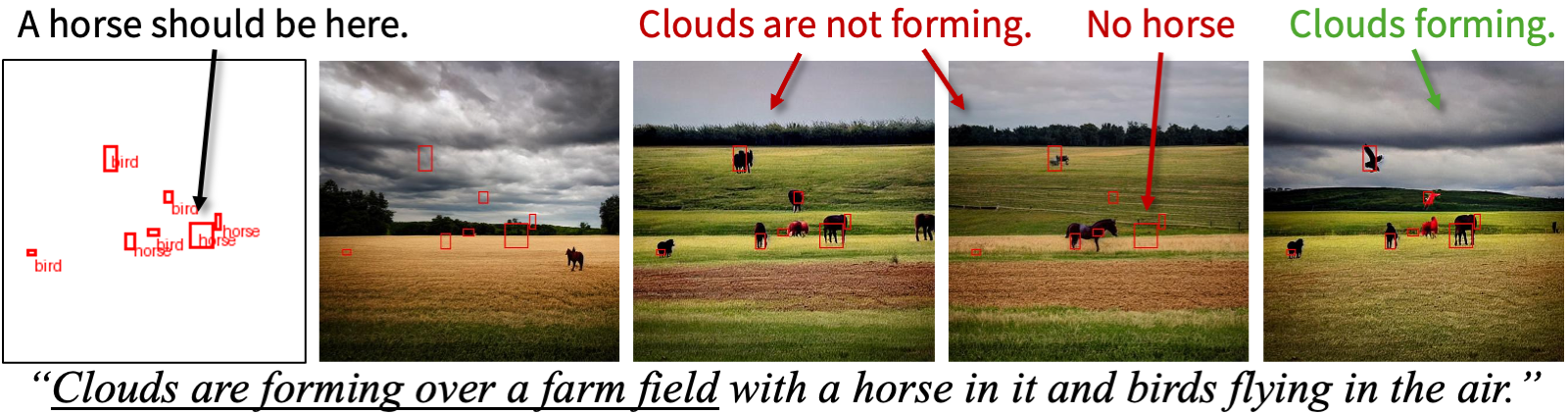}} \\
        \midrule
        %%%%%%%%%%%%%
        \multicolumn{5}{c}{\includegraphics[width=\textwidth]{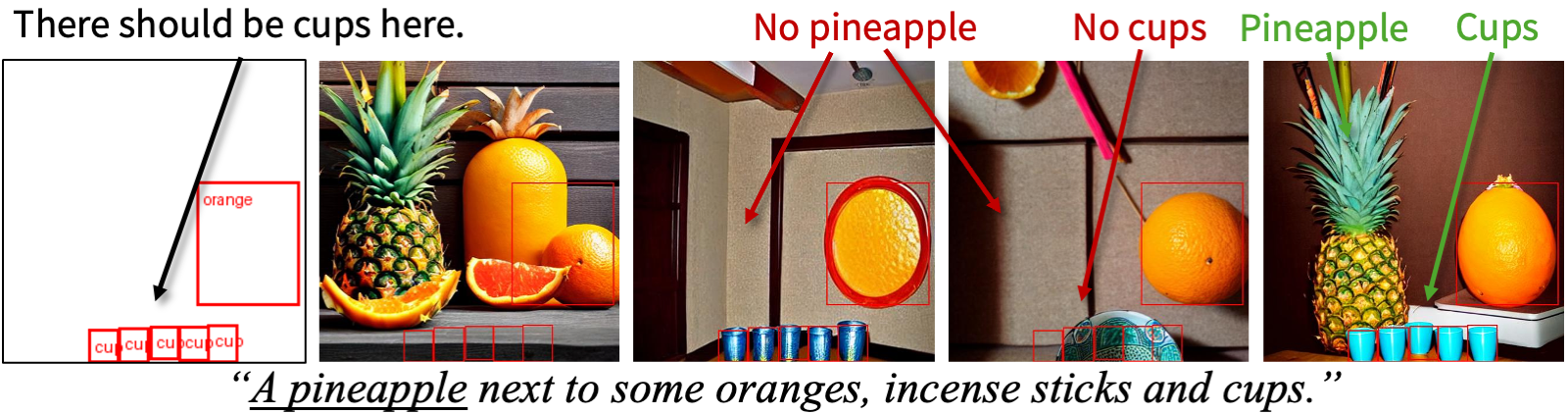}} \\
        \midrule
        % %%%%%%%%%%%%%
        % %%%%%%%%%%%%%
        \multicolumn{5}{c}{\includegraphics[width=\textwidth]{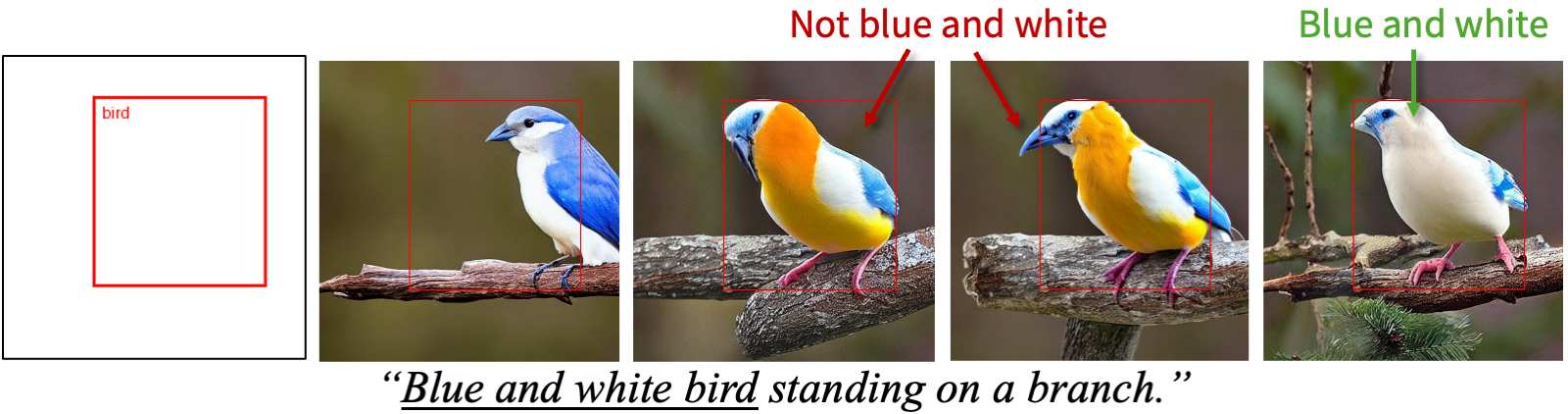}} \\
        \midrule
        % %%%%%%%%%%%%%
        \multicolumn{5}{c}{\includegraphics[width=\textwidth]{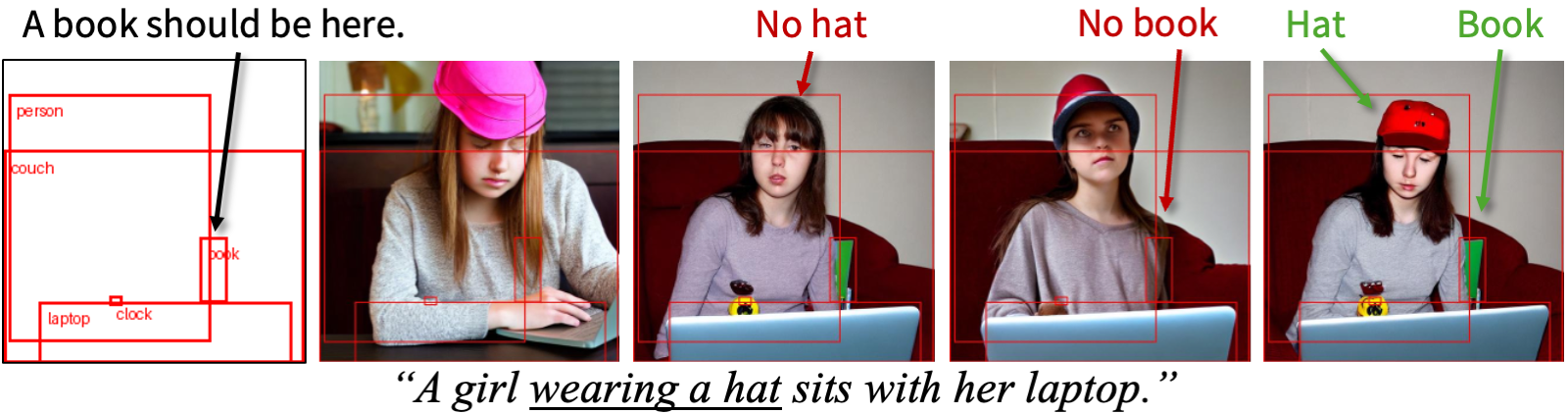}} \\
        \midrule
        % %%%%%%%%%%%%%
        \multicolumn{5}{c}{\includegraphics[width=\textwidth]{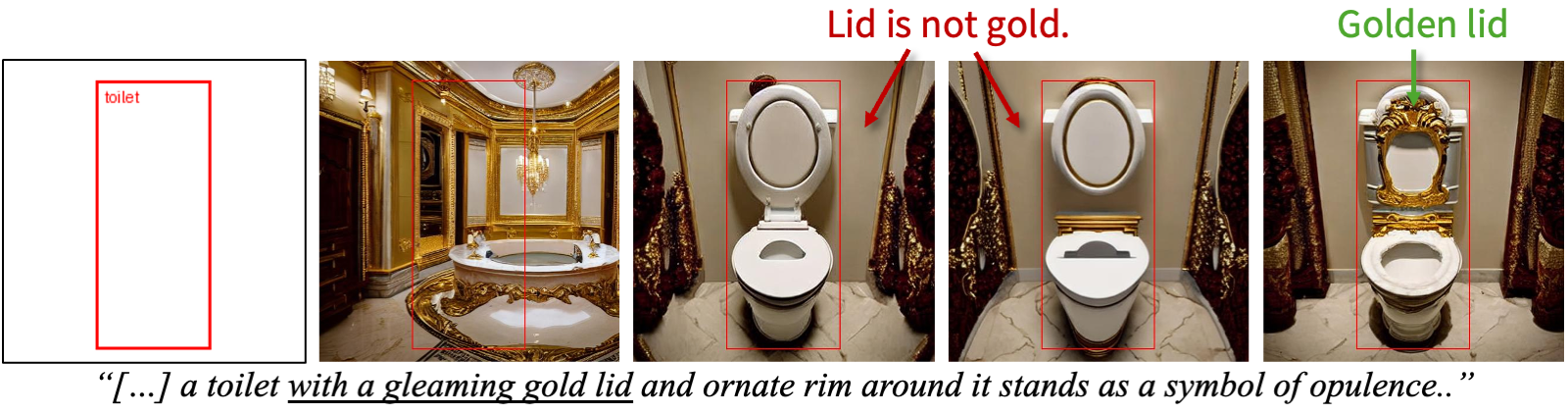}} \\
        \end{tabularx}
    }
    \label{fig:supp_more_qualitative_06}
\end{figure*}
% -----------------------------------------------------------------------

% -----------------------------------------------------------------------
\begin{figure*}[h!]
    \centering
    \footnotesize{
        \renewcommand{\arraystretch}{0.5}
        \setlength{\tabcolsep}{0.0em}
        \setlength{\fboxrule}{0.0pt}
        \setlength{\fboxsep}{0pt}
        %%%%%%%%%%%%%
        \begin{tabularx}{\textwidth}{>{\centering\arraybackslash}m{0.20\textwidth} >{\centering\arraybackslash}m{0.20\textwidth} >{\centering\arraybackslash}m{0.20\textwidth} >{\centering\arraybackslash}m{0.20\textwidth} >{\centering\arraybackslash}m{0.20\textwidth}}
        Layout & SD & GLIGEN$_{\gamma=1.0}$ & GLIGEN$_{\gamma=0.2}$ & \textbf{ReGround} \\
        %%%%%%%%%%%%%%%%%%%%%%%%
        \multicolumn{5}{c}{\includegraphics[width=\textwidth]{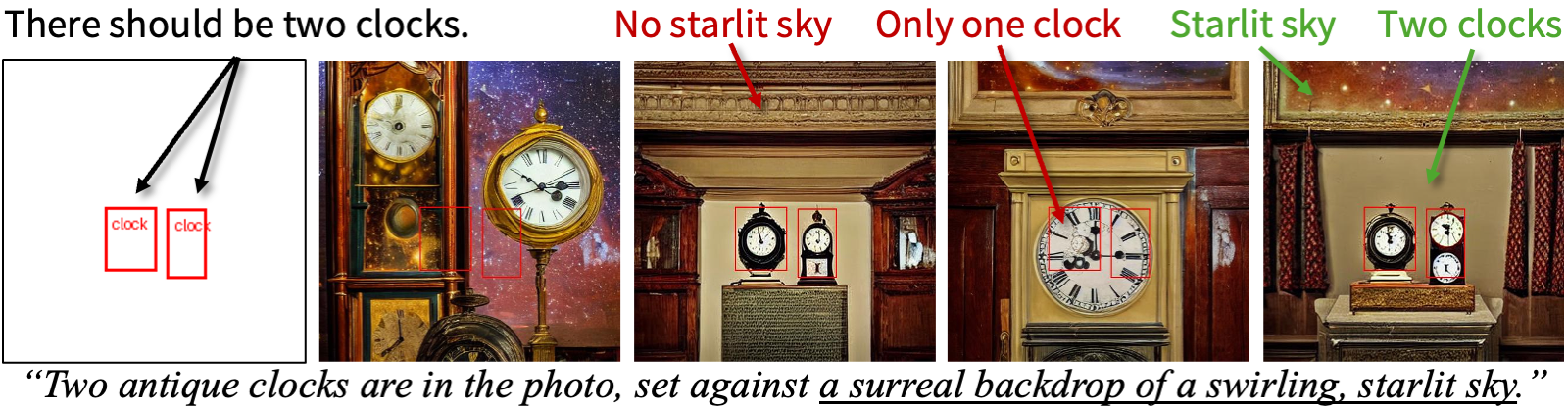}} \\
        \midrule
        %%%%%%%%%%%%%
        \multicolumn{5}{c}{\includegraphics[width=\textwidth]{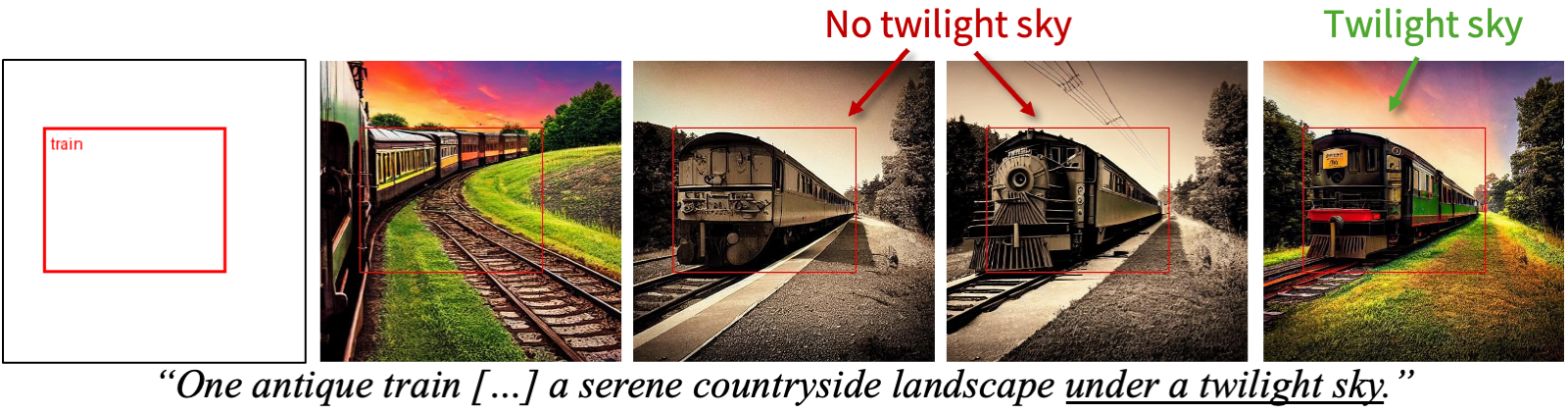}} \\
        \midrule
        % %%%%%%%%%%%%%
        % %%%%%%%%%%%%%
        \multicolumn{5}{c}{\includegraphics[width=\textwidth]{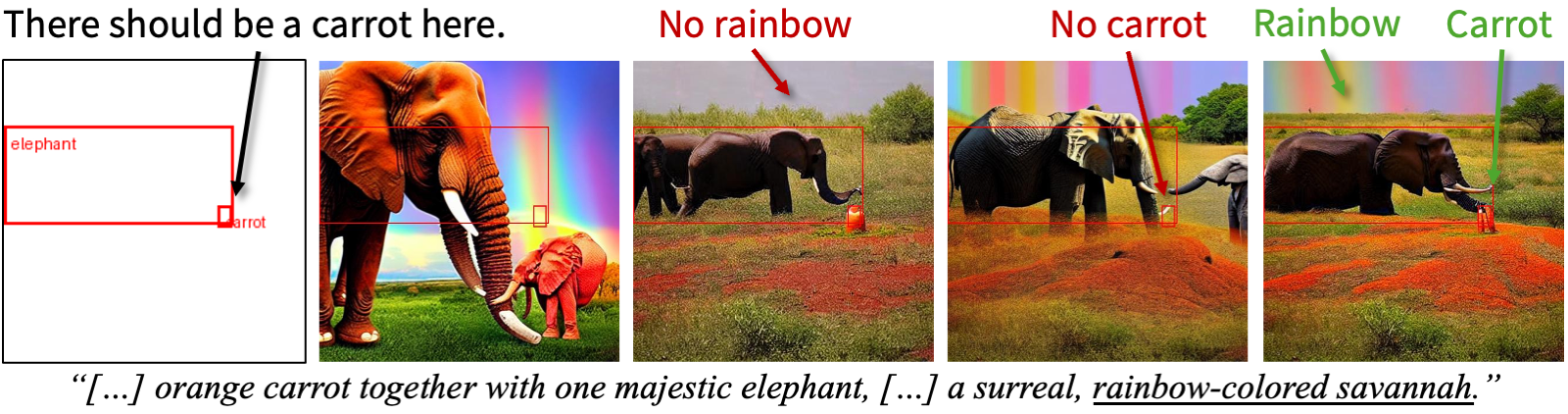}} \\
        \midrule
        % %%%%%%%%%%%%%
        \multicolumn{5}{c}{\includegraphics[width=\textwidth]{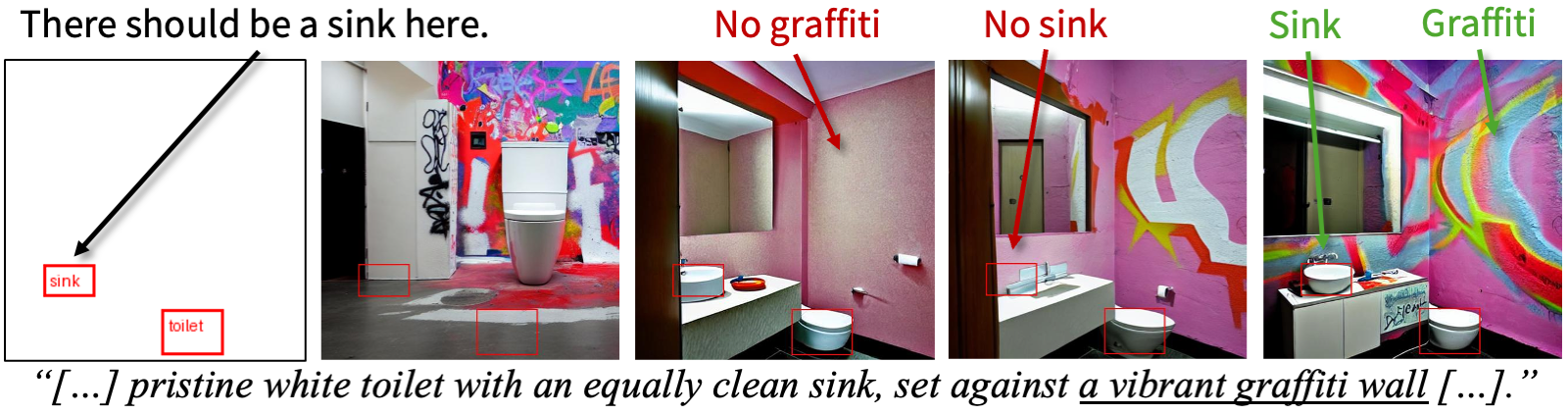}} \\
        \midrule
        % %%%%%%%%%%%%%
        \multicolumn{5}{c}{\includegraphics[width=\textwidth]{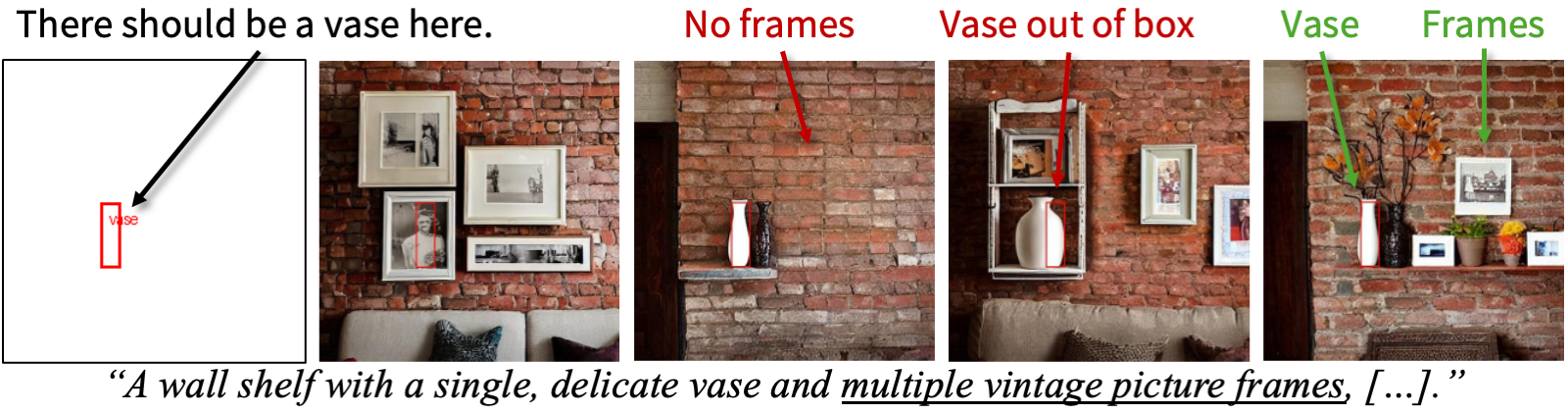}} \\
        \end{tabularx}
    }
    \label{fig:supp_more_qualitative_07}
\end{figure*}
% -----------------------------------------------------------------------

% -----------------------------------------------------------------------
\begin{figure*}[h!]
    \centering
    \footnotesize{
        \renewcommand{\arraystretch}{0.5}
        \setlength{\tabcolsep}{0.0em}
        \setlength{\fboxrule}{0.0pt}
        \setlength{\fboxsep}{0pt}
        %%%%%%%%%%%%%
        \begin{tabularx}{\textwidth}{>{\centering\arraybackslash}m{0.20\textwidth} >{\centering\arraybackslash}m{0.20\textwidth} >{\centering\arraybackslash}m{0.20\textwidth} >{\centering\arraybackslash}m{0.20\textwidth} >{\centering\arraybackslash}m{0.20\textwidth}}
        Layout & SD & GLIGEN$_{\gamma=1.0}$ & GLIGEN$_{\gamma=0.2}$ & \textbf{ReGround} \\
        %%%%%%%%%%%%%%%%%%%%%%%%
        \multicolumn{5}{c}{\includegraphics[width=\textwidth]{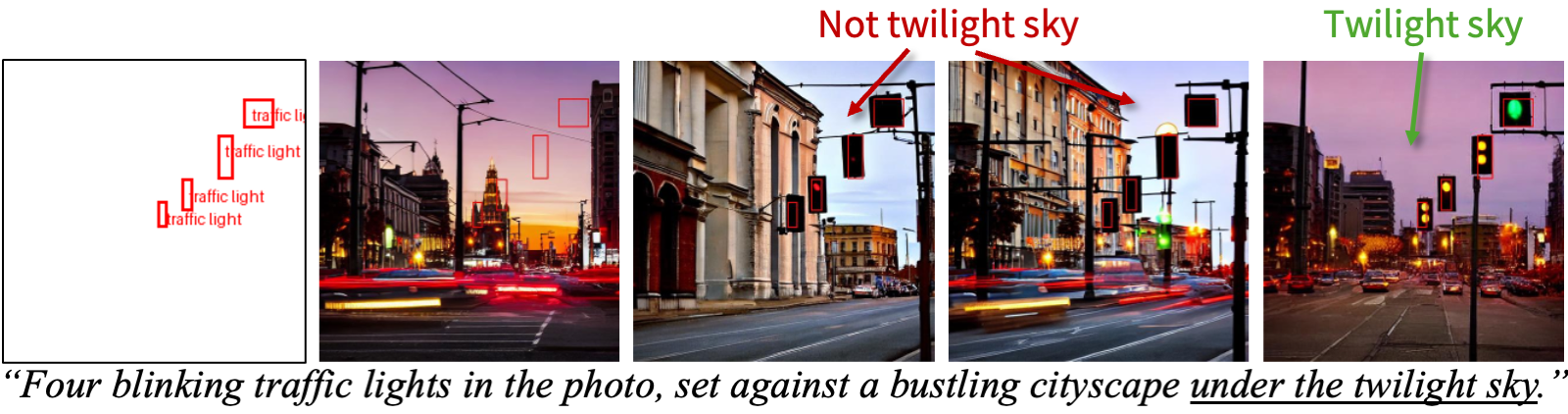}} \\
        \midrule
        %%%%%%%%%%%%%
        %%%%%%%%%%%%%%%%%%%%%%%%
        \multicolumn{5}{c}{\includegraphics[width=\textwidth]{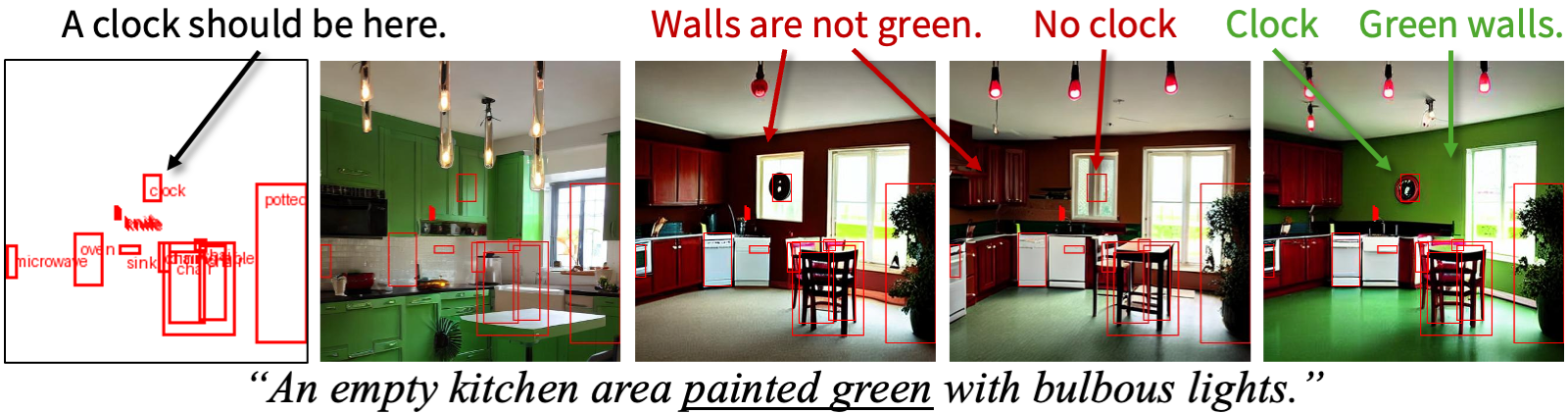}} \\
        \midrule
        %%%%%%%%%%%%%%%%%%%%%%%%
        \multicolumn{5}{c}{\includegraphics[width=\textwidth]{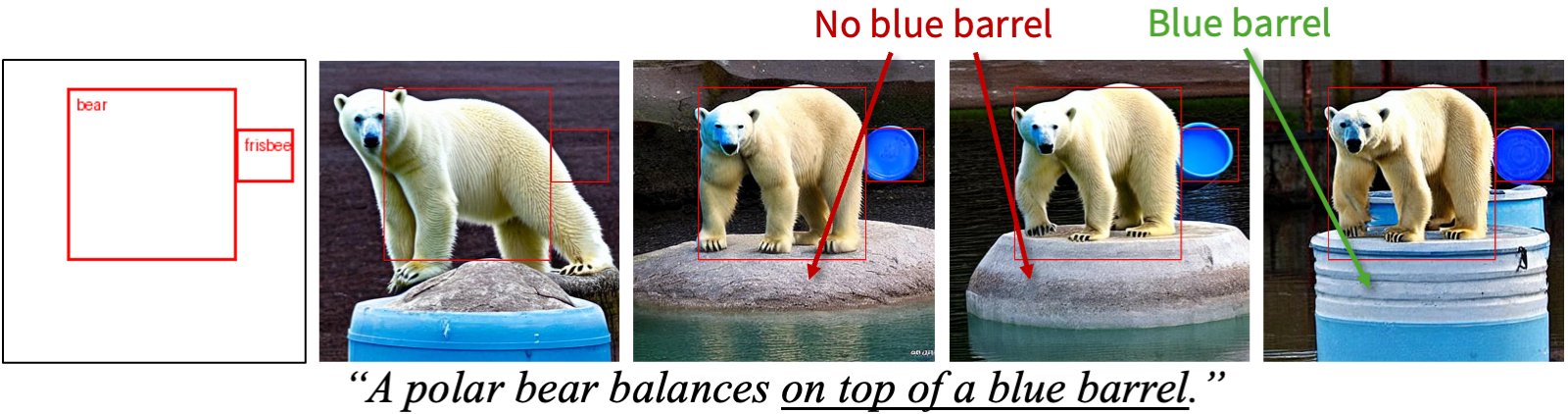}} \\
        \midrule
        %%%%%%%%%%%%%%%%%%%%%%%%
        \multicolumn{5}{c}{\includegraphics[width=\textwidth]{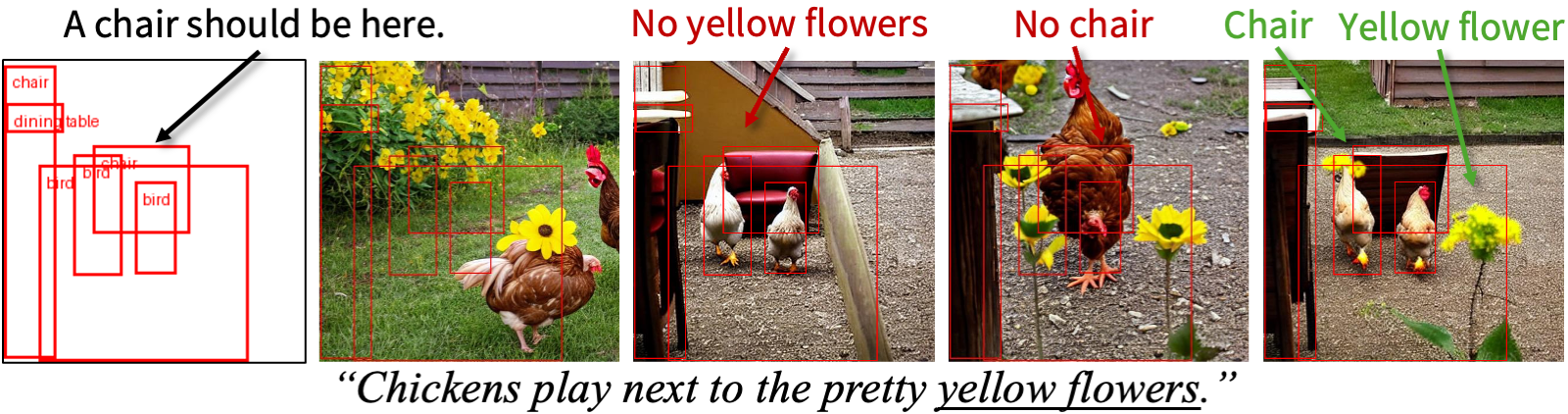}} \\
        \midrule
        %%%%%%%%%%%%%%%%%%%%%%%%
        \multicolumn{5}{c}{\includegraphics[width=\textwidth]{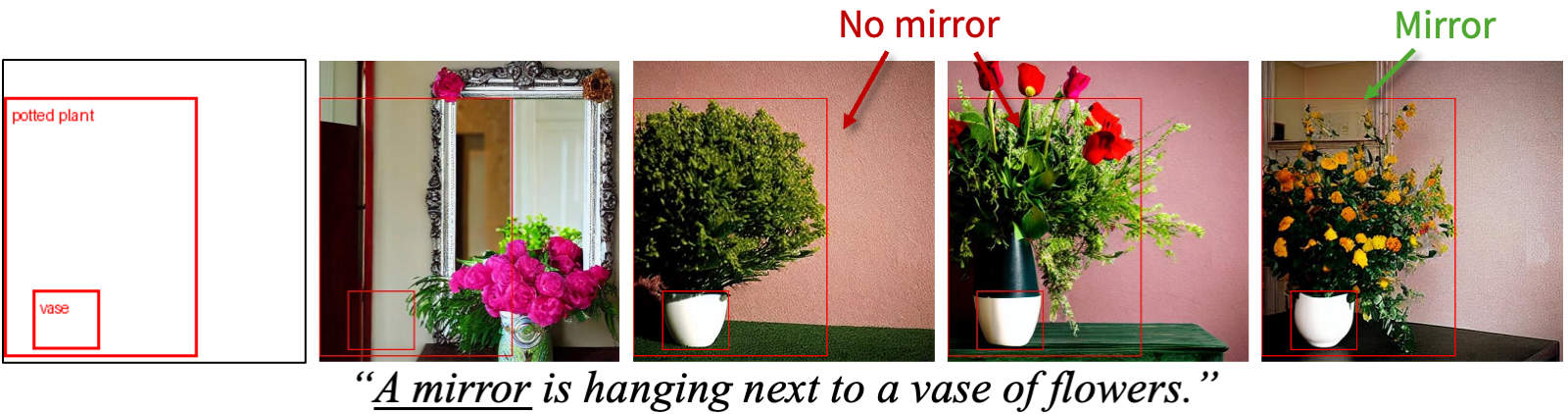}} \\
        \end{tabularx}
    }
    \label{fig:supp_more_qualitative_08}
\end{figure*}
% -----------------------------------------------------------------------

% -----------------------------------------------------------------------
\begin{figure*}[h!]
    \centering
    \footnotesize{
        \renewcommand{\arraystretch}{0.5}
        \setlength{\tabcolsep}{0.0em}
        \setlength{\fboxrule}{0.0pt}
        \setlength{\fboxsep}{0pt}
        %%%%%%%%%%%%%
        \begin{tabularx}{\textwidth}{>{\centering\arraybackslash}m{0.20\textwidth} >{\centering\arraybackslash}m{0.20\textwidth} >{\centering\arraybackslash}m{0.20\textwidth} >{\centering\arraybackslash}m{0.20\textwidth} >{\centering\arraybackslash}m{0.20\textwidth}}
        Layout & SD & GLIGEN$_{\gamma=1.0}$ & GLIGEN$_{\gamma=0.2}$ & \textbf{ReGround} \\
        %%%%%%%%%%%%%%%%%%%%%%%%
        \multicolumn{5}{c}{\includegraphics[width=\textwidth]{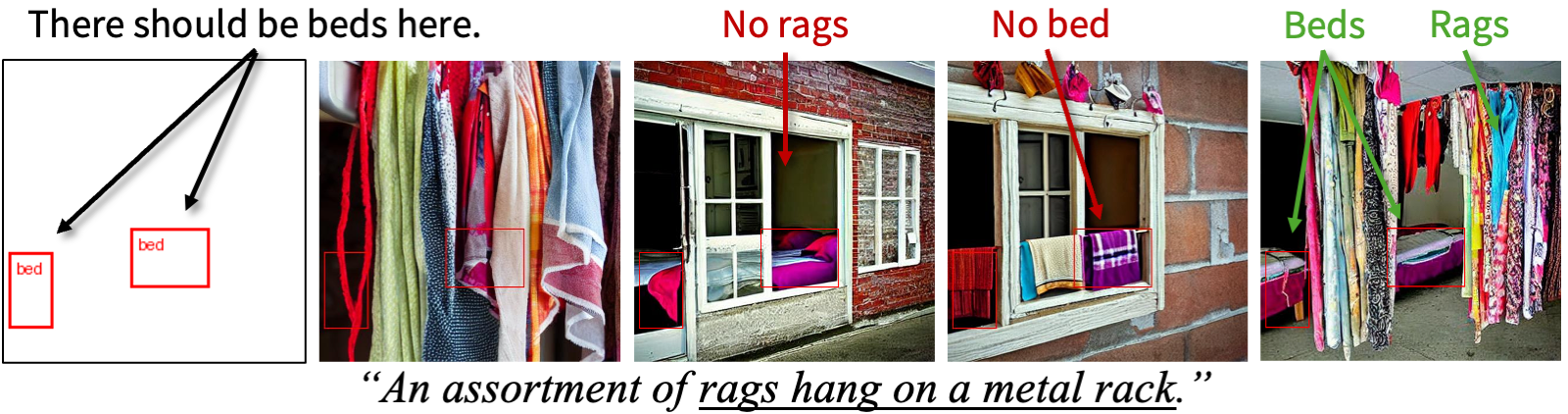}} \\
        \midrule
        %%%%%%%%%%%%%
        %%%%%%%%%%%%%%%%%%%%%%%%
        \multicolumn{5}{c}{\includegraphics[width=\textwidth]{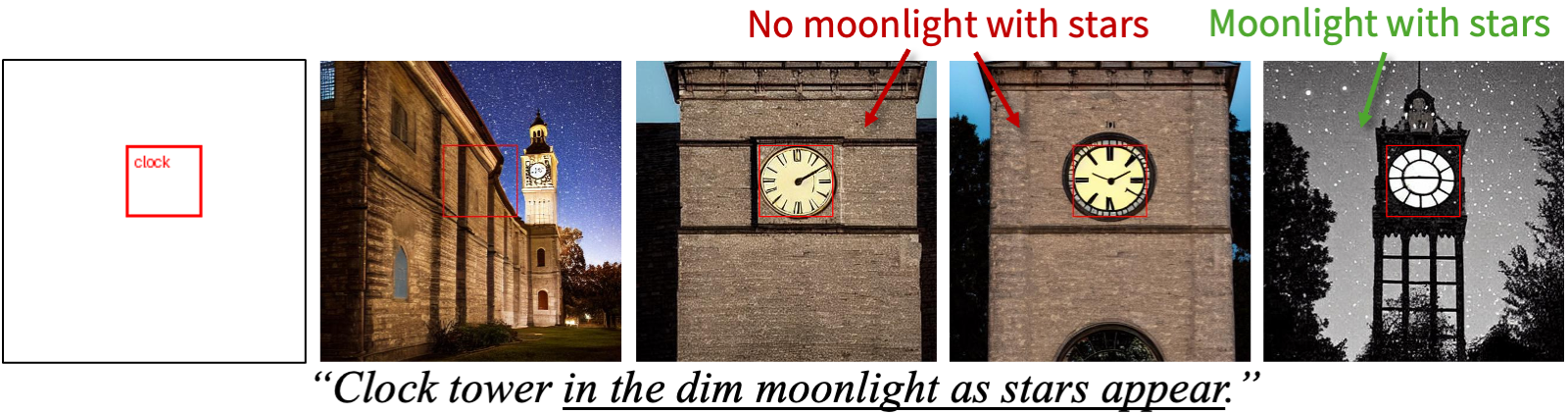}} \\
        \midrule
        %%%%%%%%%%%%%%%%%%%%%%%%
        \multicolumn{5}{c}{\includegraphics[width=\textwidth]{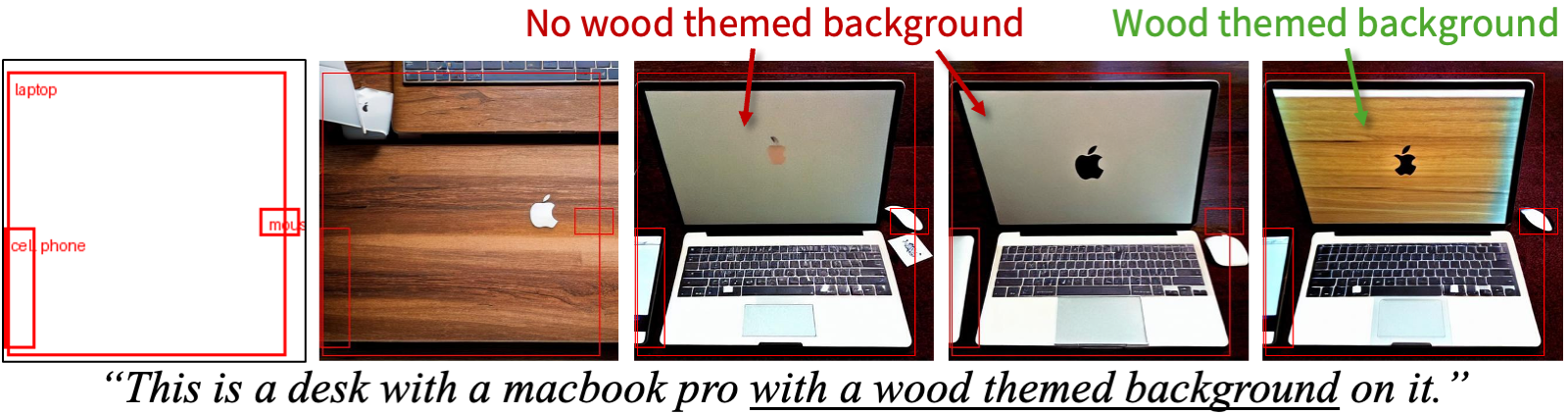}} \\
        \midrule
        %%%%%%%%%%%%%%%%%%%%%%%%
        \multicolumn{5}{c}{\includegraphics[width=\textwidth]{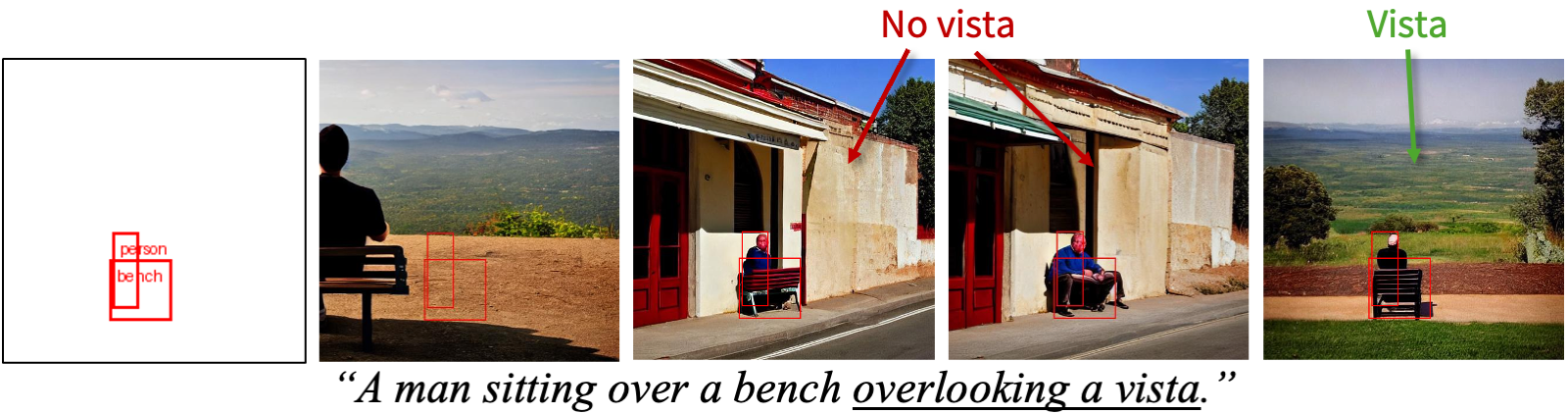}} \\
        \midrule
        %%%%%%%%%%%%%%%%%%%%%%%%
        \multicolumn{5}{c}{\includegraphics[width=\textwidth]{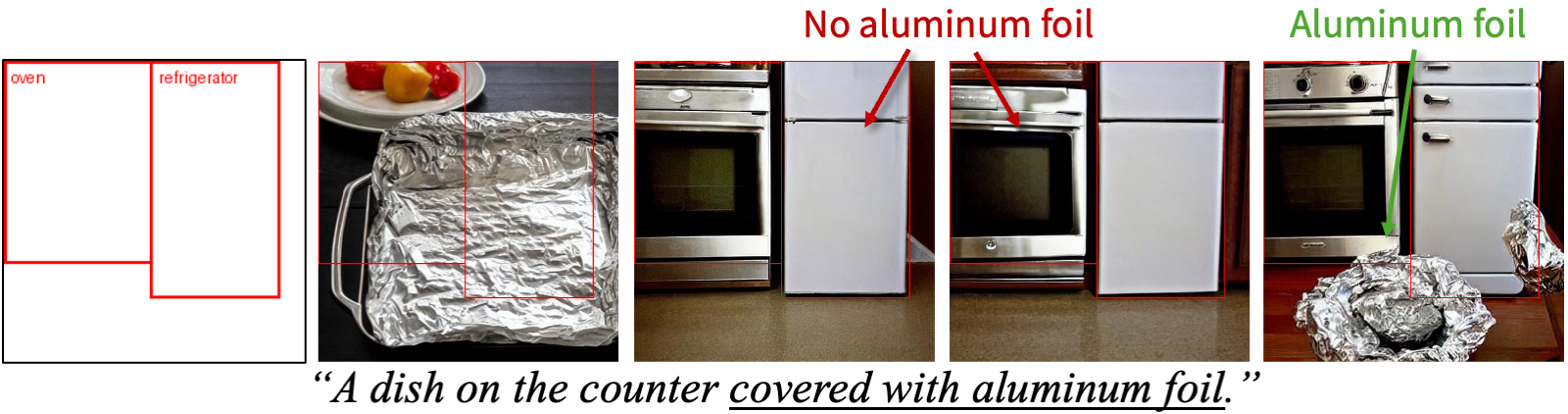}} \\
        \end{tabularx}
    }
    \label{fig:supp_more_qualitative_09}
\end{figure*}
% -----------------------------------------------------------------------

% -----------------------------------------------------------------------
\begin{figure*}[h!]
    \centering
    \footnotesize{
        \renewcommand{\arraystretch}{0.5}
        \setlength{\tabcolsep}{0.0em}
        \setlength{\fboxrule}{0.0pt}
        \setlength{\fboxsep}{0pt}
        %%%%%%%%%%%%%
        \begin{tabularx}{\textwidth}{>{\centering\arraybackslash}m{0.20\textwidth} >{\centering\arraybackslash}m{0.20\textwidth} >{\centering\arraybackslash}m{0.20\textwidth} >{\centering\arraybackslash}m{0.20\textwidth} >{\centering\arraybackslash}m{0.20\textwidth}}
        Layout & SD & GLIGEN$_{\gamma=1.0}$ & GLIGEN$_{\gamma=0.2}$ & \textbf{ReGround} \\
        %%%%%%%%%%%%%
        \multicolumn{5}{c}{\includegraphics[width=\textwidth]{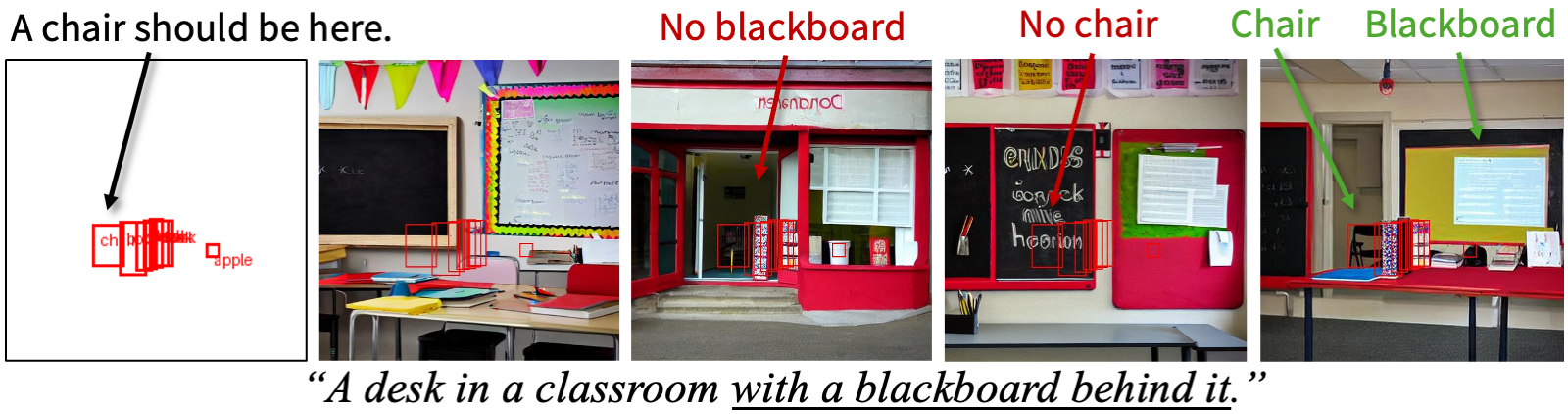}} \\
        \midrule
        %%%%%%%%%%%%%%%%%%%%%%%%
        \multicolumn{5}{c}{\includegraphics[width=\textwidth]{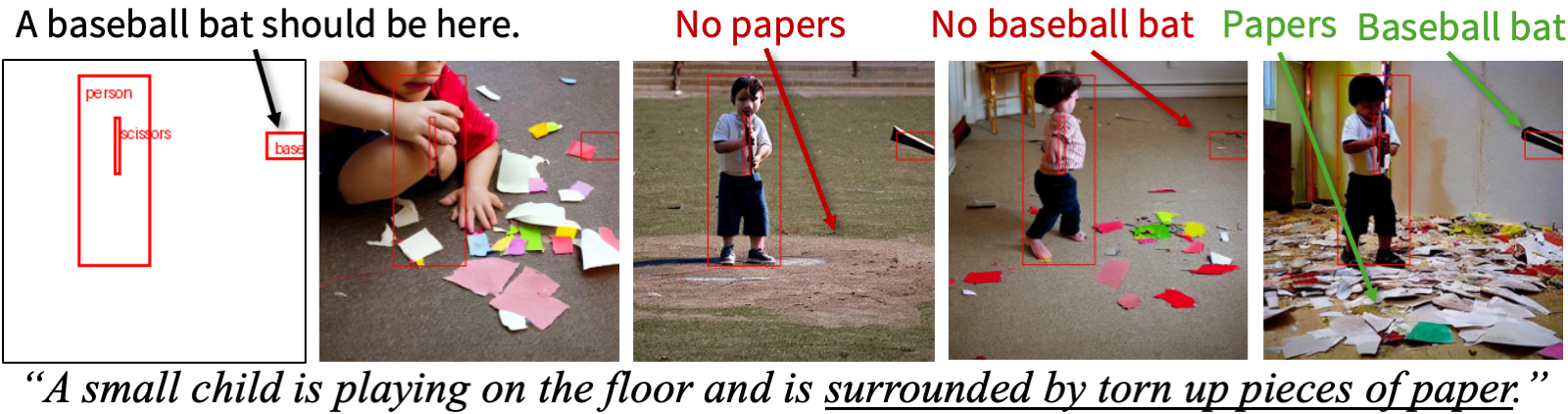}} \\
        \midrule
        %%%%%%%%%%%%%
        %%%%%%%%%%%%%%%%%%%%%%%%
        \multicolumn{5}{c}{\includegraphics[width=\textwidth]{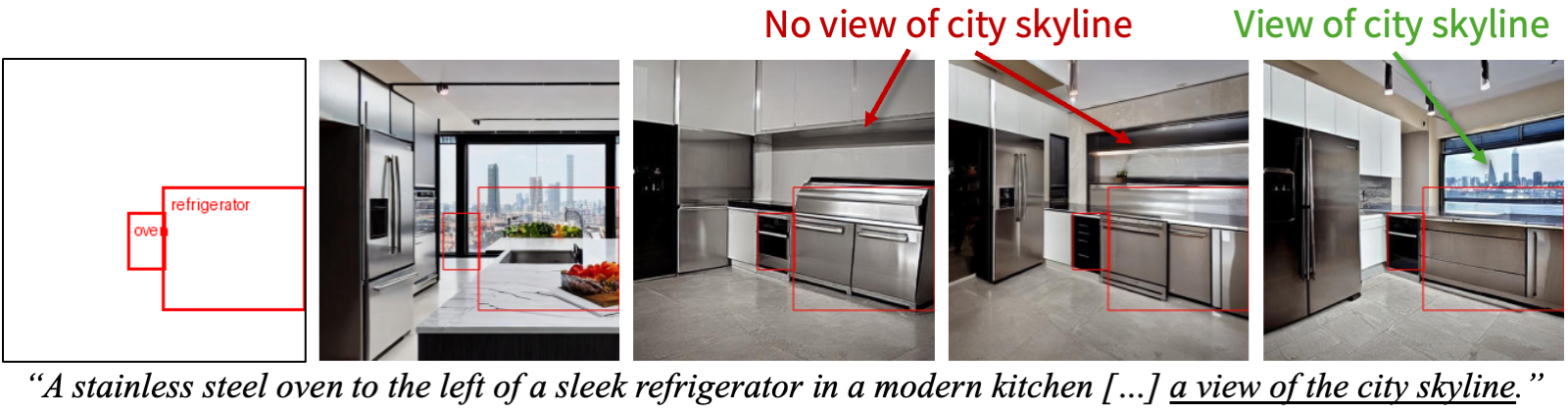}} \\
        \midrule
        %%%%%%%%%%%%%%%%%%%%%%%%
        \multicolumn{5}{c}{\includegraphics[width=\textwidth]{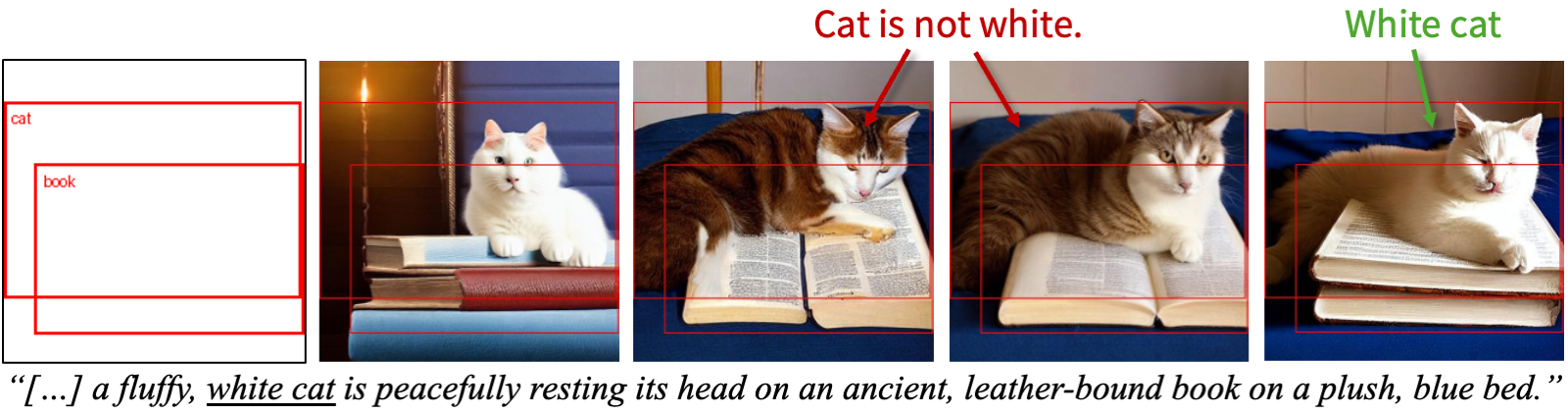}} \\
        \end{tabularx}
    }
    \label{fig:supp_more_qualitative_10}
\end{figure*}
% -----------------------------------------------------------------------
\clearpage